\documentclass[10pt,twocolumn,letterpaper]{article}

\usepackage{cvpr}
\usepackage{times}
\usepackage{epsfig}
\usepackage{graphicx}
\usepackage{amsmath, bm}
\usepackage{amssymb}
\usepackage{comment}
\usepackage{subcaption}
\usepackage{caption}
\DeclareMathOperator*{\argmax}{arg\,max}
\DeclareMathOperator*{\argmin}{arg\,min}
\DeclareMathOperator{\clamp}{clamp}

\usepackage{graphicx}
\usepackage{tabularx} 
\usepackage{hhline}
\newcommand{\rulesep}{\unskip\ \vrule\ }

\usepackage{subcaption} 
\usepackage[font=small,labelfont=bf]{caption} 

\usepackage[breaklinks=true,bookmarks=false]{hyperref}

\cvprfinalcopy 

\def\cvprPaperID{****} 

\newcommand\jj[1]{\textcolor{blue}{\textbf{#1}}}

\newcommand\pf[1]{\textcolor{red}{\textbf{#1}}}

\usepackage[symbol]{footmisc}

\renewcommand{\thefootnote}{\fnsymbol{footnote}}

\begin{document}

\title{DeepSDF: Learning Continuous Signed Distance Functions  \\ for Shape Representation}

\twocolumn[{%
\renewcommand\twocolumn[1][]{#1}%
\maketitle
\vspace{-1.3cm}
\begin{center}
		{
		\vspace{-1.0cm}
		\normalsize
		Jeong Joon Park$^{1,3\dagger}$~~~~~~Peter Florence $^{2,3\dagger}$~~~~~~Julian Straub$^3$~~~~~~Richard Newcombe$^3$~~~~~~Steven Lovegrove$^3$\\ \vspace{0.3cm}
		$^1$University of Washington~~~~~~$^2$Massachusetts Institute of Technology~~~~~~$^3$Facebook Reality Labs		}
\centering

\vspace{0.3cm}
\includegraphics[width=\linewidth]{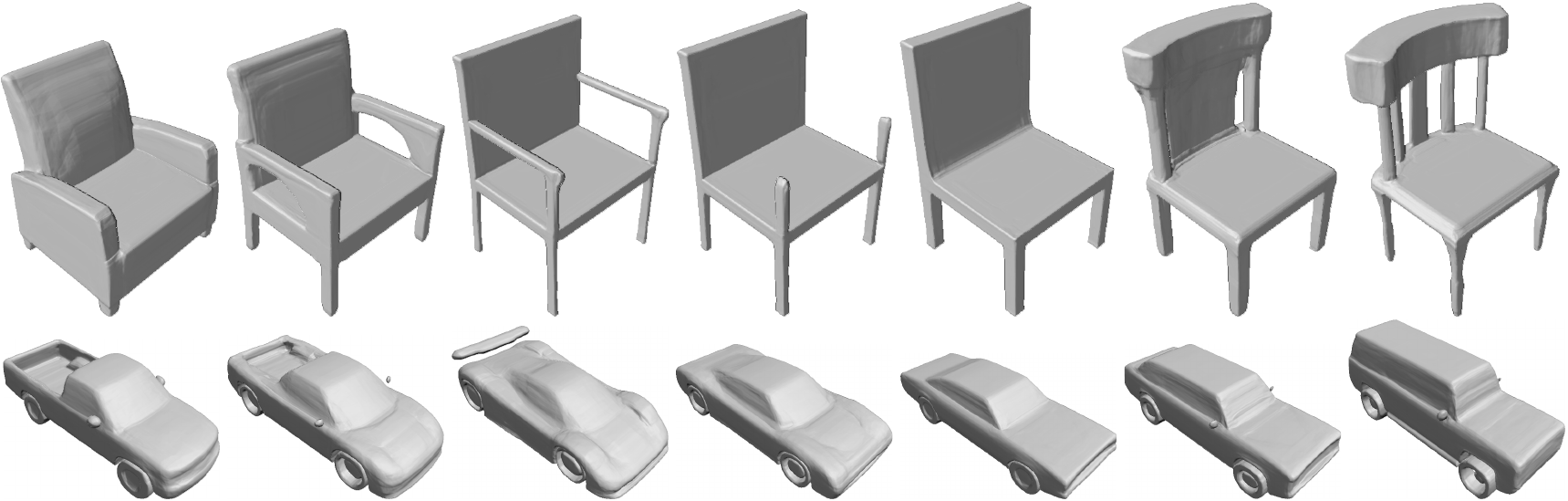}
    \captionof{figure}{DeepSDF represents signed distance functions (SDFs) of shapes via latent code-conditioned feed-forward decoder networks. Above images are raycast renderings of DeepSDF interpolating between two shapes in the learned shape latent space. Best viewed digitally.}    \label{fig:teaser}
\end{center}%
}]

\begin{abstract}
\vspace{-0.5cm}
Computer graphics, 3D computer vision and robotics communities have produced multiple approaches to representing 3D geometry for rendering and reconstruction. These provide trade-offs across fidelity, efficiency and compression capabilities. In this work, we introduce DeepSDF, a learned continuous Signed Distance Function (SDF) representation of a class of shapes that enables high quality shape representation, interpolation and completion from partial and noisy 3D input data. DeepSDF, like its classical counterpart, represents a shape's surface by a continuous volumetric field: the magnitude of a point in the field represents the distance to the surface boundary and the sign indicates whether the region is inside (-) or outside (+) of the shape, hence our representation implicitly encodes a shape's boundary as the zero-level-set of the learned function while explicitly representing the classification of space as being part of the shapes interior or not. While classical SDF's both in analytical or discretized voxel form typically represent the surface of a single shape, DeepSDF can represent an entire class of shapes. Furthermore, we show state-of-the-art performance for learned 3D shape representation and completion while reducing the model size by an order of magnitude compared with previous work.

\end{abstract}
\vspace{-1cm}
\let\thefootnote\relax\footnotetext{$\dagger$ \text{Work performed during internship at Facebook Reality Labs.}}

\section{Introduction} 

Deep convolutional networks which are a mainstay of image-based approaches grow quickly in space and time complexity when directly generalized to the 3rd spatial dimension, and more classical and compact surface representations such as triangle or quad meshes pose problems in training since we may need to deal with an unknown number of vertices and arbitrary topology. These challenges have limited the quality, flexibility and fidelity of deep learning approaches when attempting to either input 3D data for processing or produce 3D inferences for object segmentation and reconstruction.

\begin{figure}
\centering
\includegraphics[width=\linewidth]{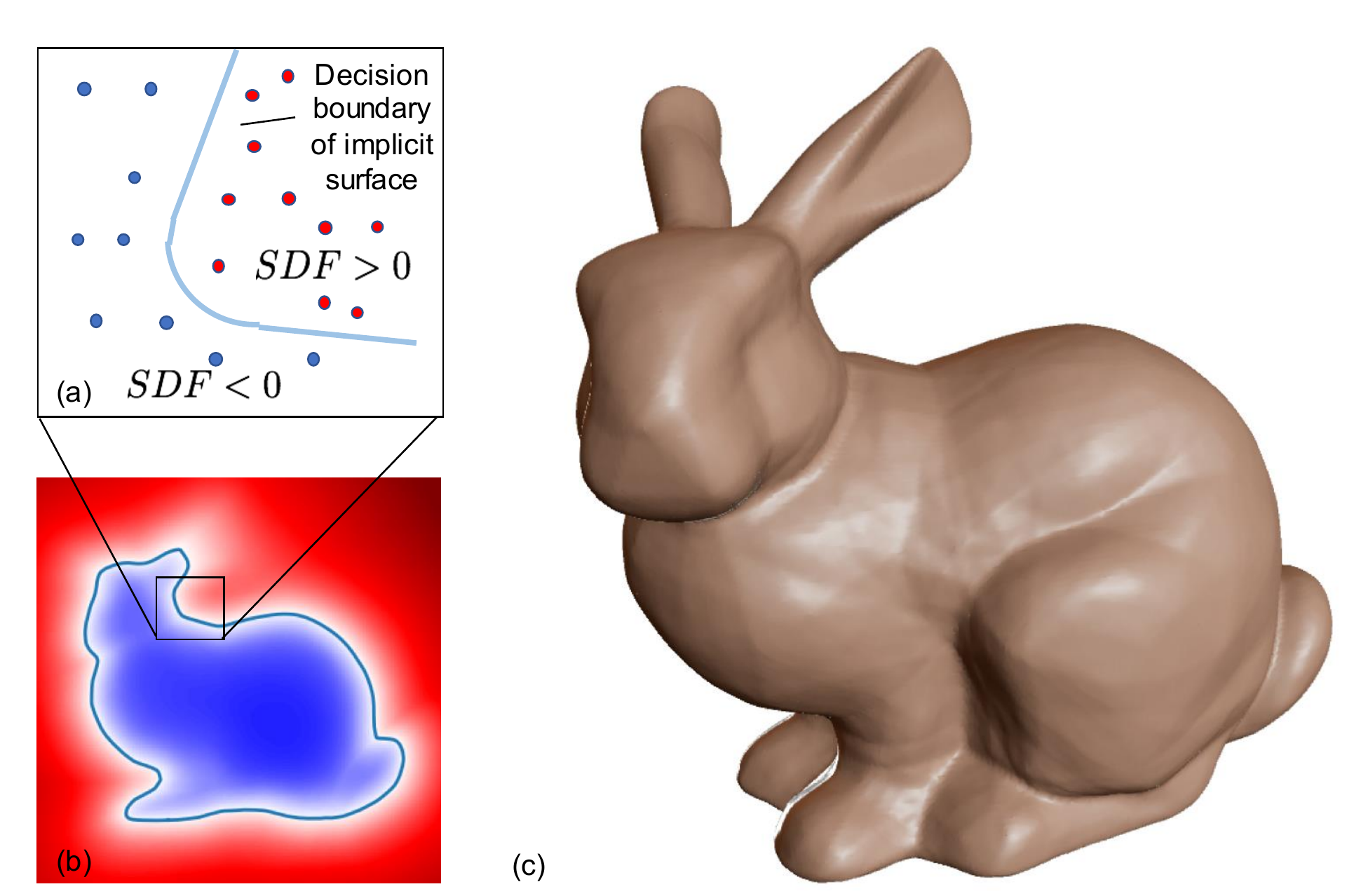}
\caption{Our DeepSDF representation applied to the Stanford Bunny: (a) depiction of the underlying implicit surface $SDF=0$ trained on sampled points inside $SDF<0$ and outside $SDF>0$ the surface, (b) 2D cross-section of the signed distance field, (c) rendered 3D surface recovered from $SDF=0$. Note that (b) and (c) are recovered via DeepSDF.}
\label{fig:deepsdfbunny}
\end{figure}

In this work, we present a novel representation and approach for generative 3D modeling that is efficient, expressive, and fully continuous. Our approach uses the concept of a SDF, but unlike common surface reconstruction techniques which discretize this SDF into a regular grid for evaluation and measurement denoising, we instead learn a generative model to produce such a continuous field.

The proposed continuous representation 
may be intuitively understood as a learned shape-conditioned classifier for which the decision boundary is the surface of the shape itself, as shown in Fig.~\ref{fig:deepsdfbunny}.   
Our approach shares the generative aspect of other works seeking to map a latent space to a distribution of complex shapes in 3D \cite{wu2016learning}, but critically differs in the central representation.  While the notion of an implicit surface defined as a SDF is widely known in the computer vision and graphics communities, to our knowledge no prior works have attempted to directly learn continuous, generalizable 3D generative models of SDFs.  

Our contributions include: (i) the formulation of generative shape-conditioned 3D modeling with a continuous implicit surface, (ii) a learning method for 3D shapes based on a probabilistic auto-decoder, and (iii) the demonstration and application of this formulation
to shape modeling and completion.  
Our models produce high quality continuous surfaces with complex topologies, and obtain state-of-the-art results in quantitative comparisons for shape reconstruction and completion.  As an example of the effectiveness of our method, our models use only 7.4 MB (megabytes) of memory to represent entire classes of shapes (for example, thousands of 3D chair models) -- this is, for example, less than half the memory footprint (16.8 MB) of a single uncompressed $512^3$ 3D bitmap.

\section{Related Work}

We review three main areas of related work: 3D representations for shape learning (Sec.~\ref{related-representations}), techniques for learning generative models (Sec.~\ref{related-generative}), and shape completion (Sec.~\ref{related-completion}).

\subsection{Representations for 3D Shape Learning}\label{related-representations}
Representations for data-driven 3D learning approaches can be largely classified into three categories: point-based, mesh-based, and voxel-based methods. 
While some applications such as 3D-point-cloud-based object classification are well suited to these representations, we address their limitations in expressing continuous surfaces with complex topologies.

\noindent\textbf{Point-based.} A point cloud 
is a lightweight 3D representation that closely matches the raw data that many sensors (i.e. LiDARs, depth cameras) provide, and hence is a natural fit for applying 3D learning. PointNet \cite{qi2017pointnet,qi2017pointnet++}, for example, uses max-pool operations to extract global shape features, and the technique is widely used as an encoder for point generation networks \cite{yang2017foldingnet,achlioptas2018learning}. There is a sizable list of related works to the PointNet style approach of learning on point clouds. 
A primary limitation, however, of learning with point clouds is that they do not describe topology and are not suitable for producing watertight surfaces. 

\noindent\textbf{Mesh-based.} Various approaches represent classes of similarly shaped objects, such as morphable human body parts, with predefined template meshes and some of these models demonstrate high fidelity shape generation results  \cite{bagautdinov2018modeling, litany2017deformable}. Other recent works \cite{baque2018geodesic} use poly-cube mapping \cite{tarini2004polycube} for shape optimization. While the use of template meshes is convenient and naturally provides 3D correspondences, it can only model shapes with fixed mesh topology. 

Other mesh-based methods use existing \cite{sinha2016deep,maron2017convolutional} or learned \cite{groueix2018atlasnet,hamu2018multi} parameterization techniques to describe 3D surfaces by morphing 2D planes. The quality of such representations depends on parameterization algorithms that are often sensitive to input mesh quality and cutting strategies. To address this, recent data-driven approaches \cite{yang2017foldingnet, groueix2018atlasnet} learn the parameterization task with deep networks. They report, however, that (a) multiple planes are required to describe complex topologies but (b) the generated surface patches are not stitched, i.e. the produced shape is not closed. To generate a closed mesh, sphere parameterization may be used \cite{groueix2018atlasnet,hamu2018multi}, but the resulting shape is limited to the topological sphere. Other works related to learning on meshes propose to use new convolution and pooling operations for meshes \cite{defferrard2016convolutional, verma2018feastnet} or general graphs \cite{bruna2013spectral}. 

\noindent\textbf{Voxel-based.} Voxels, which non-parametrically describe volumes with 3D grids of values, are perhaps the most natural extension into the 3D domain of the well-known learning paradigms (i.e., convolution) that have excelled in the 2D image domain.  
The most straightforward variant of voxel-based learning is to use a dense occupancy grid (occupied / not occupied). 
Due to the cubically growing compute and memory requirements, however, current methods are only able to handle low resolutions ($128^3$ or below). As such, voxel-based approaches do not preserve fine shape details \cite{wu20153d, choy20163d}, and additionally voxels visually appear significantly different than high-fidelity shapes, since when rendered their normals are not smooth. Octree-based methods \cite{ogn2017,riegler2017octnet,hane2017hierarchical} alleviate the compute and memory limitations of dense voxel methods, extending for example the ability to learn at up to $512^3$ resolution \cite{ogn2017}, but even this resolution is far from producing shapes that are visually compelling. 

Aside from occupancy grids, and more closely related to our approach, it is also possible to use a 3D grid of voxels to represent a signed distance function.  This inherits from the success of fusion approaches that utilize a truncated SDF (TSDF), pioneered in \cite{curless1996volumetric, newcombe2011kinectfusion}, to combine noisy depth maps into a single 3D model.  Voxel-based SDF representations have been extensively used for 3D shape learning \cite{zeng20173dmatch,dai2017shape, stutz2018learning}, but their use of discrete voxels is expensive in memory. As a result, the learned discrete SDF approaches generally present low resolution shapes. \cite{Distance_Field_Compression} reports various wavelet transform-based approaches for distance field compression, while \cite{canelhas2016eigenshapes} applies dimensionality reduction techniques on discrete TSDF volumes. These methods encode the SDF volume of each individual scene rather than a dataset of shapes.

\subsection{Representation Learning Techniques}\label{related-generative}
Modern representation learning techniques aim at automatically discovering a set of features that compactly but expressively describe data. For a more extensive review of the field, we refer to Bengio et al. \cite{bengio2013representation}.

\noindent\textbf{Generative Adversial Networks.} GANs \cite{goodfellow2014generative} and their variants \cite{chen2016infogan,radford2015unsupervised} learn deep embeddings of target data by training discriminators adversarially against generators. Applications of this class of networks \cite{isola2017image,karras2017progressive} generate realstic images of humans, objects, or scenes. On the downside, adversarial training for GANs is known to be unstable.
In the 3D domain, Wu~et~al.~\cite{wu2016learning} trains a GAN to generate objects in a voxel representation, while the recent work of Hamu~et~al.~\cite{hamu2018multi} uses multiple parameterization planes to generate shapes of topological spheres.

\noindent\textbf{Auto-encoders.}
Auto-encoder outputs are expected to replicate the original input given the constraint of an information bottleneck between the encoder and decoder. The ability of auto-encoders as a feature learning tool has been evidenced by the vast variety of 3D shape learning works in the literature \cite{dai2017shape, stutz2018learning,bagautdinov2018modeling,groueix2018atlasnet,wu2018learning} who adopt auto-encoders for representation learning. Recent 3D vision works \cite{bloesch2018codeslam, bagautdinov2018modeling, litany2017deformable} often adopt a variational auto-encoder (VAE) learning scheme, in which bottleneck features are perturbed with Gaussian noise to encourage smooth and complete latent spaces. The regularization on the latent vectors enables exploring the embedding space with gradient descent or random sampling.

\noindent\textbf{Optimizing Latent Vectors.}
Instead of using the full auto-encoder for representation learning, an alternative is to learn compact data representations by training \textit{decoder-only} networks. This idea goes back to at least the work of Tan~et~al.~\cite{tan1995reducing} which simultaneously optimizes the latent vectors assigned to each data point and the decoder weights through back-propagation. For inference, an optimal latent vector is searched to match the new observation with fixed decoder parameters. Similar approaches have been extensively studied in \cite{reddy1998input, bouakkaz2012combined, qunxiong2006dimensionality}, for applications including noise reduction, missing measurement completions, and fault detections. Recent approaches \cite{bojanowski18a, fan2018matrix} extend the technique by applying deep architectures. Throughout the paper we refer to this class of networks as \textit{auto-decoders}, for they are trained with {\em self}-reconstruction loss on decoder-only architectures.

\subsection{Shape Completion}\label{related-completion}

3D shape completion related works aim to infer unseen parts of the original shape given sparse or partial input observations. This task is anaologous to image-inpainting in 2D computer vision.

Classical surface reconstruction methods complete a point cloud into a dense surface by fitting radial basis function (RBF) \cite{carr2001reconstruction} to approximate implicit surface functions, or by casting the reconstruction from oriented point clouds as a Poisson problem \cite{kazhdan2013screened}. These methods only model a single shape rather than a dataset.

Various recent methods use data-driven approaches for the 3D completion task. Most of these methods adopt encoder-decoder architectures to reduce partial inputs of occupancy voxels \cite{wu20153d}, discrete SDF voxels \cite{dai2017shape}, depth maps \cite{rock2015completing}, RGB images \cite{choy20163d, wu2018learning} or point clouds \cite{stutz2018learning} into a latent vector and subsequently generate a prediction of full volumetric shape based on learned priors.

\section{Modeling SDFs with Neural Networks} \label{sec: csdf}

In this section we present DeepSDF, our continuous shape learning approach. We describe modeling shapes as the zero iso-surface decision boundaries of feed-forward networks trained to represent SDFs.
A signed distance function is a continuous function that, for a given spatial point, outputs the point's distance to the closest surface, whose sign encodes whether the point is inside (negative) or outside (positive) of the watertight surface:
\begin{equation} \label{eq:sdf}
   SDF(\bm{x}) = s: \bm{x}\in \mathbb{R}^3,\, s\in \mathbb{R}\,.
\end{equation} 
The underlying surface is implicitly represented by the iso-surface of $SDF(\cdot)=0$. A view of this implicit surface can be rendered through raycasting or rasterization of a mesh obtained with, for example, Marching Cubes \cite{lorensen1987marching}. 

Our key idea is to directly regress the continuous SDF from point samples using deep neural networks. The resulting trained network is able to predict the SDF value of a given query position, from which we can extract the zero level-set surface by evaluating spatial samples. Such surface representation can be intuitively understood as a learned binary classifier for which the decision boundary is the surface of the shape itself as depicted in Fig.~\ref{fig:deepsdfbunny}. As a universal function approximator \cite{hornik1989multilayer}, deep feed-forward networks in theory can learn the fully continuous shape functions with arbitrary precision. Yet, the precision of the approximation in practice is limited by the finite number of point samples that guide the decision boundaries and the finite capacity of the network due to restricted compute power.

\begin{figure} 
\begin{subfigure}[t]{0.44\linewidth}
\includegraphics[width=\linewidth]{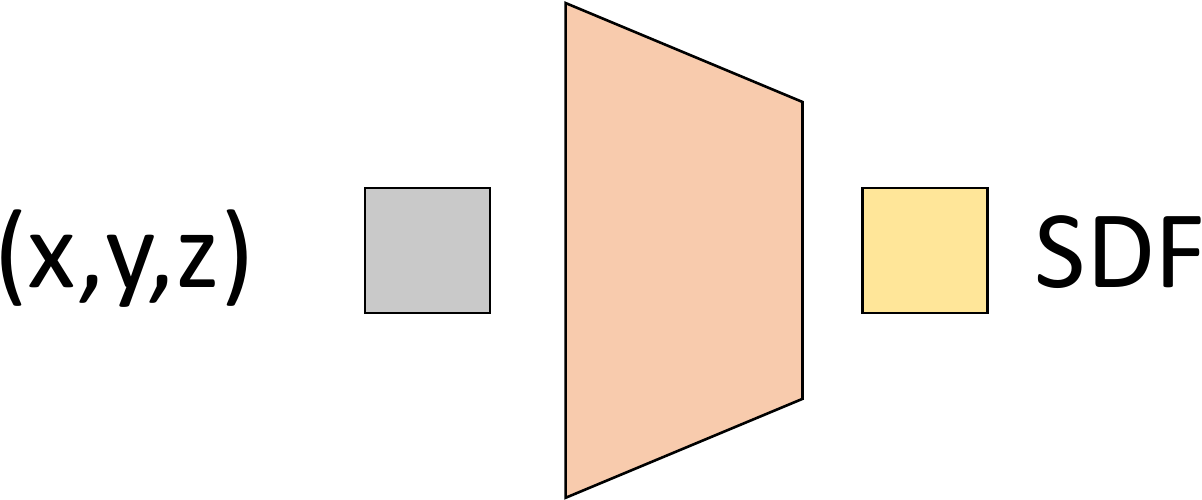}
\caption{Single Shape DeepSDF \label{fig:singleShapeDeepSDF}}
\end{subfigure}
\hfill
\begin{subfigure}[t]{0.44\linewidth}
\includegraphics[width=\linewidth]{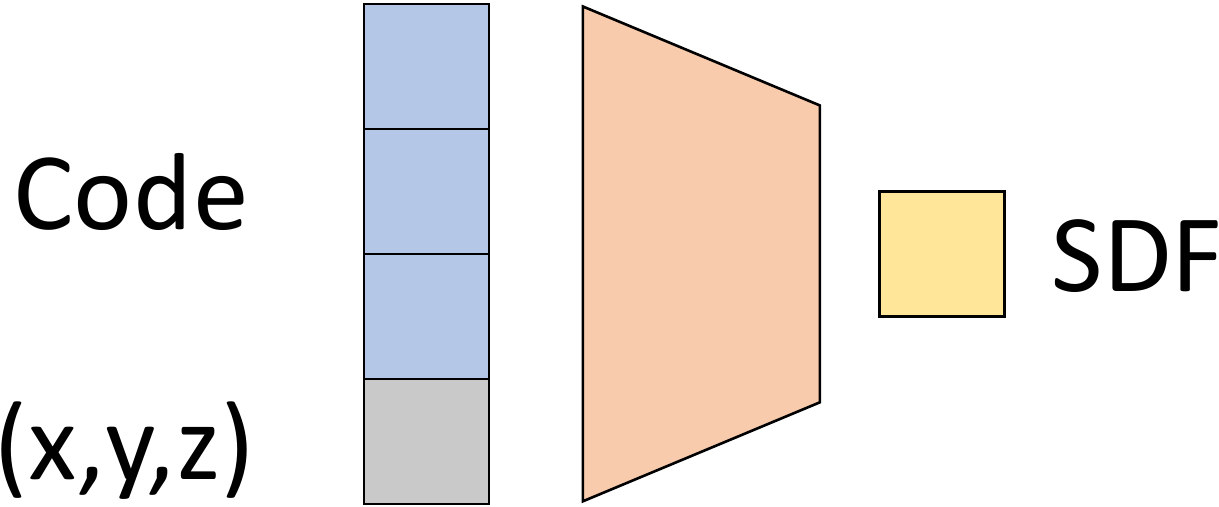}
\caption{Coded Shape DeepSDF \label{fig:codedShapeDeepSDF}}
\end{subfigure}
	\caption{In the single-shape DeepSDF instantiation, the shape information is contained in the network itself whereas the coded-shape DeepSDF, the shape information is contained in a code vector that is concatenated with the 3D sample location. In both cases, DeepSDF produces the SDF value at the 3D query location,
	}
	\label{fig: latentmodel}
\end{figure}

The most direct application of this approach is to train a single deep network for a given target shape as depicted in Fig.~\ref{fig:singleShapeDeepSDF}.
Given a target shape, we prepare a set of pairs $X$ composed of 3D point samples and their SDF values:
\begin{equation} \label{eq:sdfdef} 
    X:=\{ (\bm{x}, s) : SDF(\bm{x})=s\}\,.
\end{equation}
We train the parameters $\theta$ of a multi-layer fully-connected neural network $f_\theta$ on the training set $S$ to make $f_\theta$ a good approximator of the given SDF in the target domain $\Omega$:
\begin{equation} \label{eq:f}
  f_\theta(\bm{x})\approx SDF(\bm{x}), \,\forall \bm{x}\in \Omega\,.
\end{equation}

The training is done by minimizing the sum over losses between the predicted and real SDF values of points in $X$ under the following $L_1$ loss function:
\begin{equation}\label{eq:clampedL1}
\mathcal{L}(f_\theta(\bm{x}),s) = | \clamp(f_\theta(\bm{x}),\delta) -  \clamp(s,\delta) \ |,
\end{equation}
where $\clamp(x,\delta) := \min(\delta, \max(-\delta, x))$ introduces the parameter $\delta$ to control the distance from the surface over which we expect to maintain a metric SDF. Larger values of $\delta$ allow for fast ray-tracing since each sample gives information of safe step sizes. Smaller values of $\delta$ can be used to concentrate network capacity on details near the surface. 

To generate the 3D model shown in Fig.~\ref{fig:singleShapeDeepSDF}, we use $\delta=0.1$ and a feed-forward network composed of eight fully connected layers, each of them applied with dropouts. All internal layers are 512-dimensional and have ReLU non-linearities. The output non-linearity regressing the SDF value is tanh.
We found training with batch-normalization \cite{ioffe2015batch} to be unstable and applied the weight-normalization technique instead \cite{salimans2016weight}.
For training, we use the Adam optimizer~\cite{kingma2014adam}.
Once trained, the surface is implicitly represented as the zero iso-surface of $f_\theta(\bm{x})$, which can be visualized through raycasting or marching cubes. Another nice property of this approach is that accurate normals can be analytically computed by calculating the spatial derivative $\partial f_\theta(\bm{x}) / \partial \bm{x}$ via back-propogation through the network.

\section{Learning the Latent Space of Shapes}

Training a specific neural network for each shape is neither feasible nor very useful. Instead, we want a model that can represent a wide variety of shapes, discover their common properties, and embed them in a low dimensional latent space. 
To this end, we introduce a latent vector $\bm{z}$, which can be thought of as encoding the desired shape, as a second input to the neural network as depicted in Fig.~\ref{fig:codedShapeDeepSDF}. 
Conceptually, we map this latent vector to a 3D shape represented by a continuous SDF. Formally, for some shape indexed by $i$,  $f_\theta$ is now a function of a latent code $\bm{z}_i$ and a query 3D location $\bm{x}$, and outputs the shape's approximate SDF:
\begin{equation} 
\label{eq:latent}
f_\theta(\bm{z}_i,\bm{x})\approx SDF^i(\bm{x}).
\end{equation}
By conditioning the network output on a latent vector, this formulation allows modeling multiple SDFs with a single neural network.
Given the decoding model $f_{\theta}$, the continuous surface associated with a latent vector $\bm{z}$ is similarly represented with the decision boundary of $f_{\theta}(\bm{z},\bm{x})$, and the shape can again be discretized for visualization by, for example, raycasting or Marching Cubes.

Next, we motivate the use of encoder-less training before introducing the `auto-decoder' formulation of the shape-coded DeepSDF.

\begin{figure} 
\centering
\begin{subfigure}[t]{0.44\linewidth}
\includegraphics[width=\linewidth]{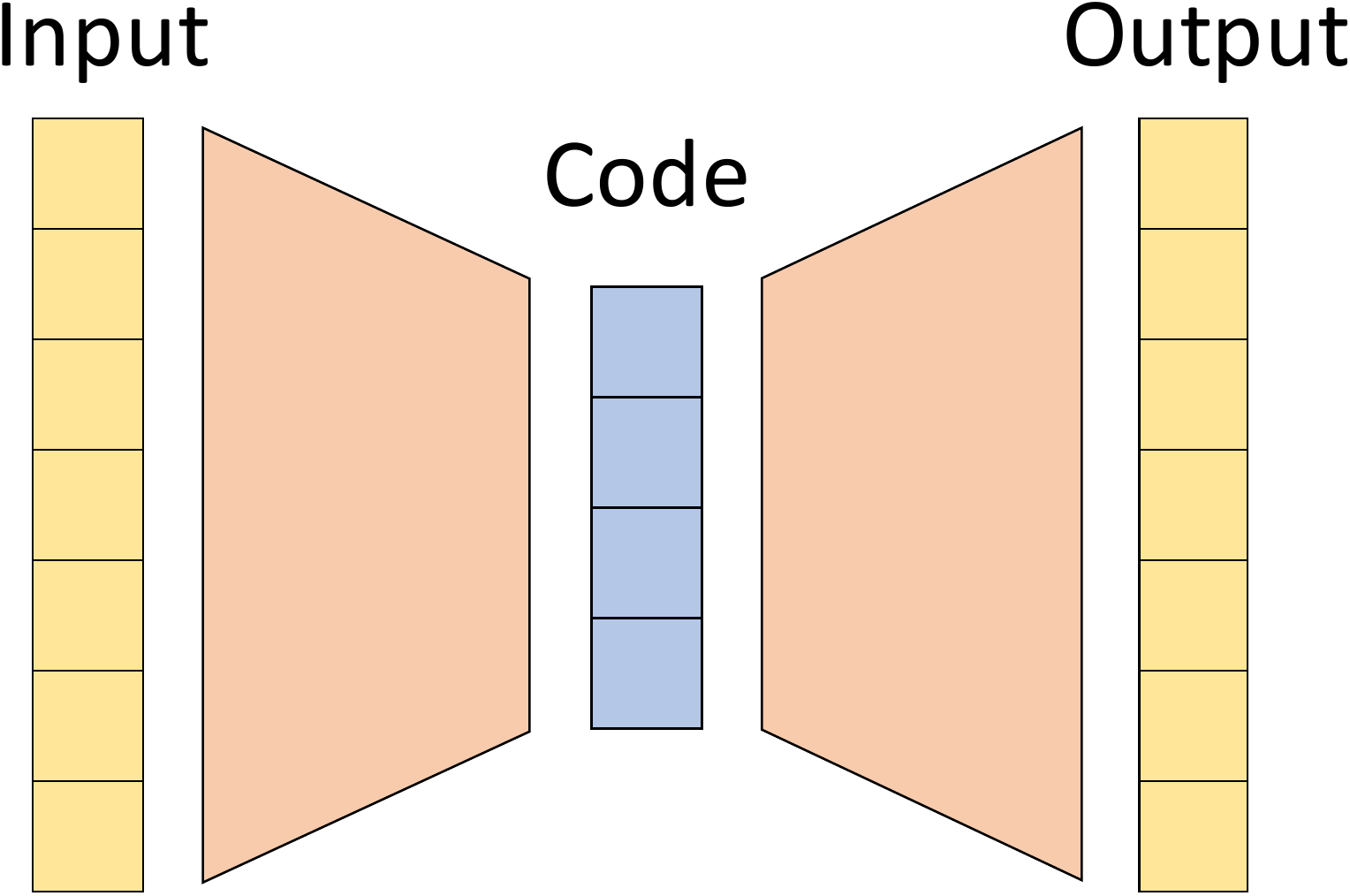}
\caption{Auto-encoder}
\end{subfigure}
\begin{subfigure}[t]{0.44\linewidth}
\includegraphics[width=\linewidth]{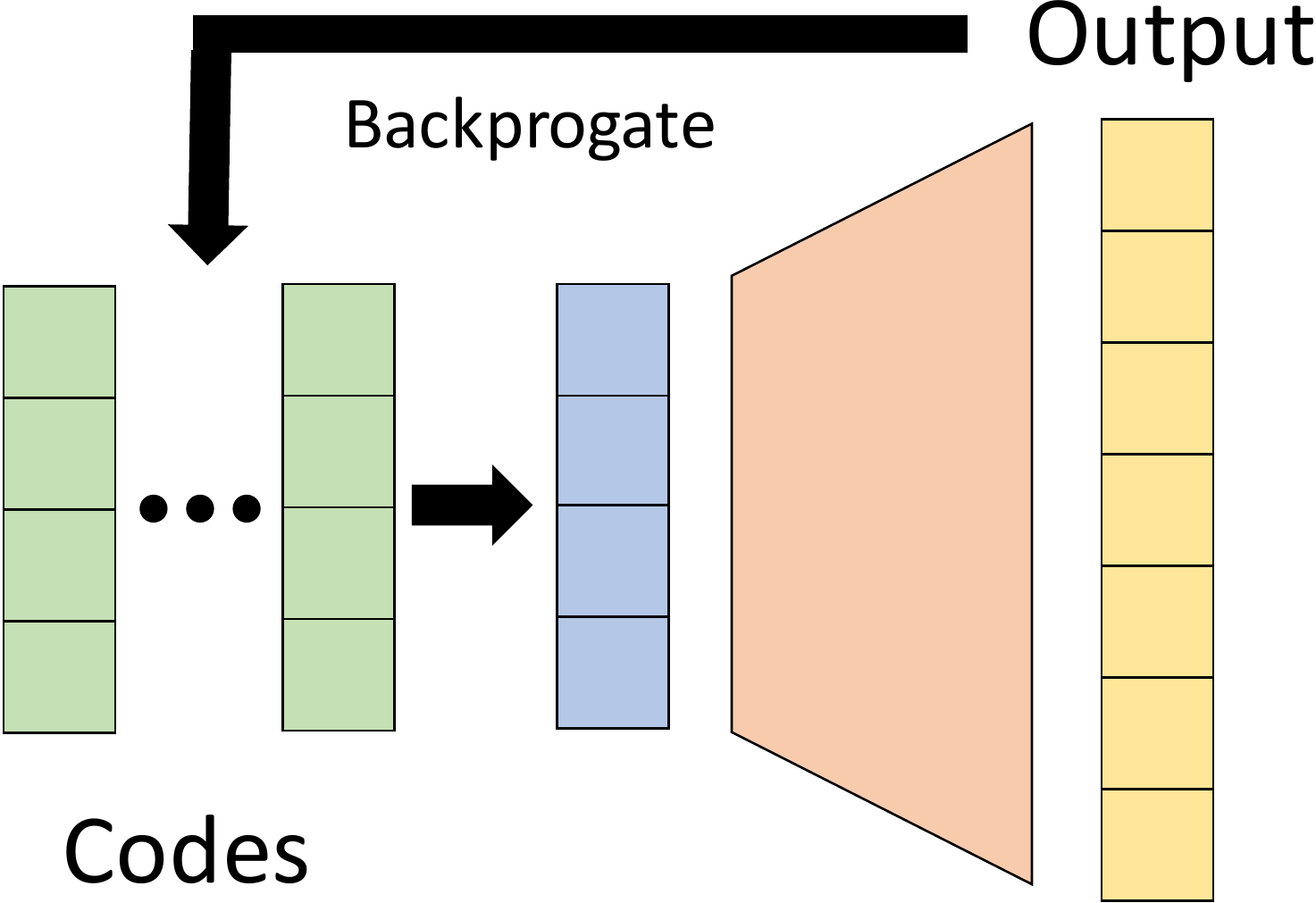}
\caption{Auto-decoder}
\end{subfigure}%
	\caption{Different from an auto-encoder whose latent code is produced by the encoder, an auto-decoder directly accepts a latent vector as an input. A randomly initialized latent vector is assigned to each data point in the beginning of training, and the latent vectors are optimized along with the decoder weights through standard backpropagation. During inference, decoder weights are fixed, and an optimal latent vector is estimated.}
	\label{fig: autodecoder}
\end{figure}

\subsection{Motivating Encoder-less Learning}

Auto-encoders and encoder-decoder networks are widely used for representation learning as their bottleneck features tend to form natural latent variable representations. 

Recently, in applications such as modeling depth maps \cite{bloesch2018codeslam}, faces \cite{bagautdinov2018modeling}, and body shapes \cite{litany2017deformable} a full auto-encoder is trained but only the decoder is retained for inference, where they search for an optimal latent vector given some input observation.
However, since the trained encoder is unused at test time, it is unclear whether using the encoder is the most effective use of computational resources during training.
This motivates us to use an \textit{auto-}\textbf{decoder} for learning a shape embedding without an encoder as depicted in Fig.~\ref{fig: autodecoder}. 

We show that applying an auto-decoder to learn continuous SDFs leads to high quality 3D generative models. Further, we develop a probabilistic formulation for training and testing the auto-decoder that naturally introduces latent space regularization for improved generalization. To the best of our knowledge, this work is the first to introduce the auto-decoder learning method to the 3D learning community.

\subsection{Auto-decoder-based DeepSDF Formulation} \label{sec:autodecoder}
To derive the auto-decoder-based shape-coded DeepSDF formulation we adopt a probabilistic perspective. 
Given a dataset of $N$ shapes represented with signed distance function ${SDF^i}_{i=1}^{N}$, we prepare a set of $K$ point samples and their signed distance values:
\begin{equation} \label{eq:datapoint}
X_i = \{(\bm{x}_j,s_j):s_j=SDF^i(\bm{x}_j)\}\,.
\end{equation}

For an auto-decoder, as there is no encoder, each latent code $\bm{z}_i$ is paired with training shape $X_i$. 
The posterior over shape code $\bm{z}_i$ given the shape SDF samples $X_i$ can be decomposed as:
\begin{equation}
p_{\theta}(\bm{z}_i|X_i)=
p(\bm{z}_i) \textstyle \prod_{(\bm{x}_j,\bm{s}_j)\in X_i}p_{\theta}(\bm{s}_j | z_i; \bm{x}_j) \,,
\end{equation}
where $\theta$ parameterizes the SDF likelihood. In the latent shape-code space, we assume the prior distribution over codes $p(\bm{z_i})$ to be a zero-mean multivariate-Gaussian with a spherical covariance $\sigma^2 I$. This prior encapsulates the notion that the shape codes should be concentrated and we empirically found it was needed to infer a compact shape manifold and
to help converge to good solutions.

In the auto-decoder-based DeepSDF formulation we express the SDF likelihood via a deep feed-forward network $f_\theta(\bm{z}_i,\bm{x}_j)$ and, without loss of generality, assume that the likelihood takes the form:
\begin{equation}
   p_{\theta}(\bm{s}_j | z_i; \bm{x}_j) = \exp(-\mathcal{L}(f_\theta (\bm{z}_i,\bm{x}_j),s_j)) \,.
\end{equation}
The SDF prediction $\tilde{s}_j = f_\theta (\bm{z}_i,\bm{x}_j)$ is represented using a fully-connected network. $\mathcal{L}(\tilde{s}_j, s_j)$ is a loss function penalizing the deviation of the network prediction from the actual SDF value $s_j$. One example for the cost function is the standard $L_2$ loss function which amounts to assuming Gaussian noise on the SDF values. In practice we use the clamped $L_1$ cost from Eq.~\ref{eq:clampedL1} for reasons outlined previously.

At training time we maximize the joint log posterior over all training shapes with respect to the individual shape codes $\{z_i\}_{i=1}^N$ and the network parameters $\theta$:
\begin{equation} \label{eq:objective}
\argmin_{\theta, \{\bm{z}_i\}_{i=1}^N} \sum_{i=1}^N \left( \sum_{j=1}^K \mathcal{L}(f_\theta (\bm{z}_i,\bm{x}_j),s_j)+\frac{1}{\sigma^{2}}||\bm{z}_i ||_2^2 \right).
\end{equation}

At inference time, after training and fixing $\theta$, a shape code $\bm{z}_i$ for shape $X_i$ can be estimated via Maximum-a-Posterior (MAP) estimation as:
 \begin{equation} \label{eq:testtime}
\hat{\bm{z}}= \argmin_{\bm{z}} \sum_{(\bm{x}_j,\bm{s}_j)\in X} \mathcal{L}(f_\theta (\bm{z},\bm{x}_j),s_j)+\frac{1}{\sigma^{2}}||\bm{z}||_2^2 \,.
\end{equation}

Crucially, this formulation is valid for SDF samples $X$ of arbitrary size and distribution because the gradient of the loss with respect to $\bm{z}$ can be computed separately for each SDF sample. This implies that DeepSDF can handle any form of partial observations such as depth maps. 
This is a major advantage over the auto-encoder framework whose encoder expects a test input similar to the training data, e.g. shape completion networks of \cite{dai2017shape,yuan2018pcn} require preparing training data of partial shapes.

To incorporate the latent shape code, we stack the code vector and the sample location as depicted in Fig.~\ref{fig:codedShapeDeepSDF} and feed it into the same fully-connected NN described previously at the input layer and additionally at the 4th layer.
We again use the Adam optimizer~\cite{kingma2014adam}.
The latent vector $\bm{z}$ is initialized randomly from $\mathcal{N}(0,0.01^2)$.

Note that while both VAE and the proposed auto-decoder formulation share the zero-mean Gaussian prior on the latent codes, we found that the the stochastic nature of the VAE optimization did not lead to good training results. 

\begin{figure} 
\begin{subfigure}[t]{0.495\linewidth}
\includegraphics[width=0.9\linewidth]{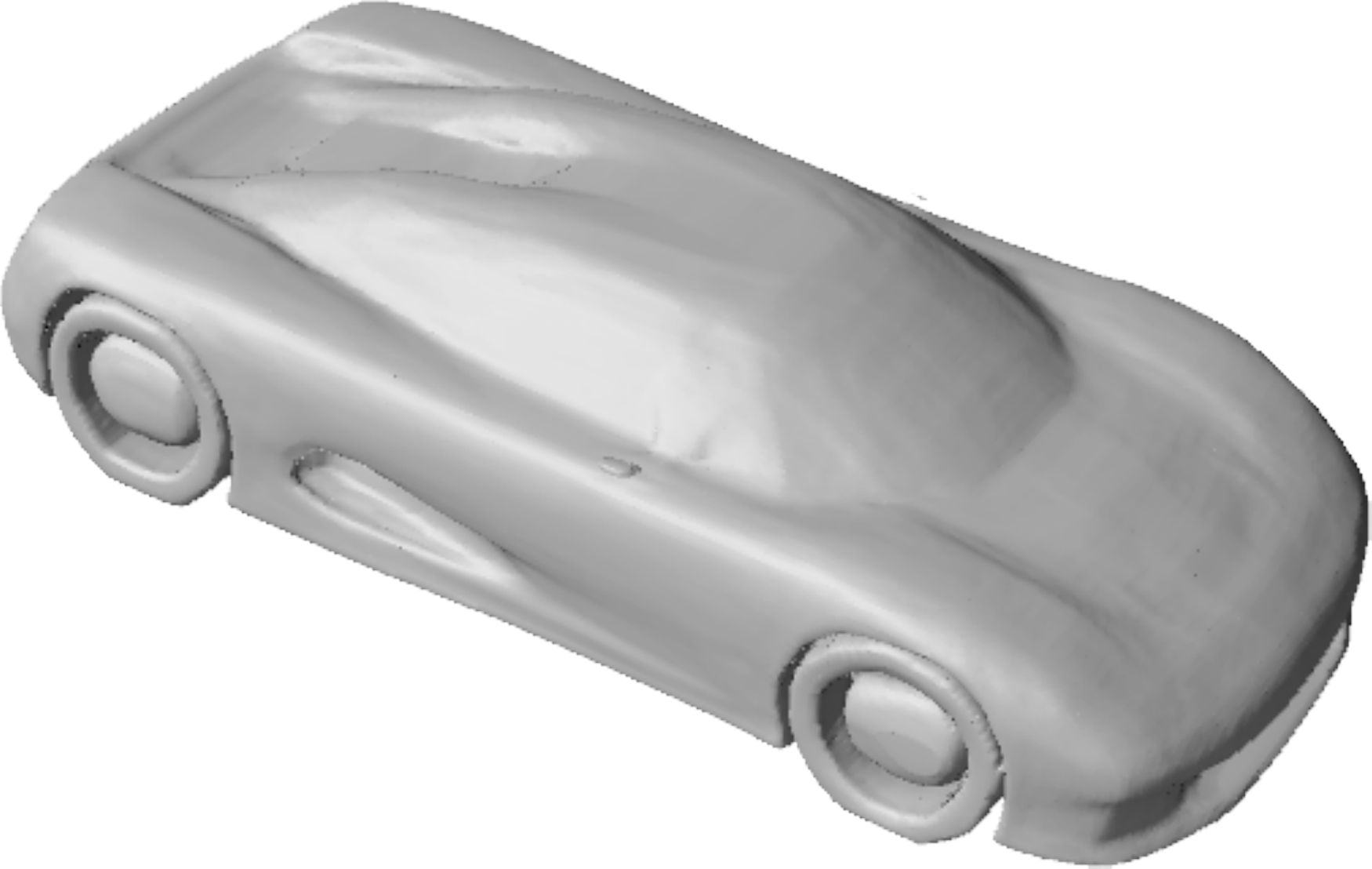}
\end{subfigure}
\hfill
\begin{subfigure}[t]{0.495\linewidth}
\includegraphics[width=0.9\linewidth]{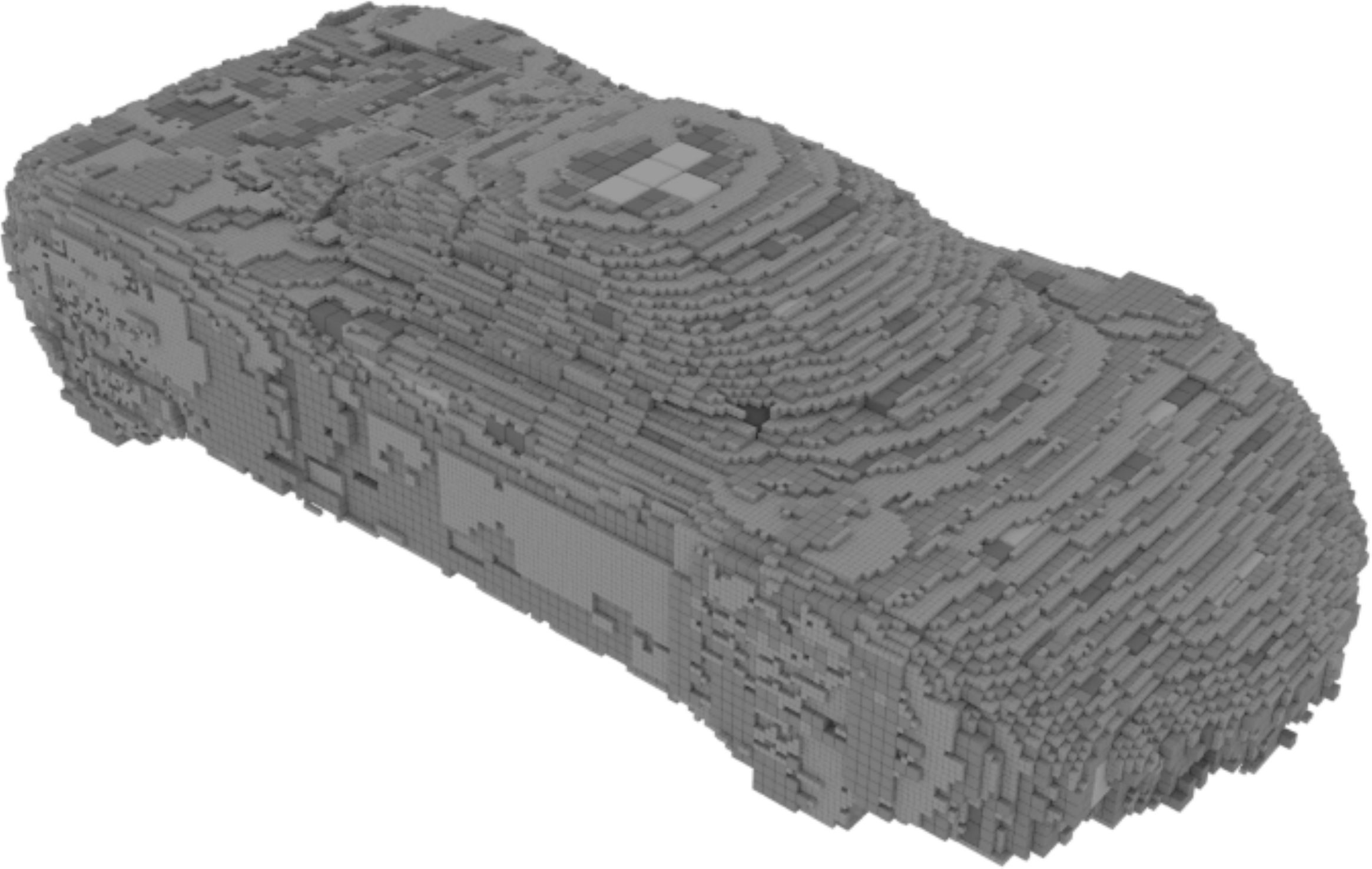}
\end{subfigure}%
\caption{Compared to car shapes memorized using OGN~\cite{ogn2017} (right), our models (left) preserve details and render visually pleasing results as DeepSDF provides oriented surace normals.}
	\label{fig: ogn-compare}
\end{figure}
\begin{table*}[t]
  \centering
  \footnotesize
  \begin{tabular}{|l|c|c|c|c|c|c|c|c|}
\hline
              &          &    & Complex & Closed & Surface & Model & Inf. & Eval. \\
Method  & Type & Discretization     & topologies & surfaces & normals & size (GB)  (s) & time (s) & tasks \\
 \hhline{|=|=|=|=|=|=|=|=|=|}
3D-EPN \cite{dai2017shape}  & Voxel SDF & $32^3$ \text{ voxels} & \checkmark & \checkmark  & \checkmark  & 0.42   &- & C\\ \hline
OGN \cite{ogn2017}               & Octree & $256^3$ \text{ voxels} & \checkmark & \checkmark &  & 0.54   & 0.32  & K \\ \hline
AtlasNet  &  Parametric & 1 patch &  & \checkmark &  & 0.015  & \textbf{0.01} & K, U  \\ 
-Sphere  \cite{groueix2018atlasnet}                                                                    & mesh           &             &   &                    &   &            &  &         \\   \hline
AtlasNet  & Parametric  & 25 patches & \checkmark & &  & 0.172   & 0.32 &  K, U \\ 
-25  \cite{groueix2018atlasnet}                                                              & mesh           &             &   &                    &   &            &   & \\       \hline
DeepSDF  & Continuous  & \textbf{none} & \checkmark & \checkmark & \checkmark & \textbf{0.0074}  & 9.72 & K, U, C \\ 
(ours)                                                                      & SDF           &             &   &                    &   &            &      &  \\     \hline
  \end{tabular}
  \caption{Overview of the benchmarked methods. 
  AtlasNet-Sphere can only describe topological-spheres, voxel/octree occupancy methods (i.e. OGN) only provide 8 directions for normals, and AtlasNet does not provide oriented normals. Our tasks evaluated are: (K) representing known shapes, (U) representing unknown shapes, and (C) shape completion.} 
  \label{tab:1}
\end{table*}

\section{Data Preparation} \label{sec: data}
To train our continuous SDF model, we prepare the SDF samples $X$ (Eq.~\ref{eq:sdfdef}) for each mesh, which consists of 3D points and their SDF values. While SDF can be computed through a distance transform for any watertight shapes  from real or synthetic data, we train with synthetic objects, (e.g. ShapeNet \cite{chang2015shapenet}), for which we are provided complete 3D shape meshes. To prepare data, we start by normalizing each mesh to a unit sphere and sampling 500,000 spatial points $\bm{x}$'s: we sample more aggressively near the surface of the object as we want to capture a more detailed SDF near the surface. 
For an ideal oriented watertight mesh, computing the signed distance value of $\bm{x}$ would only involve finding the closest triangle, but we find that human designed meshes are commonly not watertight and contain undesired internal structures. To obtain the {\em shell} of a mesh with proper orientation, we set up equally spaced virtual cameras around the object, and densely sample surface points, denoted $P_s$, with surface normals oriented towards the  camera. Double sided triangles visible from both orientations (indicating that the shape is not closed) cause problems in this case, so we discard mesh objects containing too many of such faces. Then, for each $\bm{x}$, we find the closest point in $P_s$, from which the $SDF(\bm{x})$ can be computed. We refer  readers to supplementary material for further details.

\section{Results} 
We conduct a number of experiments to show the representational power of DeepSDF, both in terms of its ability to describe geometric details and its generalization capability to learn a desirable shape embedding space. Largely, we propose four main experiments designed to test its ability to 1)  represent training data, 2) use learned feature representation to reconstruct unseen shapes, 3) apply shape priors to complete partial shapes, and 4) learn smooth and complete shape embedding space from which we can sample new shapes. For all experiments we use the popular ShapeNet \cite{chang2015shapenet} dataset.

We select a representative set of 3D learning approaches to comparatively evaluate aforementioned criteria: a recent octree-based method (OGN) \cite{ogn2017}, a mesh-based method (AtlasNet) \cite{groueix2018atlasnet}, and a volumetric SDF-based shape completion method (3D-EPN) \cite{dai2017shape} (Table \ref{tab:1}). These works show state-of-the-art performance in their respective representations and tasks, so we omit comparisons with the  works that have already been compared: e.g. OGN's octree model outperforms regular voxel approaches, while AtlasNet compares itself with various points, mesh, or voxel based methods and 3D-EPN with various completion methods.

\begin{table}[t]
\centering
\footnotesize
 \begin{tabular}{|l|c|c|c|c|}
 \hline
                                                  &   CD,    & CD,      & EMD,  & EMD, \\
  Method \textbackslash metric &   mean & median &  mean & median\\
 \hhline{|=|=|=|=|=|}
 OGN            & 0.167  & 0.127 & \bf{0.043} & \bf{0.042} \\
 AtlasNet-Sph.  &  0.210 & 0.185 & 0.046 & 0.045  \\
 AtlasNet-25    & 0.157 & 0.140 & 0.060 & 0.060   \\
 DeepSDF        & \bf{0.084} & \bf{0.058} & \bf{0.043} & \bf{0.042}  \\
 \hline
   \end{tabular}
   \caption{Comparison for representing known shapes (K) for cars trained on ShapeNet. CD = Chamfer Distance ($30,000$ points) multiplied by $10^3$, EMD = Earth Mover's Distance ($500$ points).} 
   \label{tab:2}
\end{table}

\subsection{Representing Known 3D  Shapes} \label{sec:representing}

First, we evaluate the capacity of the model to represent \textit{known} shapes, i.e.~shapes that were in the training set, from only a restricted-size latent code --- testing the limit of expressive capability of the representations. 

Quantitative comparison in Table~\ref{tab:2} shows that the proposed DeepSDF significantly beats OGN and AtlasNet in Chamfer distance against the true shape computed with a large number of points (30,000). The difference in earth mover distance (EMD) is smaller because 500 points do not well capture the additional precision. Figure~\ref{fig: ogn-compare} shows a qualitative comparison of DeepSDF to OGN.

\subsection{Representing Test 3D Shapes (auto-encoding)}

\begin{figure*} 
\begin{subfigure}[t]{0.13\linewidth}
\includegraphics[width=0.9\linewidth]{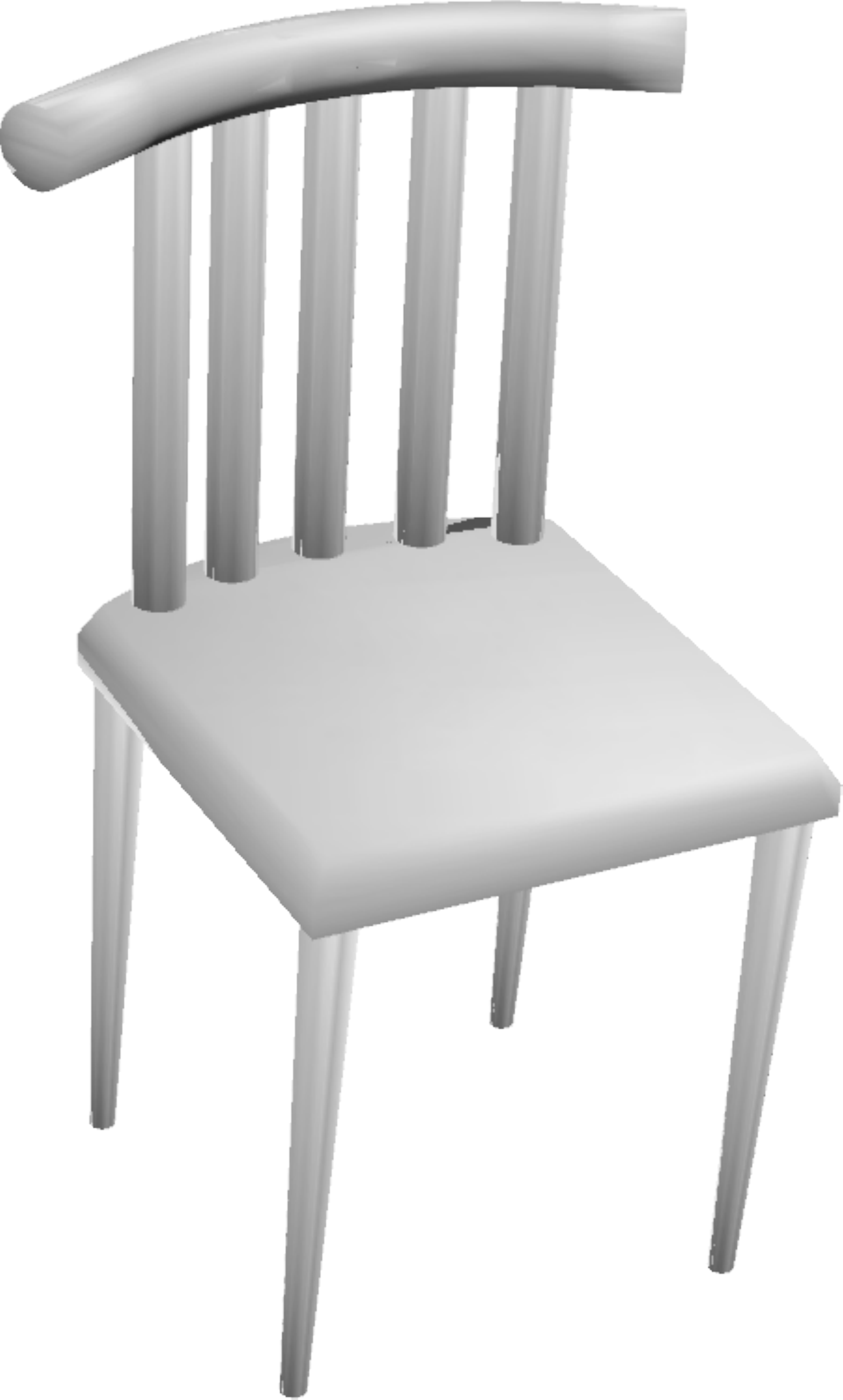}
\caption{Ground-truth}
\end{subfigure}
\hfill
\begin{subfigure}[t]{0.13\linewidth}
\includegraphics[width=0.9\linewidth]{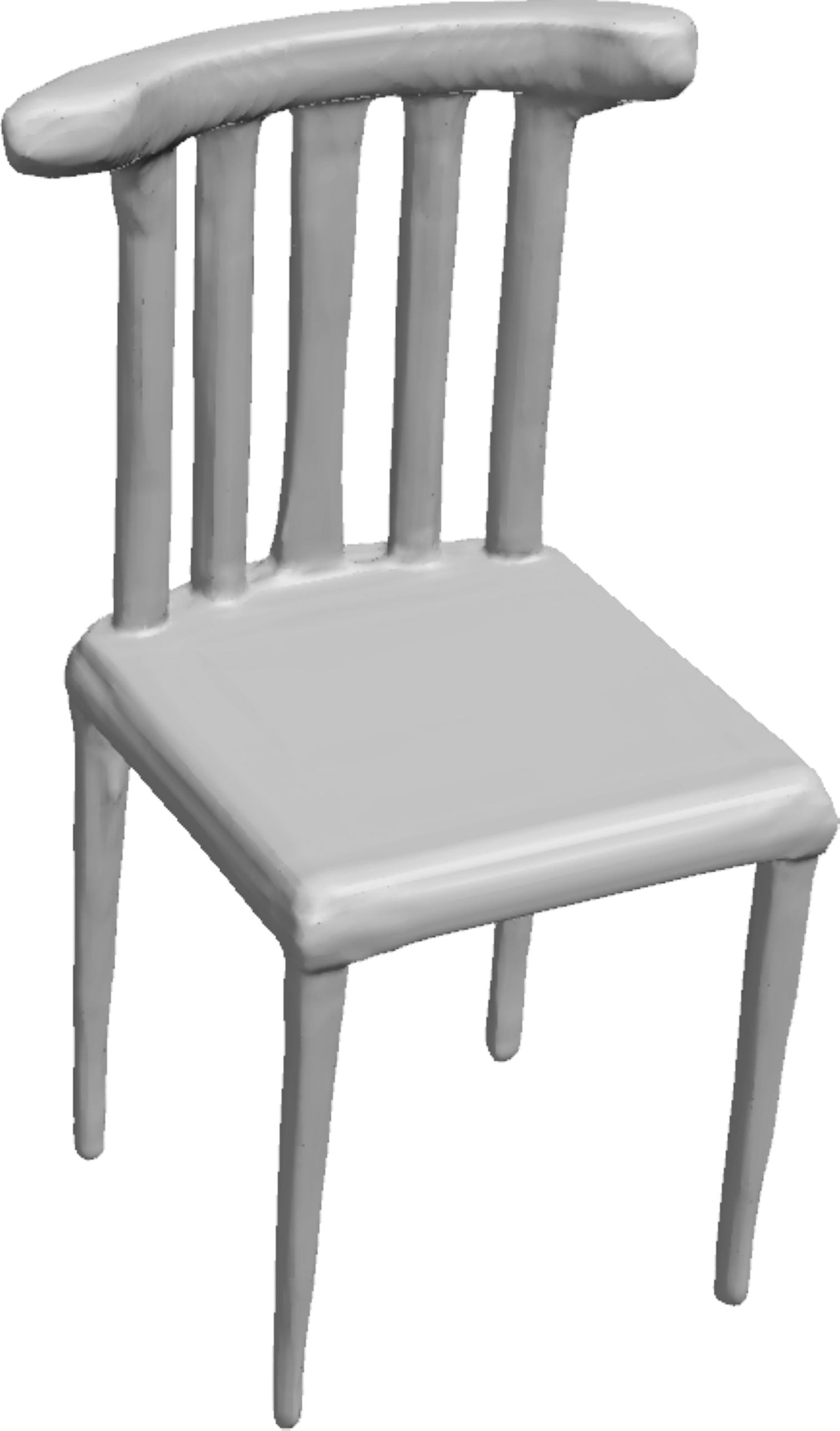}
\caption{Our Result}
\end{subfigure}
\hfill
\begin{subfigure}[t]{0.135\linewidth}
\includegraphics[width=0.9\linewidth]{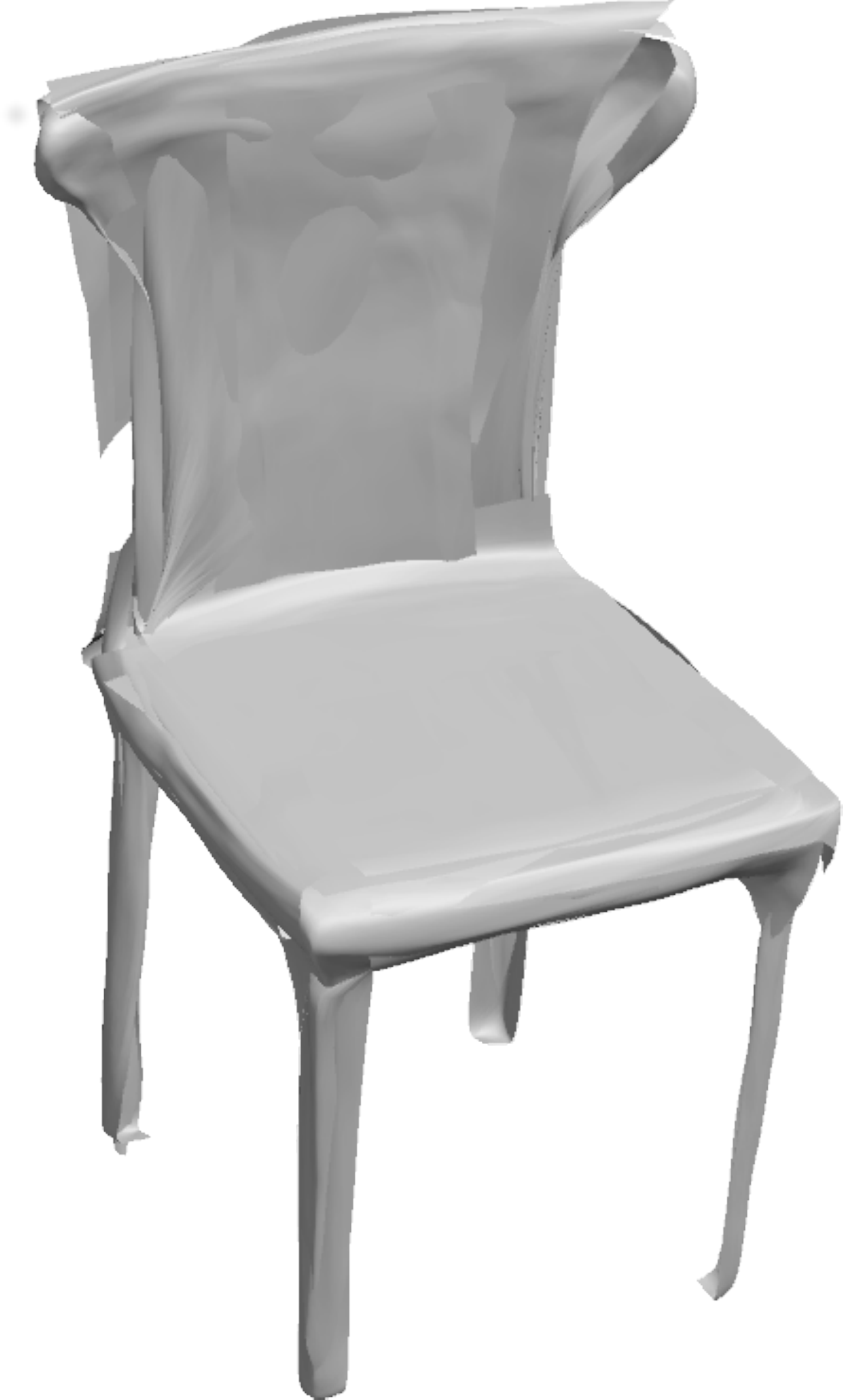}
\caption{\cite{groueix2018atlasnet}-25 patch}
\end{subfigure}
\hfill
\begin{subfigure}[t]{0.13\linewidth}
\includegraphics[width=0.9\linewidth]{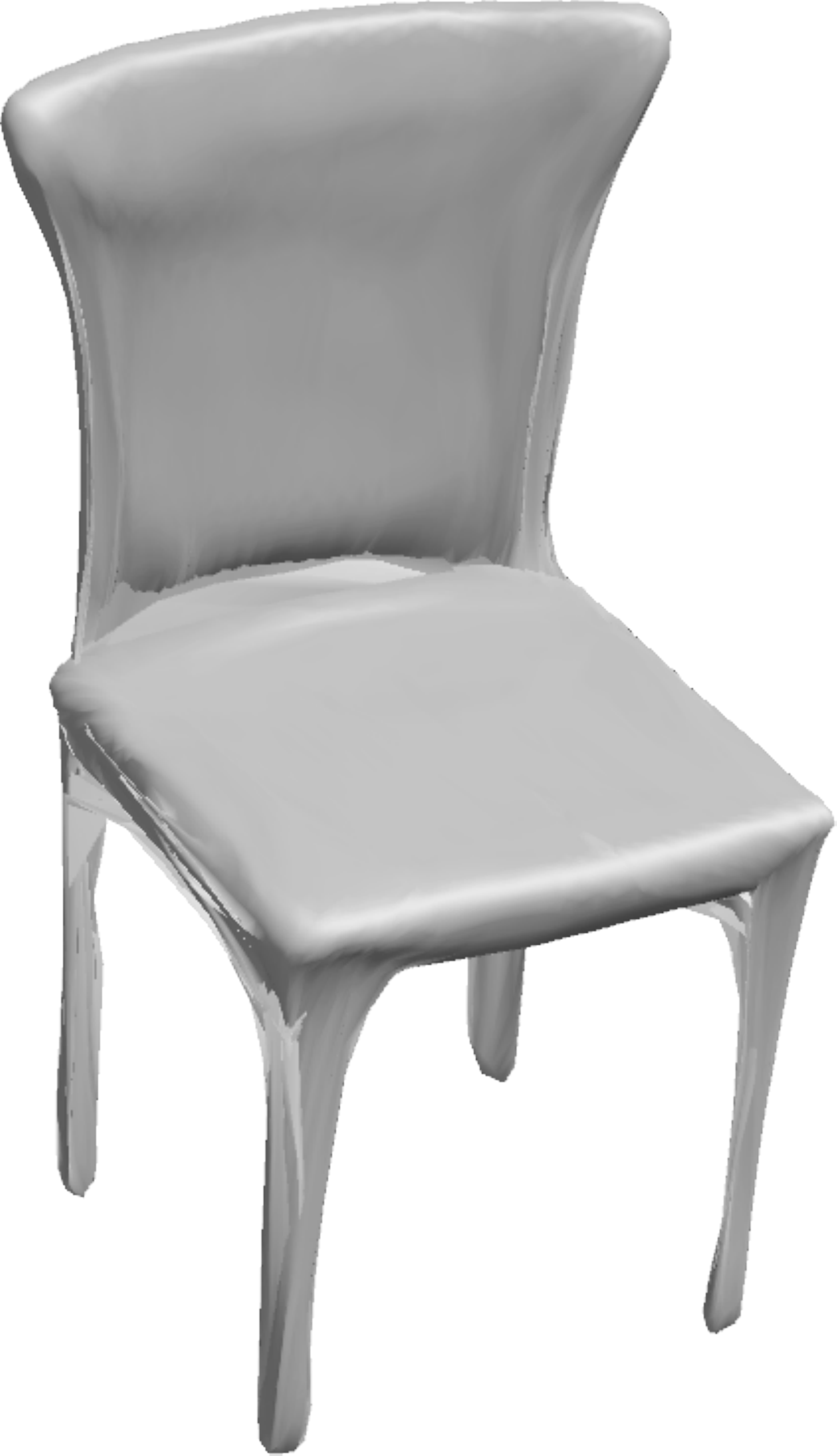}
\caption{\cite{groueix2018atlasnet}-sphere}
\end{subfigure}
\hfill
\begin{subfigure}[t]{0.225\linewidth}
\includegraphics[width=0.9\linewidth]{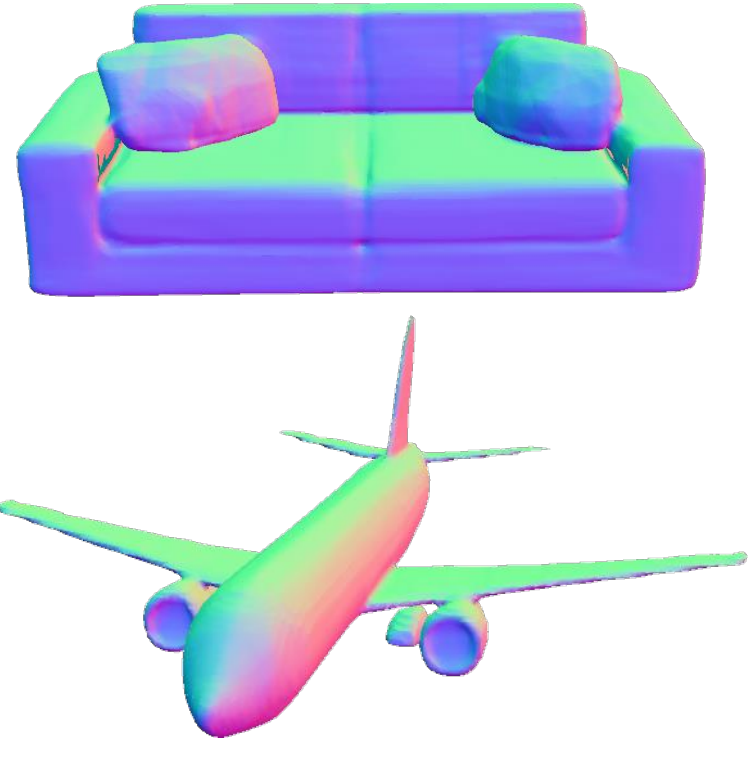}
\caption{Our Result}
\end{subfigure}
\hfill
\begin{subfigure}[t]{0.225\linewidth}
\includegraphics[width=0.9\linewidth]{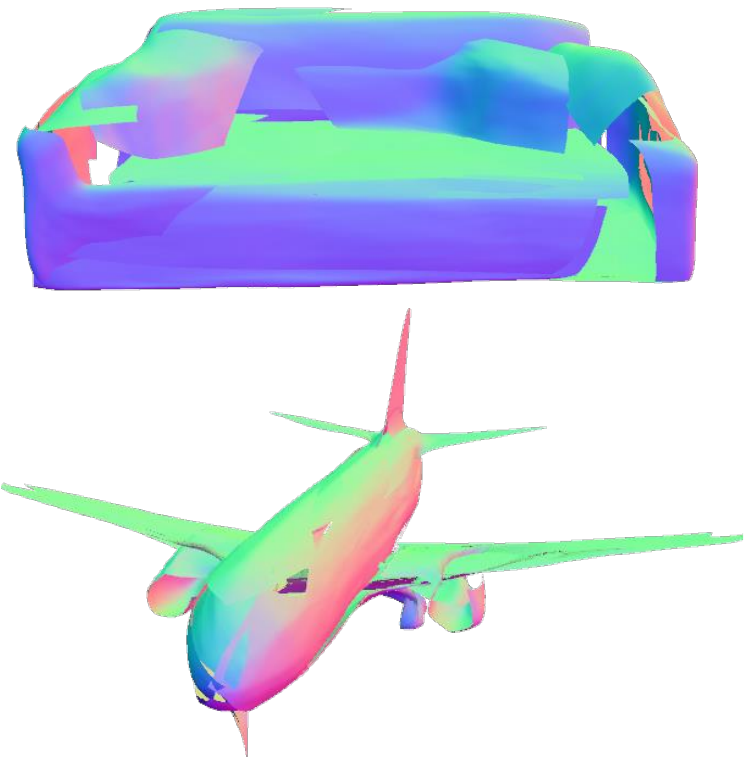}
\caption{\cite{groueix2018atlasnet}-25 patch}
\end{subfigure}
\caption{Reconstruction comparison between DeepSDF and AtlasNet~\cite{groueix2018atlasnet} (with 25-plane and sphere parameterization) for test shapes. Note that AtlasNet fails to capture the fine details of the chair, and that (f) shows holes on the surface of sofa and the plane.}
	\label{fig: recon-compare}
\end{figure*}

\begin{figure*} 
\begin{subfigure}[t]{0.10\linewidth}
\includegraphics[width=0.9\linewidth]{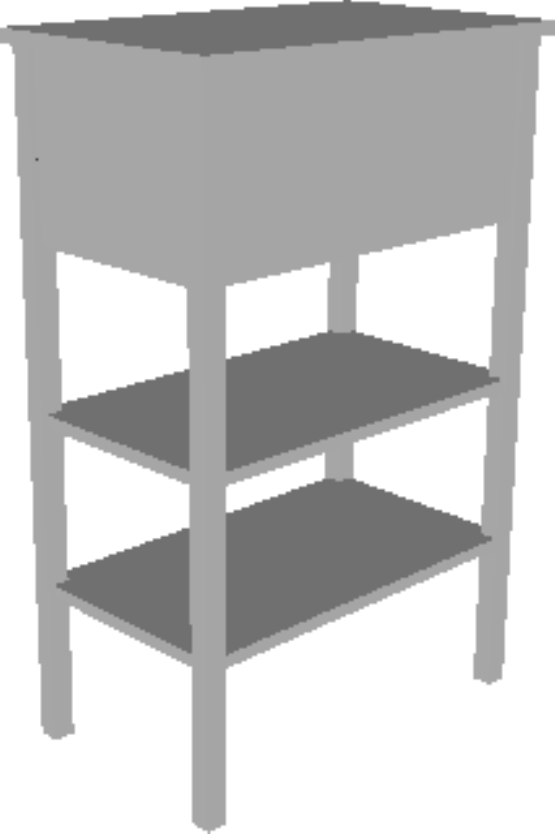}
\end{subfigure}
\begin{subfigure}[t]{0.098\linewidth}
\includegraphics[width=0.9\linewidth]{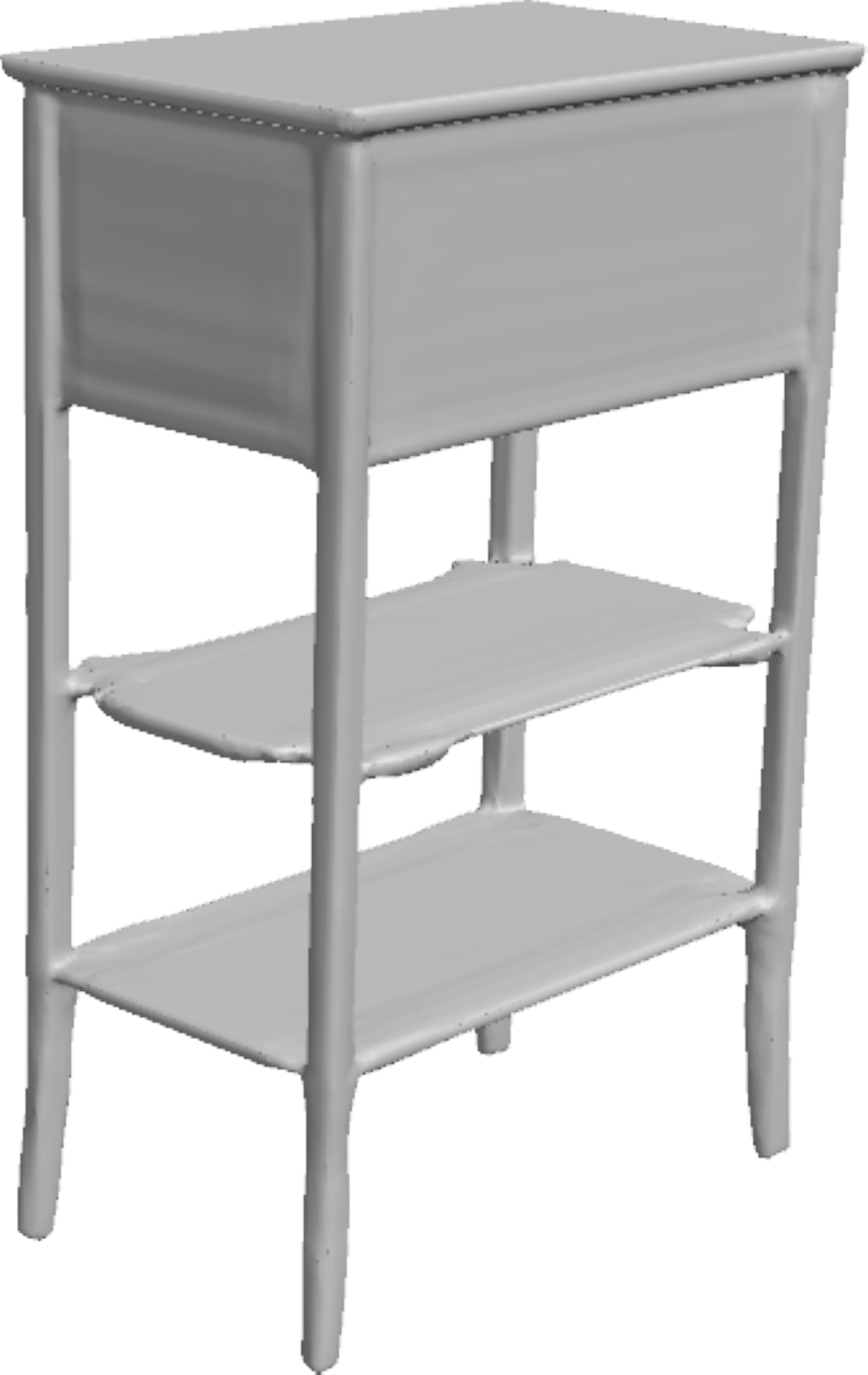}
\end{subfigure}
\hfill
\begin{subfigure}[t]{0.135\linewidth}
\includegraphics[width=0.9\linewidth]{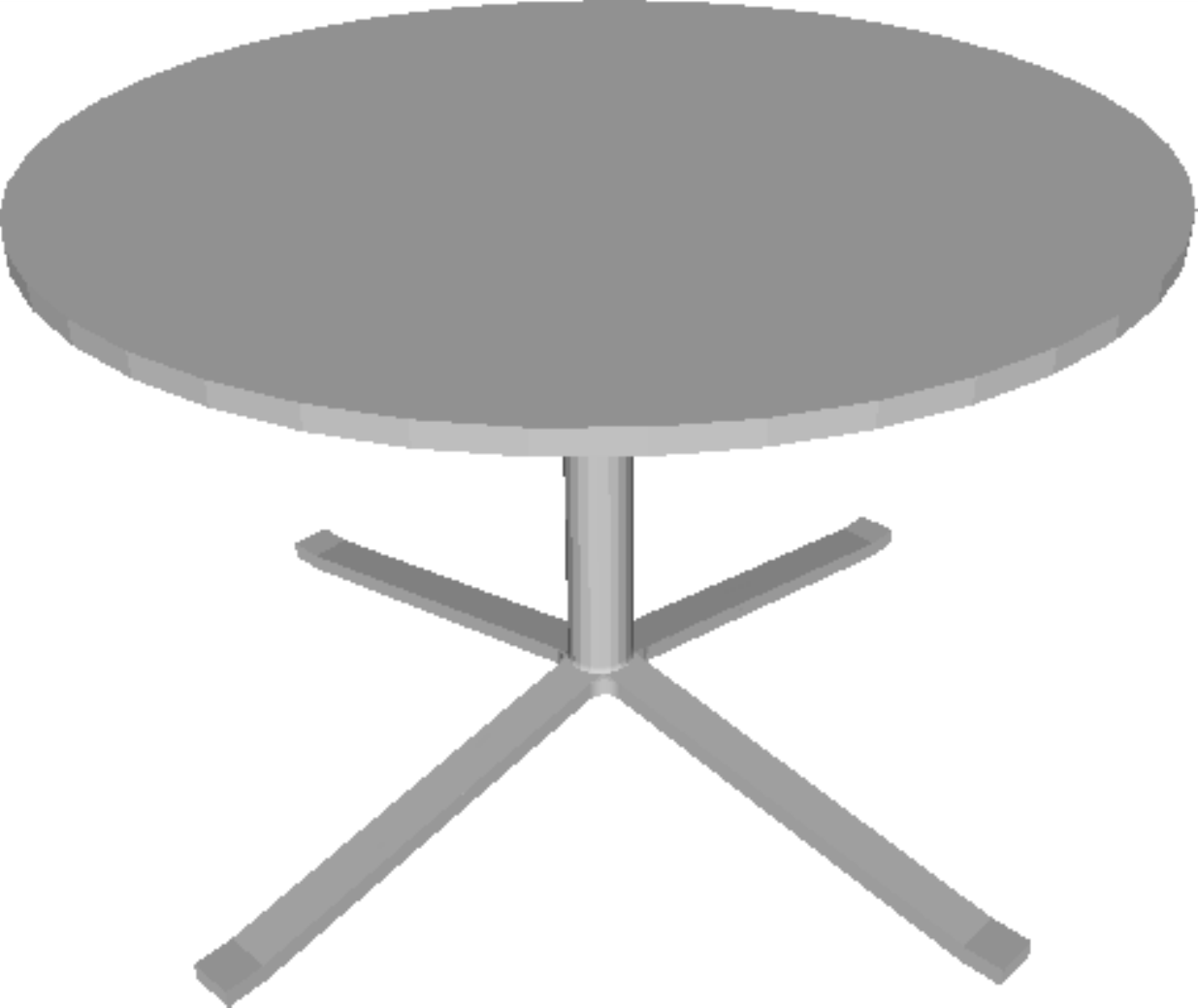}
\end{subfigure}
\begin{subfigure}[t]{0.13\linewidth}
\includegraphics[width=0.9\linewidth]{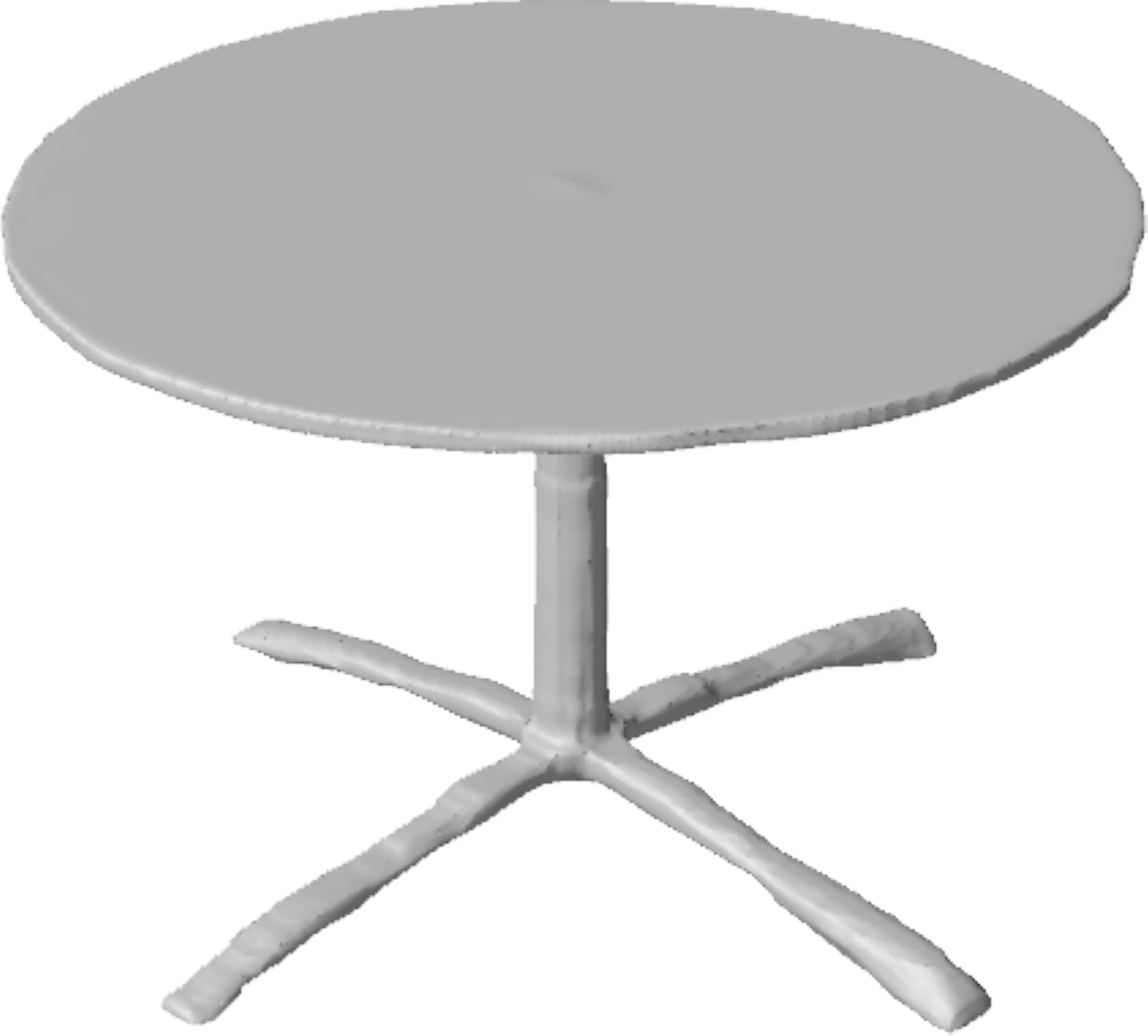}
\end{subfigure}
\hfill
\begin{subfigure}[t]{0.14\linewidth}
\includegraphics[width=0.9\linewidth]{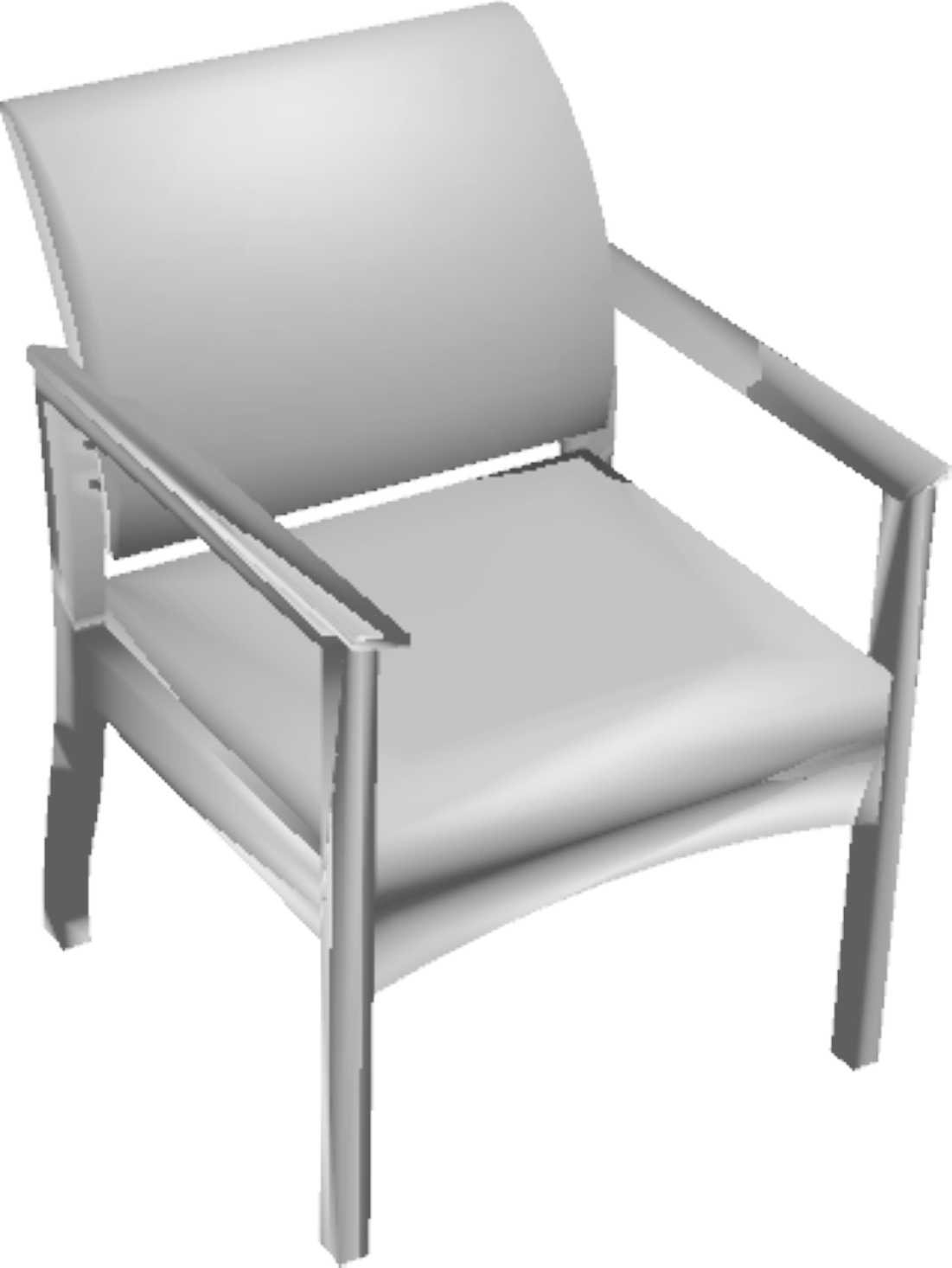}
\end{subfigure}
\begin{subfigure}[t]{0.14\linewidth}
\includegraphics[width=0.9\linewidth]{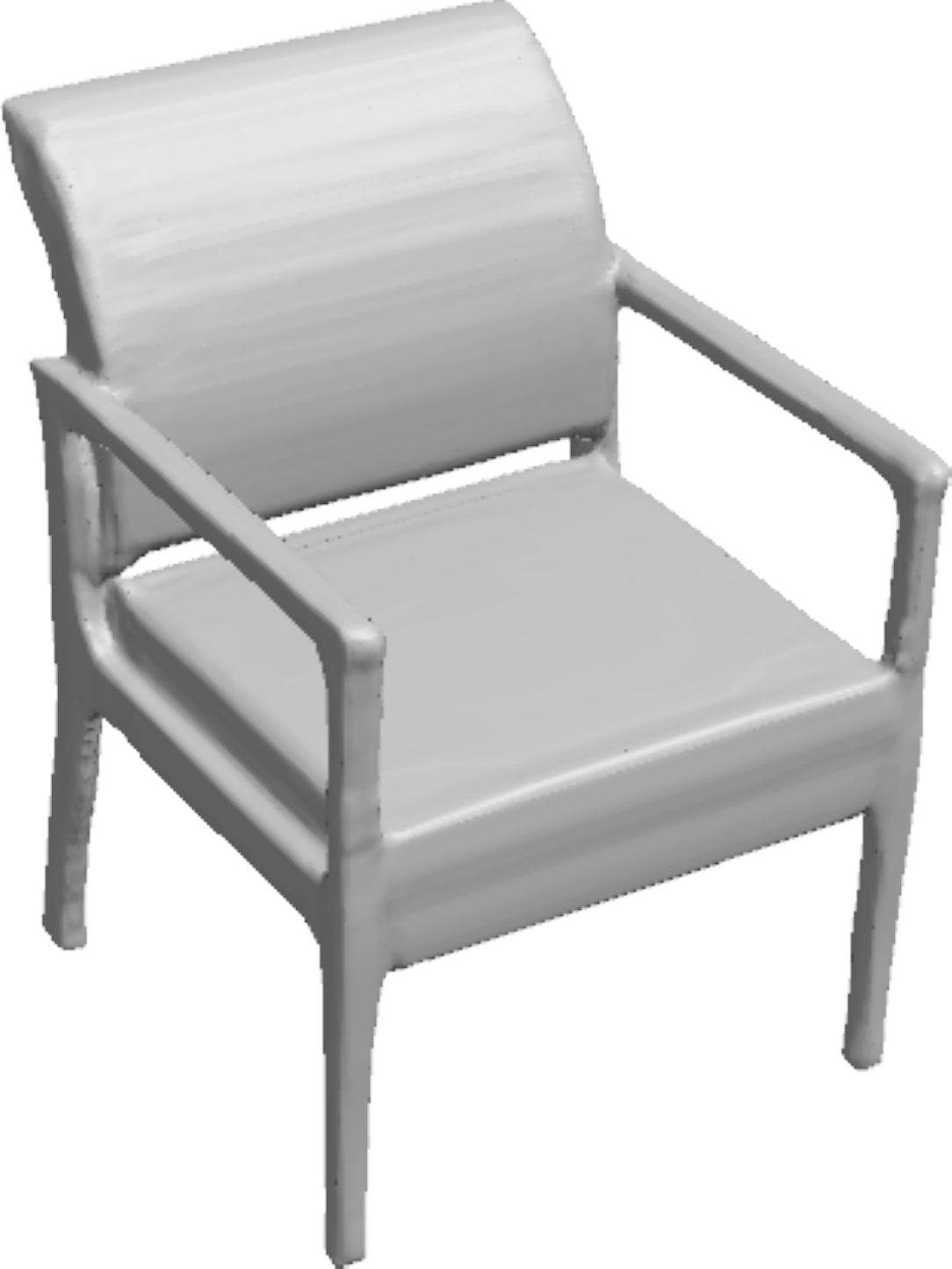}
\end{subfigure}
\hfill
\rulesep
\hfill
\begin{subfigure}[t]{0.13\linewidth}
\includegraphics[width=0.9\linewidth]{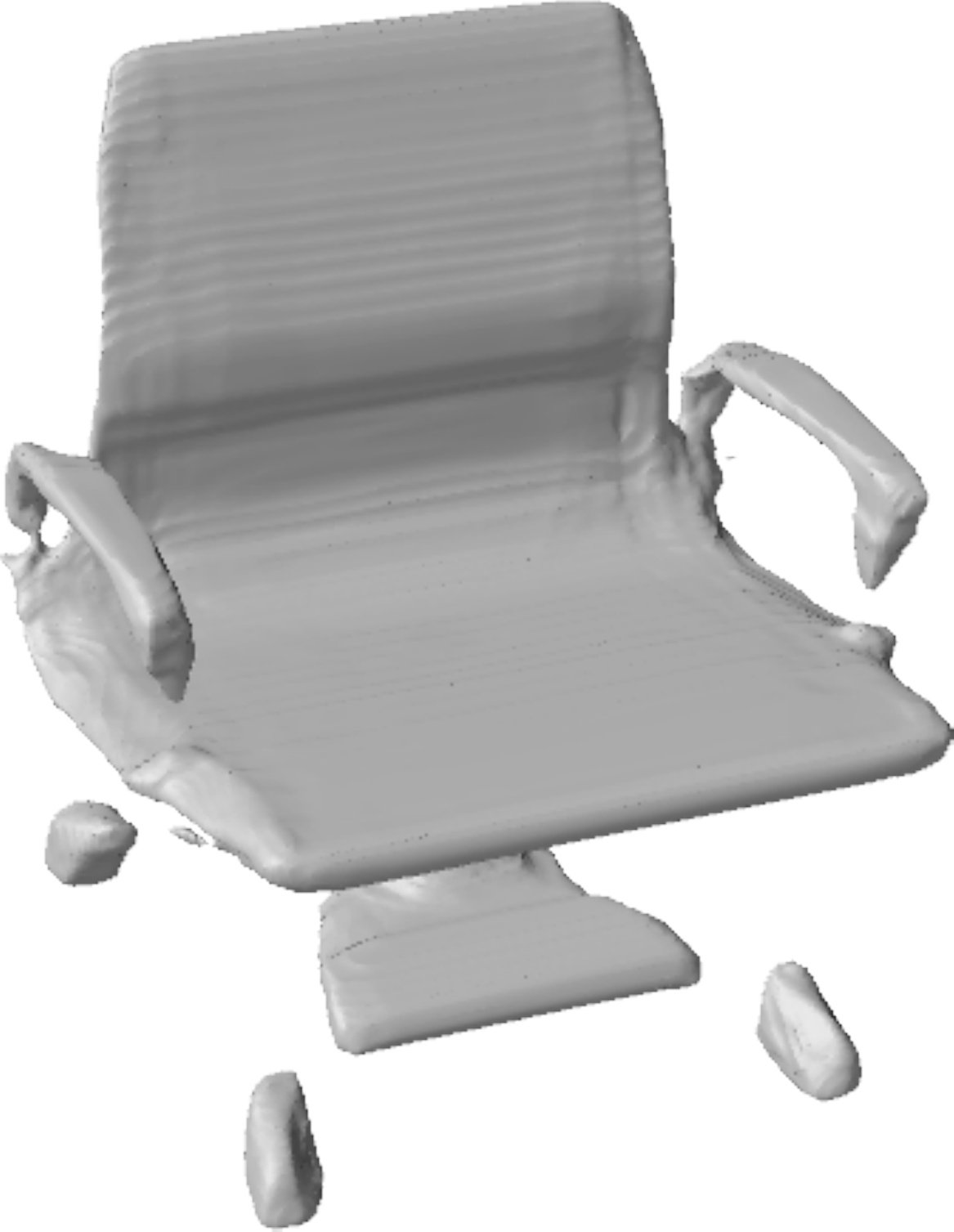}
\end{subfigure}
\begin{subfigure}[t]{0.08\linewidth}
\includegraphics[width=0.9\linewidth]{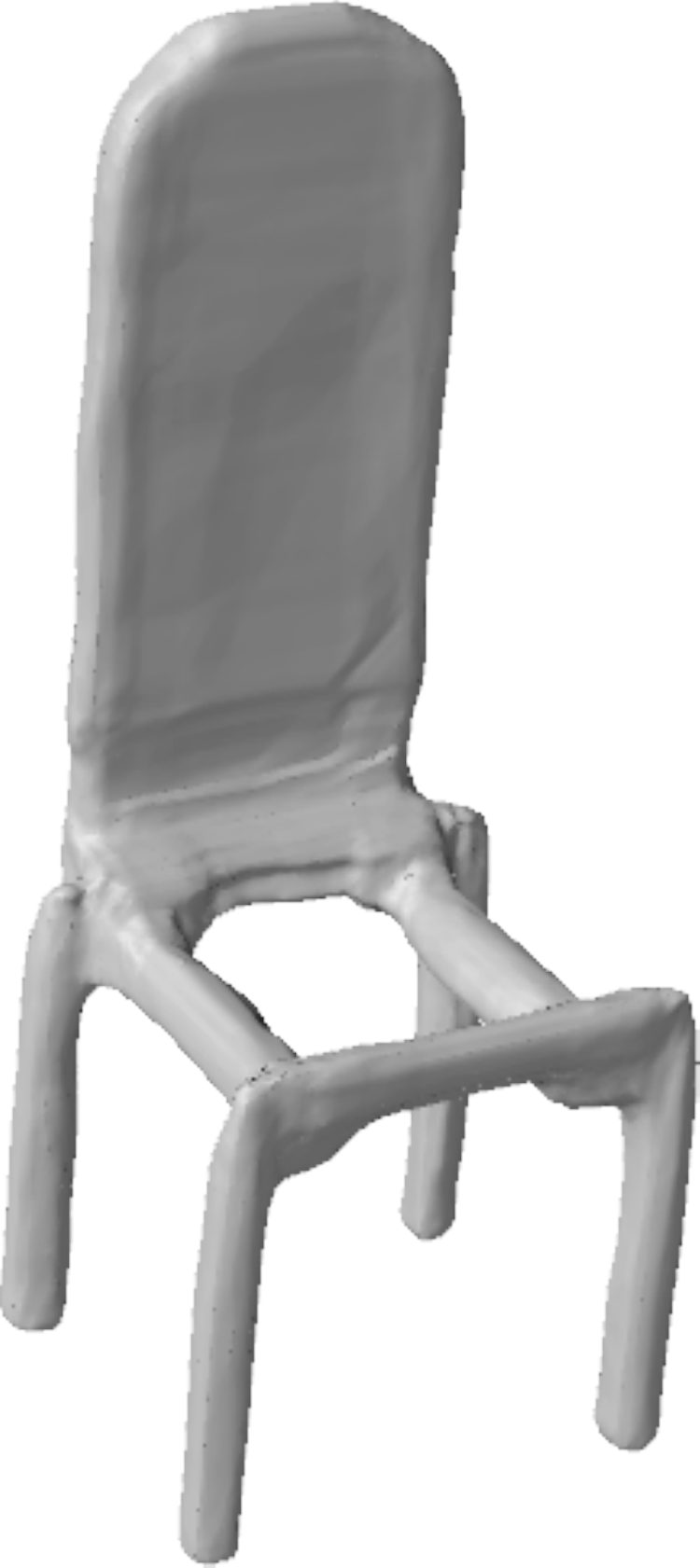}
\end{subfigure}
\caption{Reconstruction of test shapes. From left to right alternating: ground truth shape and our reconstruction. The two right most columns show failure modes of DeepSDF. These failures are likely due to lack of training data and failure of minimization convergence.} 
	\label{fig: reconstruction}
\end{figure*}

For encoding \textit{unknown} shapes, i.e.~shapes in the held-out test set, DeepSDF again significantly outperforms AtlasNet on a wide variety of shape classes and metrics as shown in Table~\ref{tab:3}. Note that AtlasNet performs reasonably well at classes of shapes that have mostly consistent topology without holes (like planes) but struggles more on classes that commonly have holes, like chairs. This is shown in Fig.~\ref{fig: recon-compare} where AtlasNet fails to represent the fine detail of the back of the chair. Figure~\ref{fig: reconstruction} shows more examples of detailed reconstructions on test data from DeepSDF as well as two example failure cases.

 \begin{table}[t]
\centering
\footnotesize
 \begin{tabular}{|l|r|r|r|r|r|}
 \hhline{|=|=|=|=|=|=|}
  \multicolumn{1}{|l|}{CD, mean} & \multicolumn{1}{c|}{chair} & \multicolumn{1}{c|}{plane} & \multicolumn{1}{c|}{table} & \multicolumn{1}{c|}{lamp} & \multicolumn{1}{c|}{sofa}  \\
 \hline
 AtlasNet-Sph.  & 0.752 & 0.188 & 0.725 & 2.381 & 0.445\\
 AtlasNet-25      & 0.368 & 0.216 & \bf{0.328} & 1.182 & 0.411\\
 DeepSDF        &  \bf{0.204} & \bf{0.143} & 0.553 & \bf{0.832} & \bf{0.132}  \\
 \hline
   \hhline{|=|=|=|=|=|=|}
  \multicolumn{1}{|l}{CD, median} &  \multicolumn{1}{c}{ \ } &  \multicolumn{1}{c}{ \ } &  \multicolumn{1}{c}{ \ } &  \multicolumn{1}{c}{ \ }  &  \multicolumn{1}{c|}{} \\
 \hline
  AtlasNet-Sph.  & 0.511 & 0.079 & 0.389 & 2.180 & 0.330 \\
 AtlasNet-25      & 0.276 & 0.065 & 0.195 & 0.993 & 0.311  \\
 DeepSDF        &  \bf{0.072} & \bf{0.036} & \bf{0.068} & \bf{0.219} & \bf{0.088}  \\
 \hline
\hhline{|=|=|=|=|=|=|}
  \multicolumn{1}{|l}{EMD, mean} &  \multicolumn{1}{c}{ \ } &  \multicolumn{1}{c}{ \ } &  \multicolumn{1}{c}{ \ } &  \multicolumn{1}{c}{ \ }  &  \multicolumn{1}{c|}{} \\
 \hline
  AtlasNet-Sph.  & 0.071 & 0.038 & 0.060 & 0.085 & 0.050  \\
 AtlasNet-25      & 0.064 & 0.041 & 0.073 & 0.062 & 0.063  \\
 DeepSDF        &  \bf{0.049} & \bf{0.033} & \bf{0.050} & \bf{0.059} & \bf{0.047}  \\
 \hline

\hhline{|=|=|=|=|=|=|}
  \multicolumn{1}{|l}{Mesh acc., mean} &  \multicolumn{1}{c}{ \ } &  \multicolumn{1}{c}{ \ } &  \multicolumn{1}{c}{ \ } &  \multicolumn{1}{c}{ \ }  &  \multicolumn{1}{c|}{} \\
 \hline
  AtlasNet-Sph.  & 0.033 & 0.013 & 0.032 & 0.054 & 0.017  \\
 AtlasNet-25      & 0.018 & 0.013 & 0.014 & 0.042 & 0.017  \\
 DeepSDF        & \bf{0.009} & \bf{0.004} & \bf{0.012} & \bf{0.013} & \bf{0.004}  \\
 \hline

   \end{tabular}
   \caption{Comparison for representing unknown shapes (U) for various classes of ShapeNet.  Mesh accuracy as defined in \cite{seitz2006comparison} is the minimum distance $d$ such that 90\% of generated points are within $d$ of the ground truth mesh.  Lower is better for all metrics.} 
   \label{tab:3}
\end{table}

\begin{table}[t]
\centering
\footnotesize
\begin{tabular}{|l|c|c|c|c||c|c|}
\hline
 & \multicolumn{4}{c||}{\footnotesize \textit{lower is better}} & \multicolumn{2}{c|}{\footnotesize \textit{higher is better}} \\
Method                  & CD, &  CD,        &    &     Mesh & Mesh & Cos \\
 \textbackslash Metric & med. & mean & EMD & acc. & comp. &  sim.\\
\hhline{|=|=|=|=|=||=|=|} \hline
\multicolumn{7}{|l|}{chair} \\ \hline
3D-EPN   & 2.25 & 2.83  & 0.084 & 0.059 & 0.209 & 0.752\\
DeepSDF  & \bf{1.28} & \bf{2.11}   & \bf{0.071} & \bf{0.049} & \bf{0.500} & \bf{0.766} \\ \hline
\multicolumn{7}{|l|}{plane} \\ \hline
3D-EPN   & 1.63  & 2.19 & 0.063 & 0.040 & 0.165 & 0.710\\
DeepSDF & \bf{0.37} & \bf{1.16} & \bf{0.049} & \bf{0.032} & \bf{0.722} & \bf{0.823} \\ \hline
\multicolumn{7}{|l|}{sofa} \\ \hline
3D-EPN   & 2.03  & 2.18 & 0.071 & 0.049 & 0.254 & 0.742\\
DeepSDF & \bf{0.82} & \bf{1.59} & \bf{0.059} & \bf{0.041} & \bf{0.541} & \bf{0.810} \\ \hline
\end{tabular}
\caption{Comparison for shape completion (C) from partial range scans of unknown shapes from ShapeNet.}
\label{tab:5}
\end{table}

\subsection{Shape Completion}\label{sec:completion}
A major advantage of the proposed DeepSDF approach for representation learning is that inference can be performed from an arbitrary number of SDF samples. In the DeepSDF framework, shape completion amounts to solving for the shape code that best explains a partial shape observation via Eq.~\ref{eq:testtime}. Given the shape-code a complete shape can be rendered using the priors encoded in the decoder.

We test our completion scheme using single view depth observations 
which is a common use-case and maps well to our architecture without modification. Note that we currently require the depth observations in the canonical shape frame of reference.

\begin{figure*} 
\begin{subfigure}[t]{0.12\linewidth}
\includegraphics[width=0.8\linewidth]{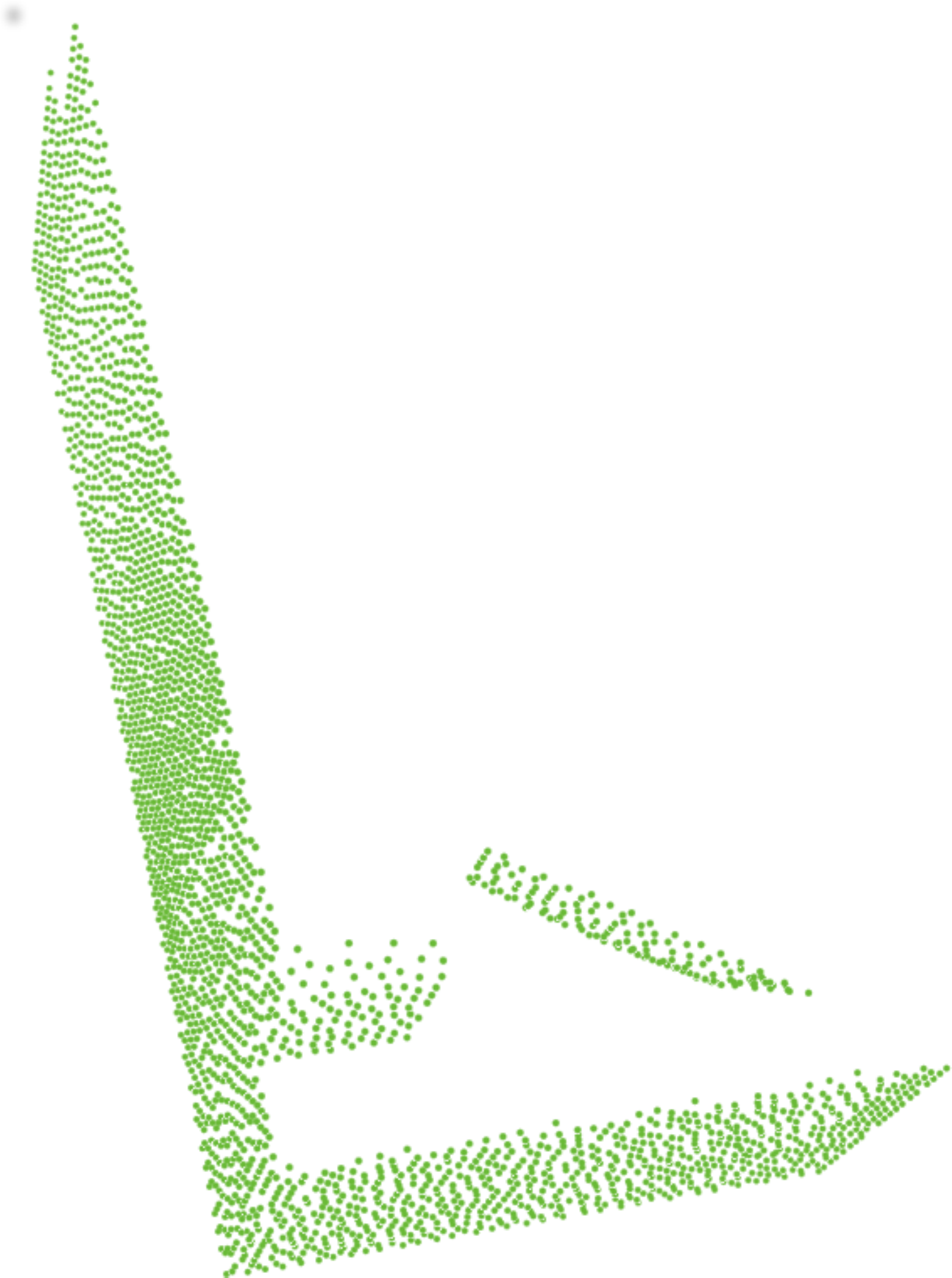}
\end{subfigure}
\hfill
\begin{subfigure}[t]{0.12\linewidth}
\includegraphics[width=0.8\linewidth]{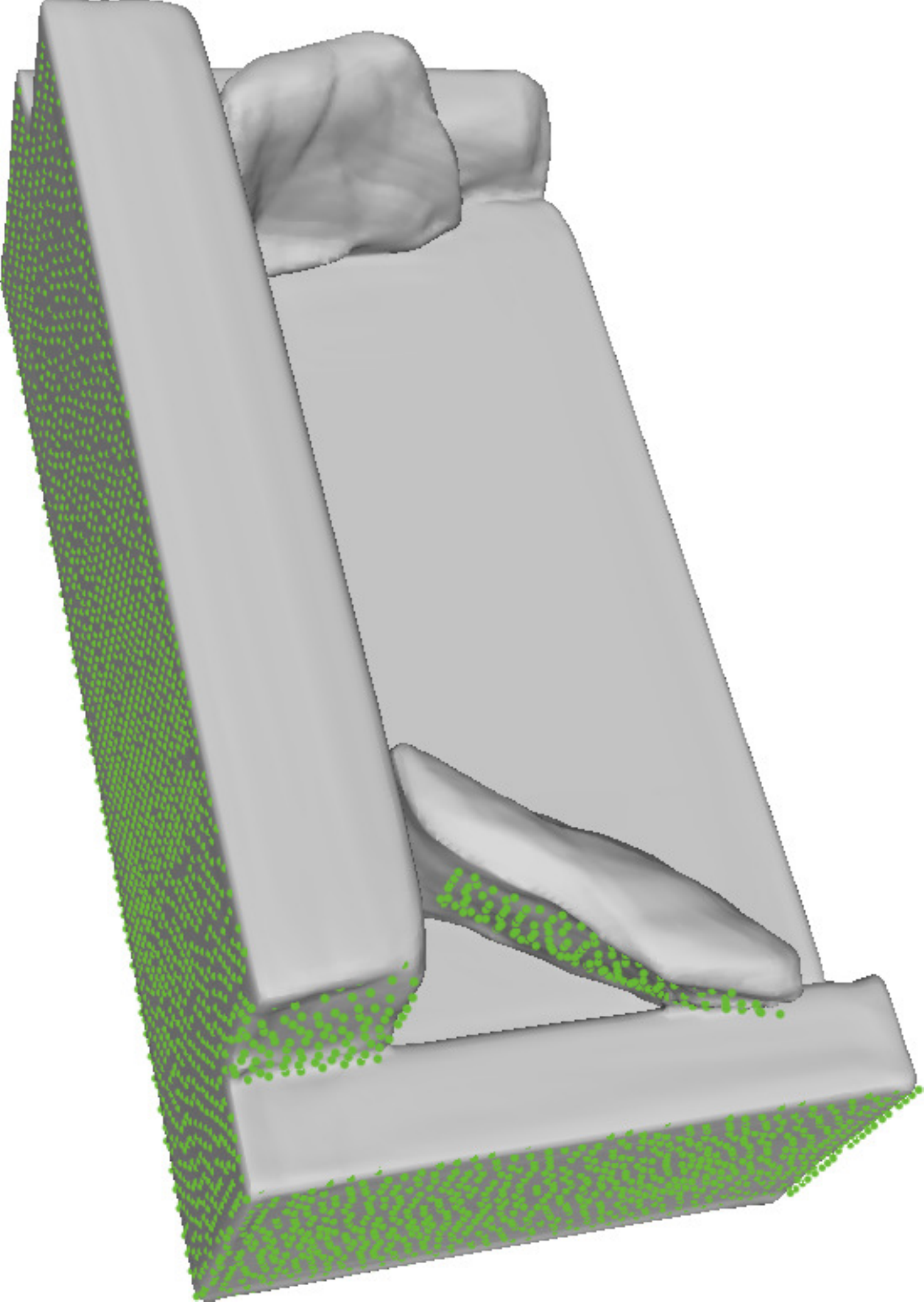}
\end{subfigure}
\hfill
\begin{subfigure}[t]{0.24\linewidth}
\includegraphics[width=0.8\linewidth]{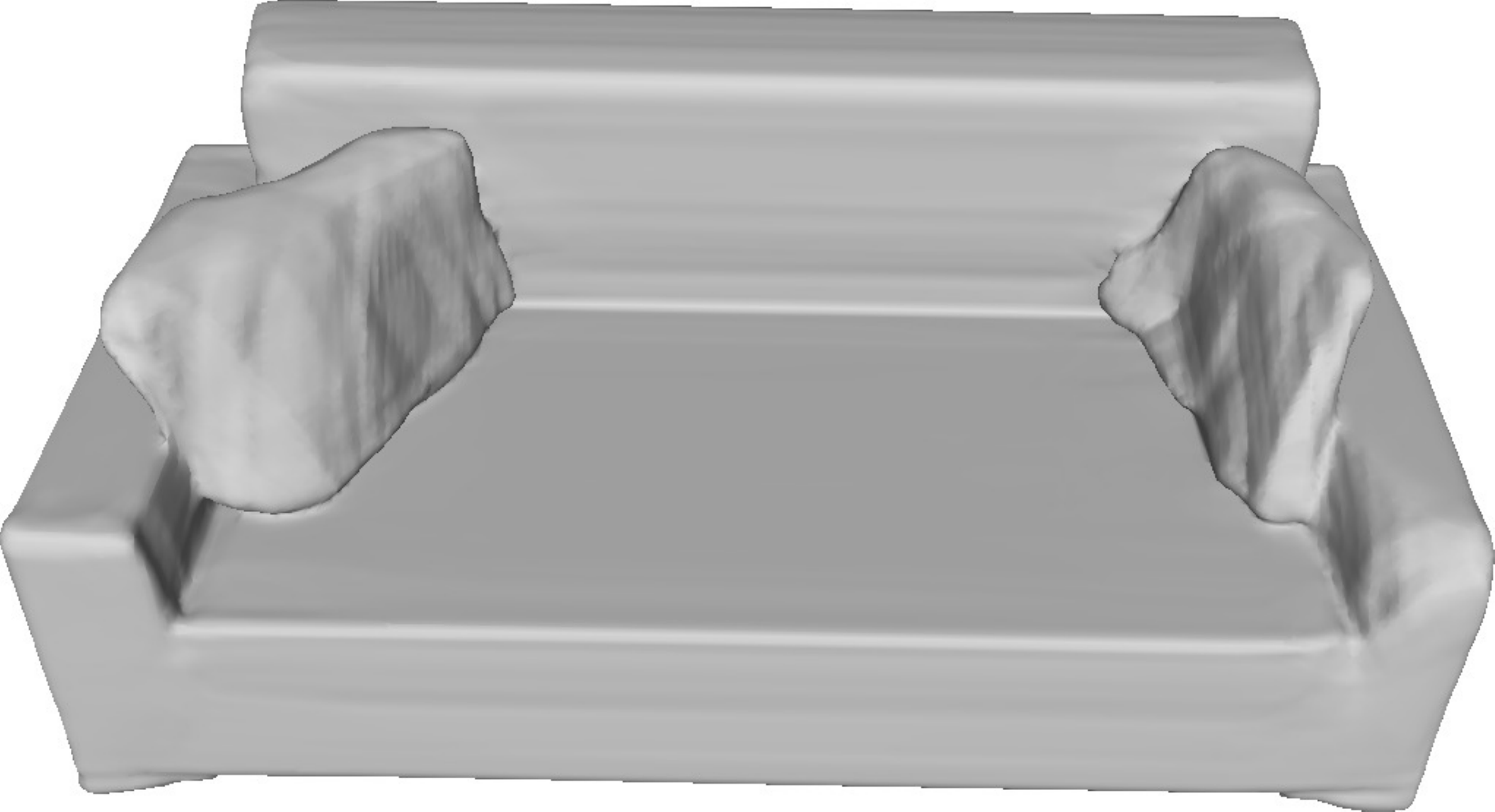}
\end{subfigure}
\begin{subfigure}[t]{0.24\linewidth}
\includegraphics[width=0.8\linewidth]{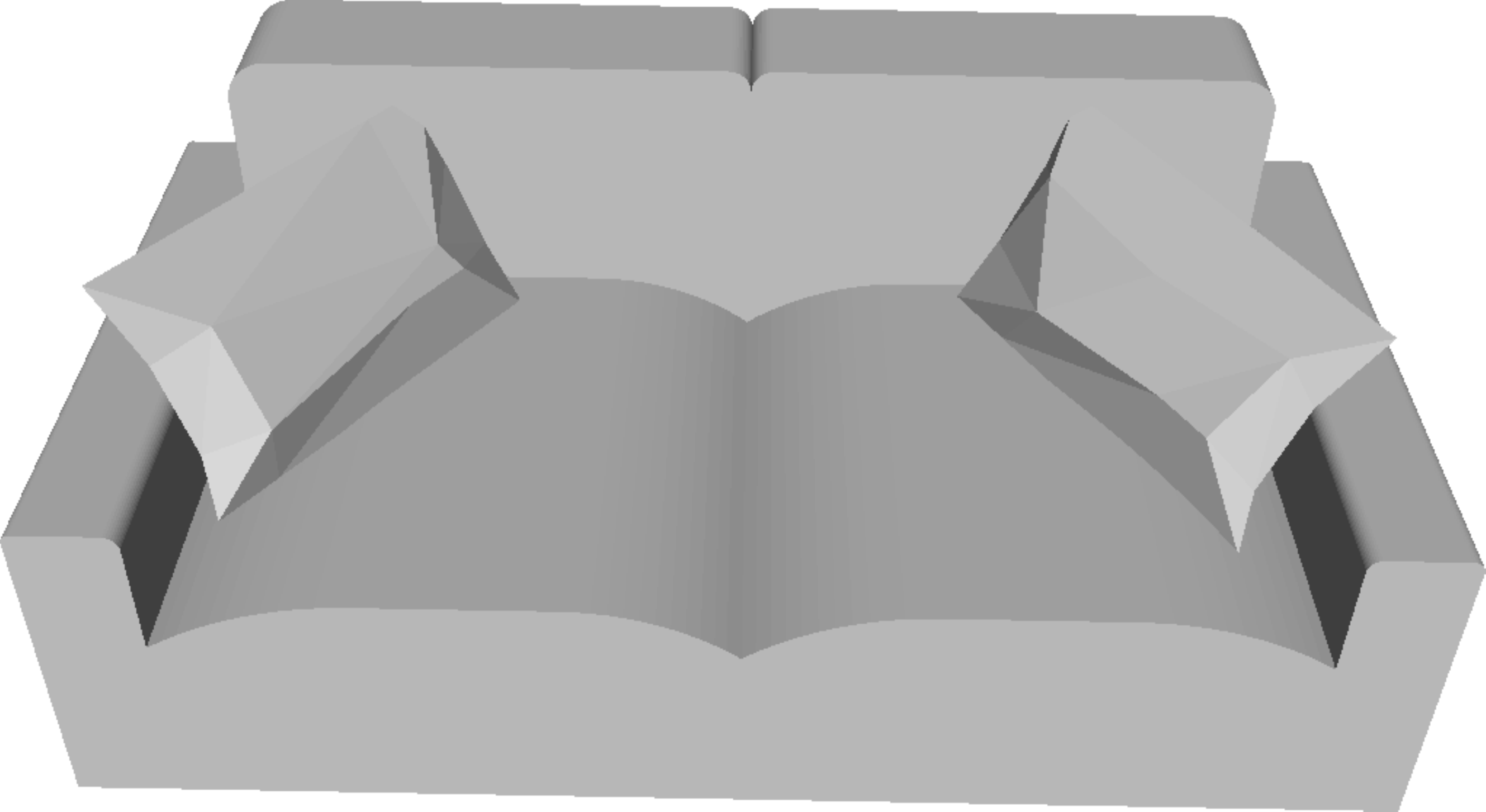}
\end{subfigure}
\begin{subfigure}[t]{0.23\linewidth}
\includegraphics[width=0.8\linewidth]{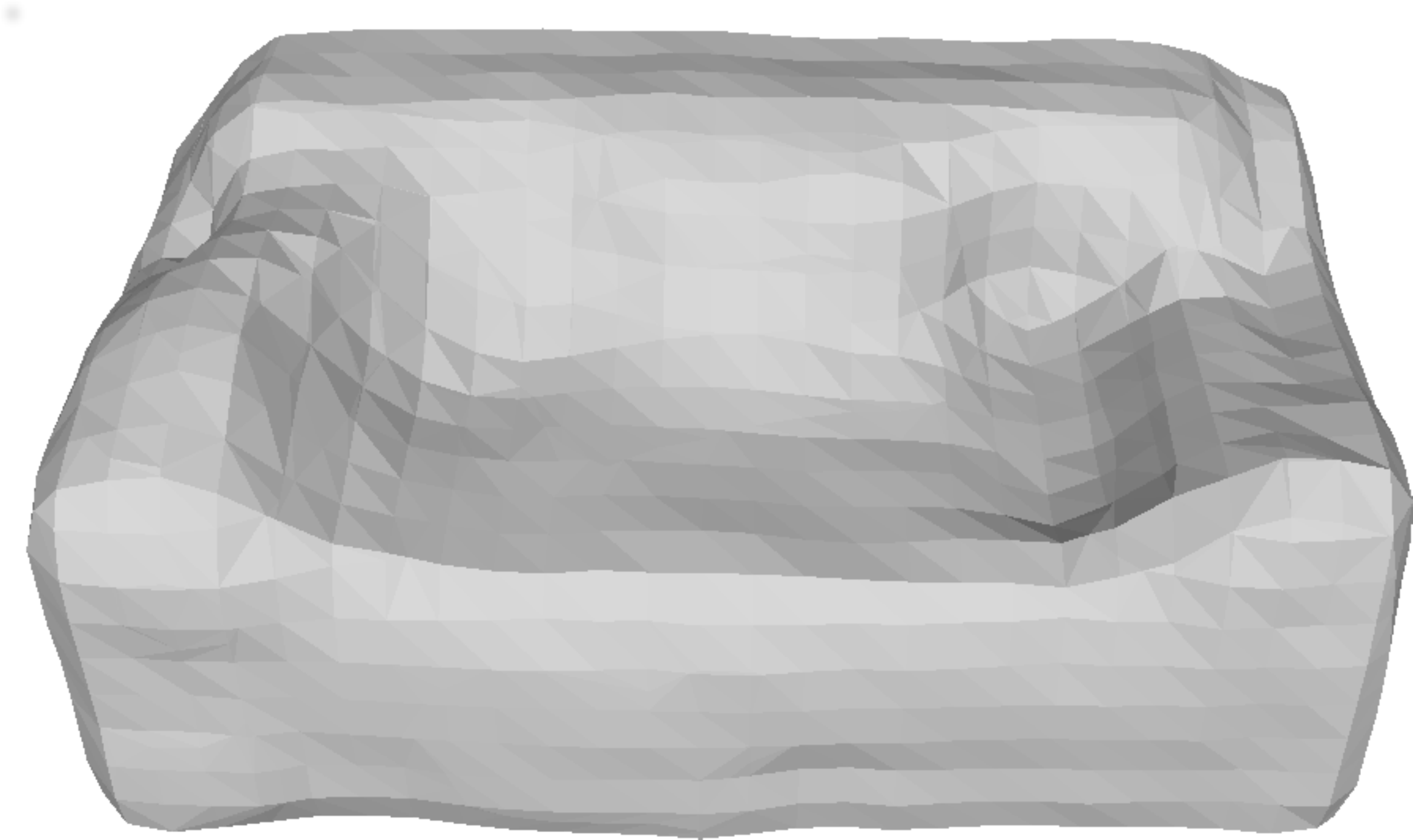}
\end{subfigure}%

\vspace{0.1cm}
\begin{subfigure}[t]{0.19\linewidth}
\includegraphics[width=0.9\linewidth]{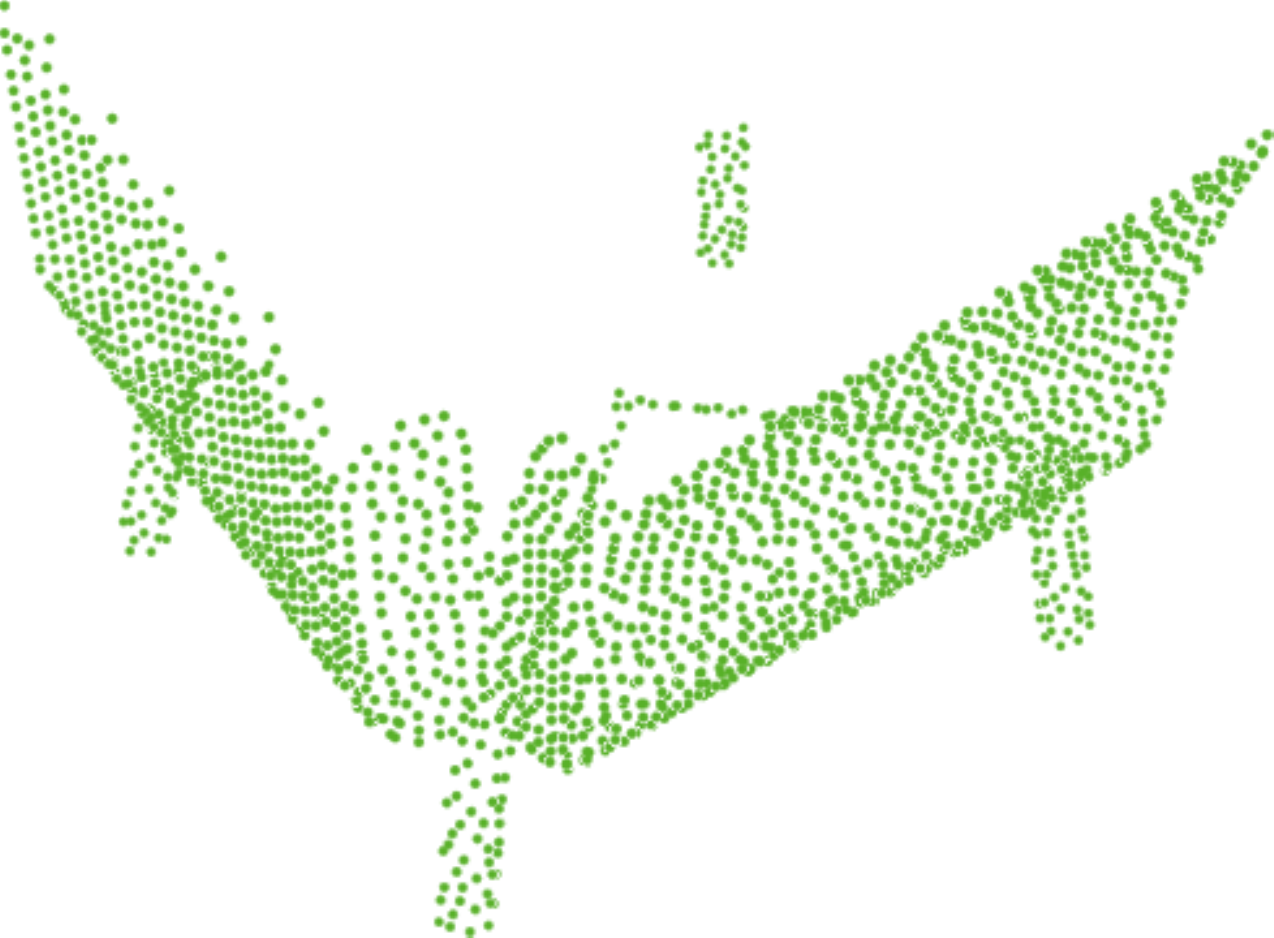}
\end{subfigure}
\hfill
\begin{subfigure}[t]{0.19\linewidth}
\includegraphics[width=0.9\linewidth]{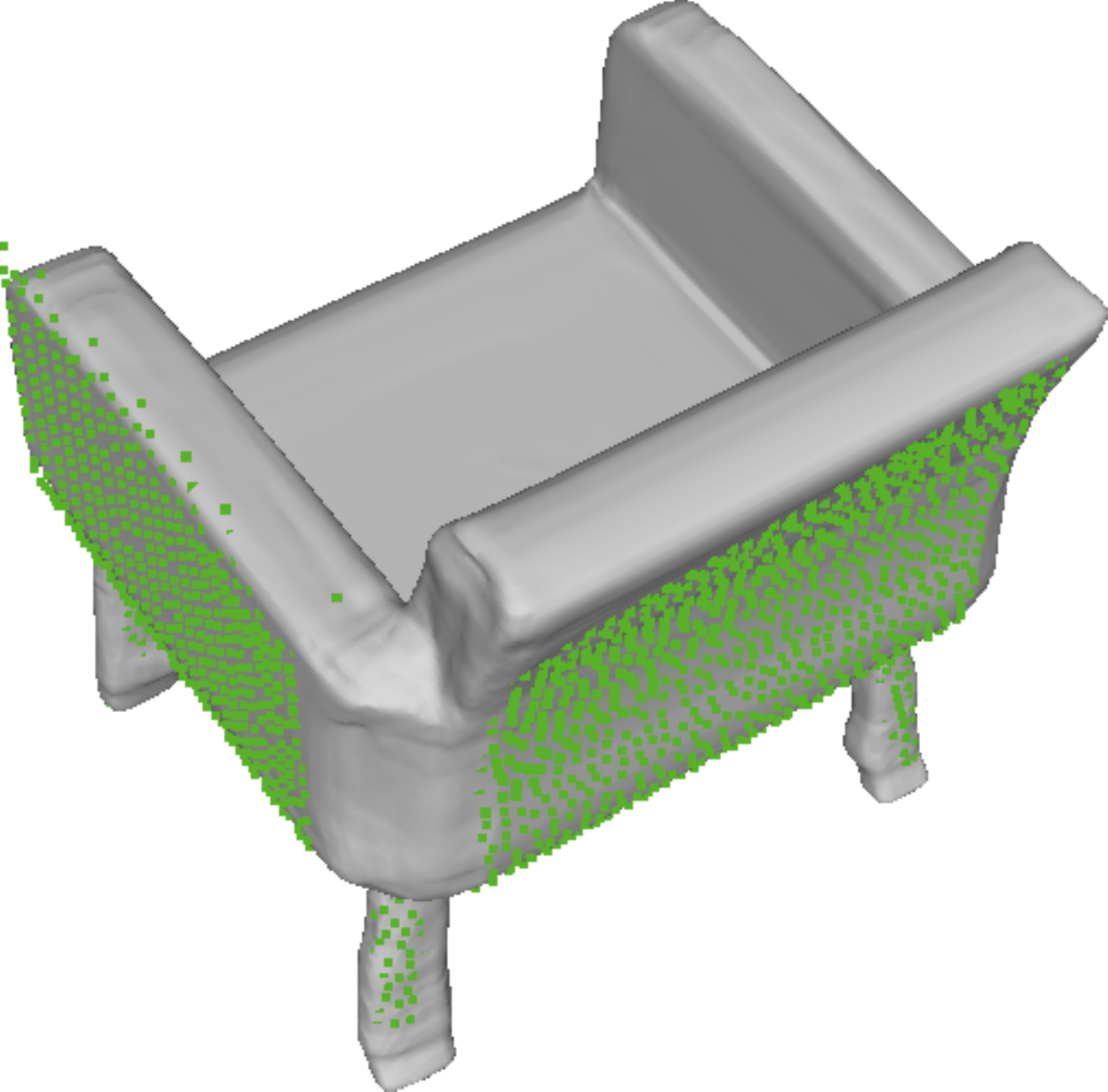}
\end{subfigure}
\hfill
\begin{subfigure}[t]{0.16\linewidth}
\includegraphics[width=0.9\linewidth]{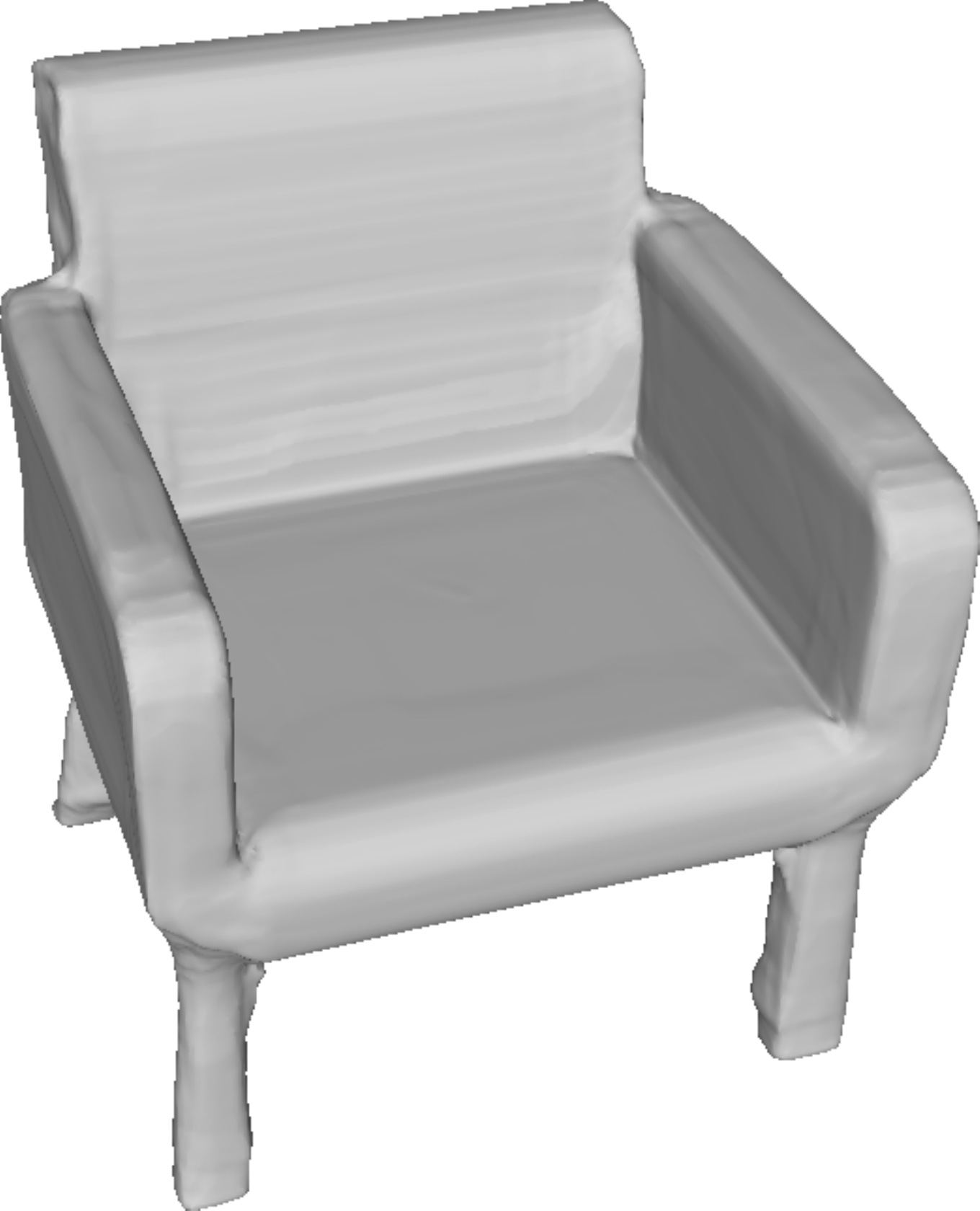}
\end{subfigure}
\hfill
\begin{subfigure}[t]{0.16\linewidth}
\includegraphics[width=0.9\linewidth]{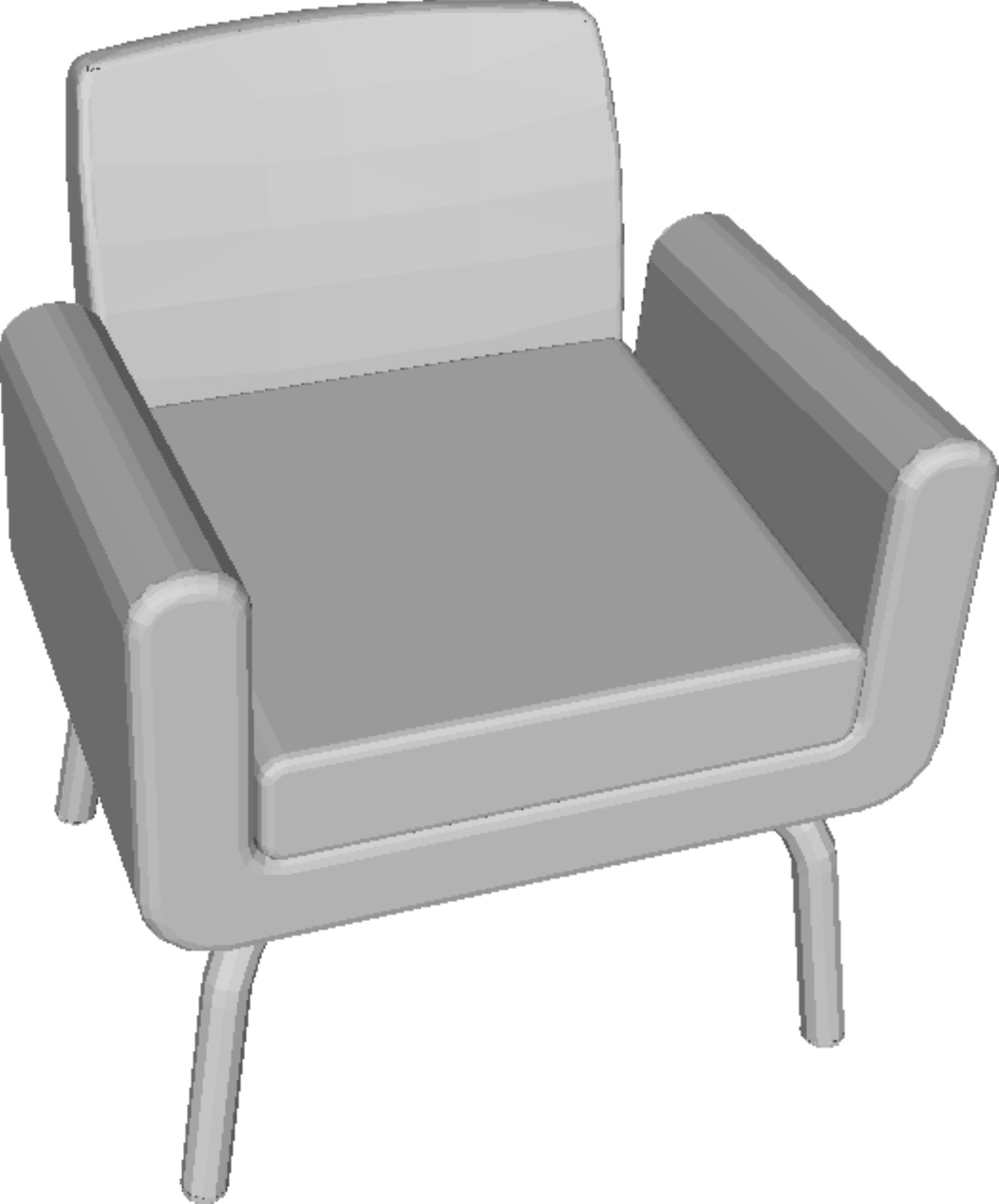}
\end{subfigure}
\hfill
\begin{subfigure}[t]{0.14\linewidth}
\includegraphics[width=0.9\linewidth]{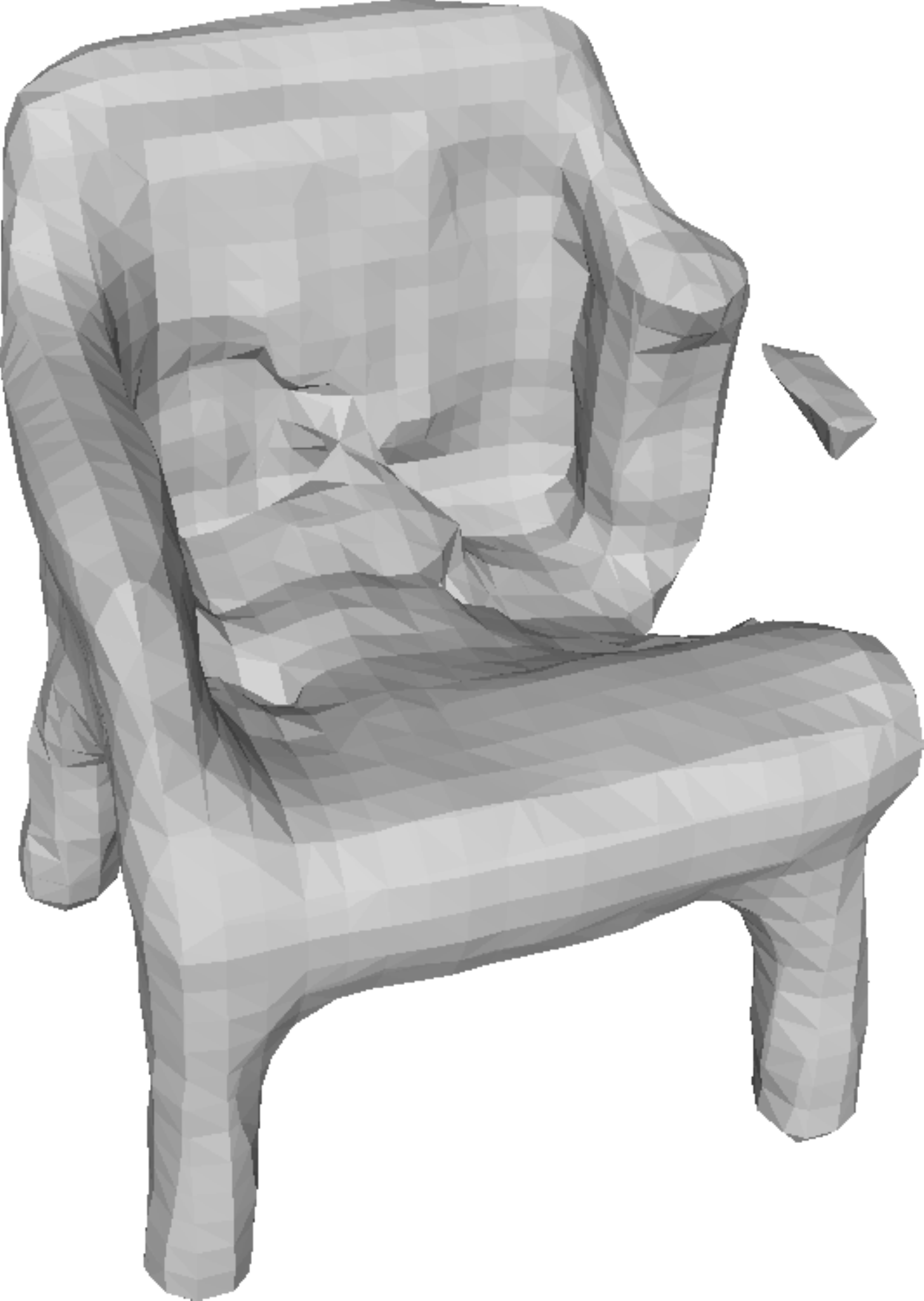}
\end{subfigure}%

\begin{subfigure}[t]{0.21\linewidth}
\includegraphics[width=0.9\linewidth]{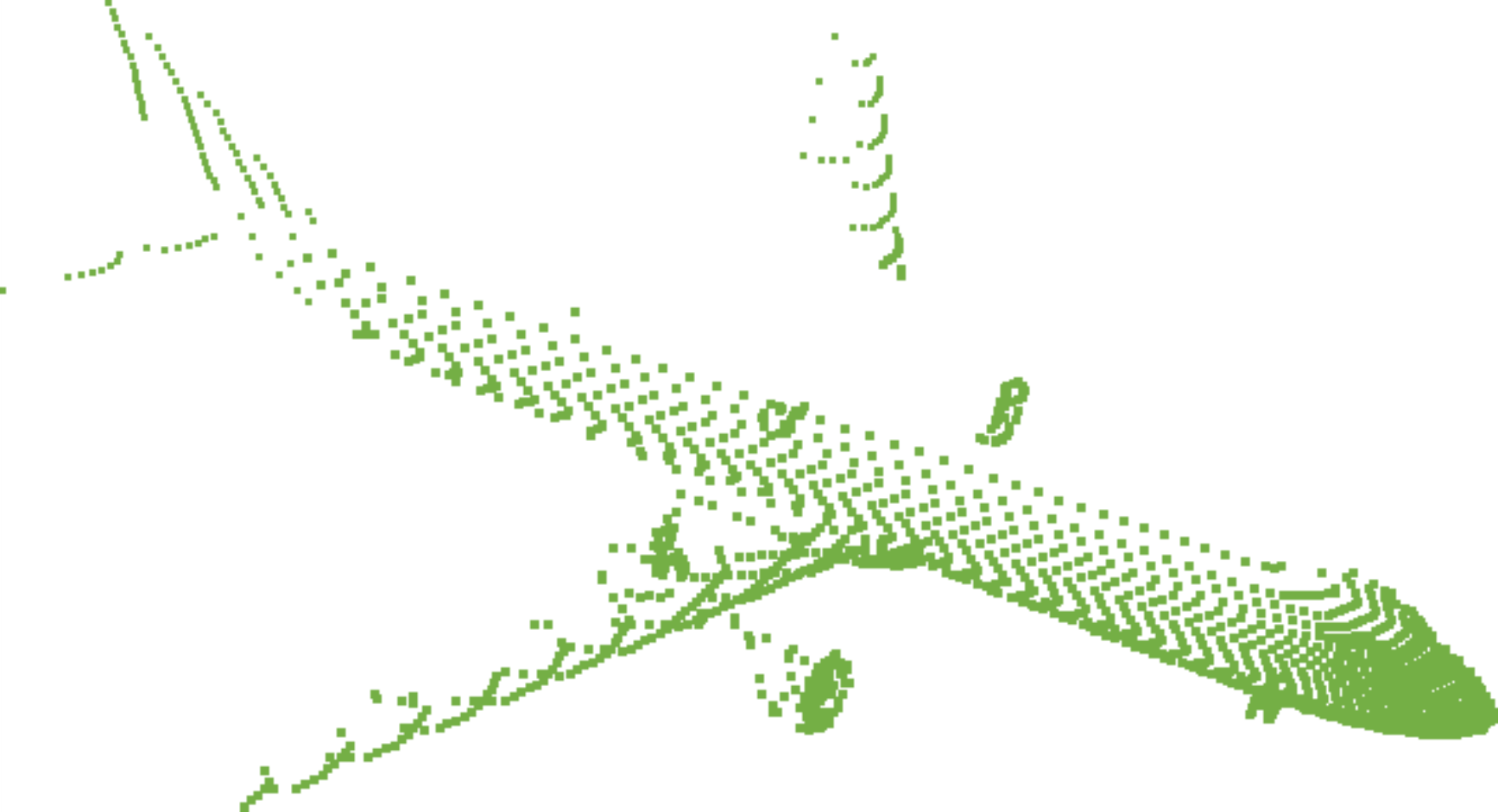}
\caption{Input Depth}
\end{subfigure}
\hfill
\begin{subfigure}[t]{0.21\linewidth}
\includegraphics[width=0.9\linewidth]{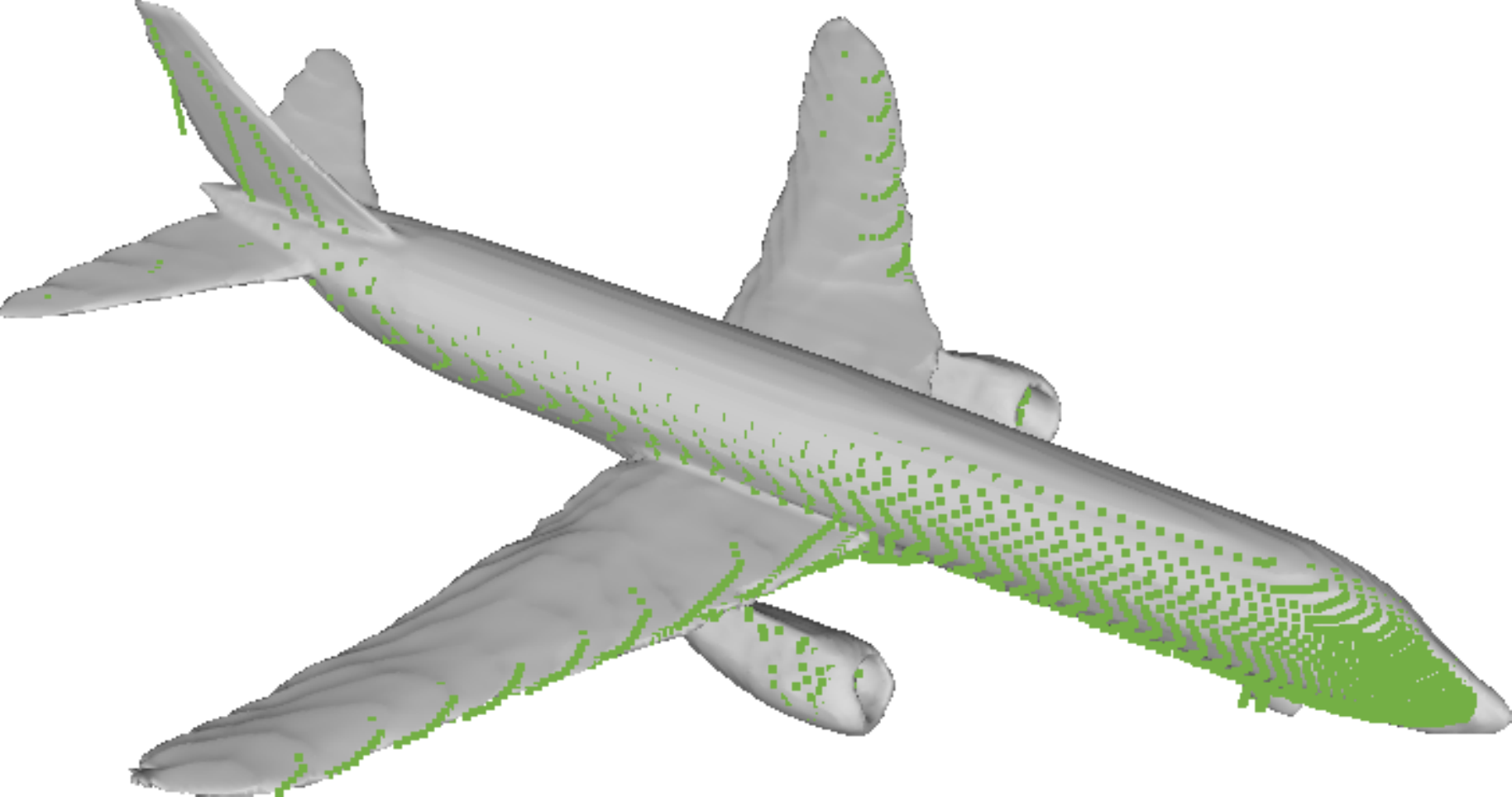}
\caption{Completion (ours)}
\end{subfigure}
\hfill
\begin{subfigure}[t]{0.18\linewidth}
\includegraphics[width=0.9\linewidth]{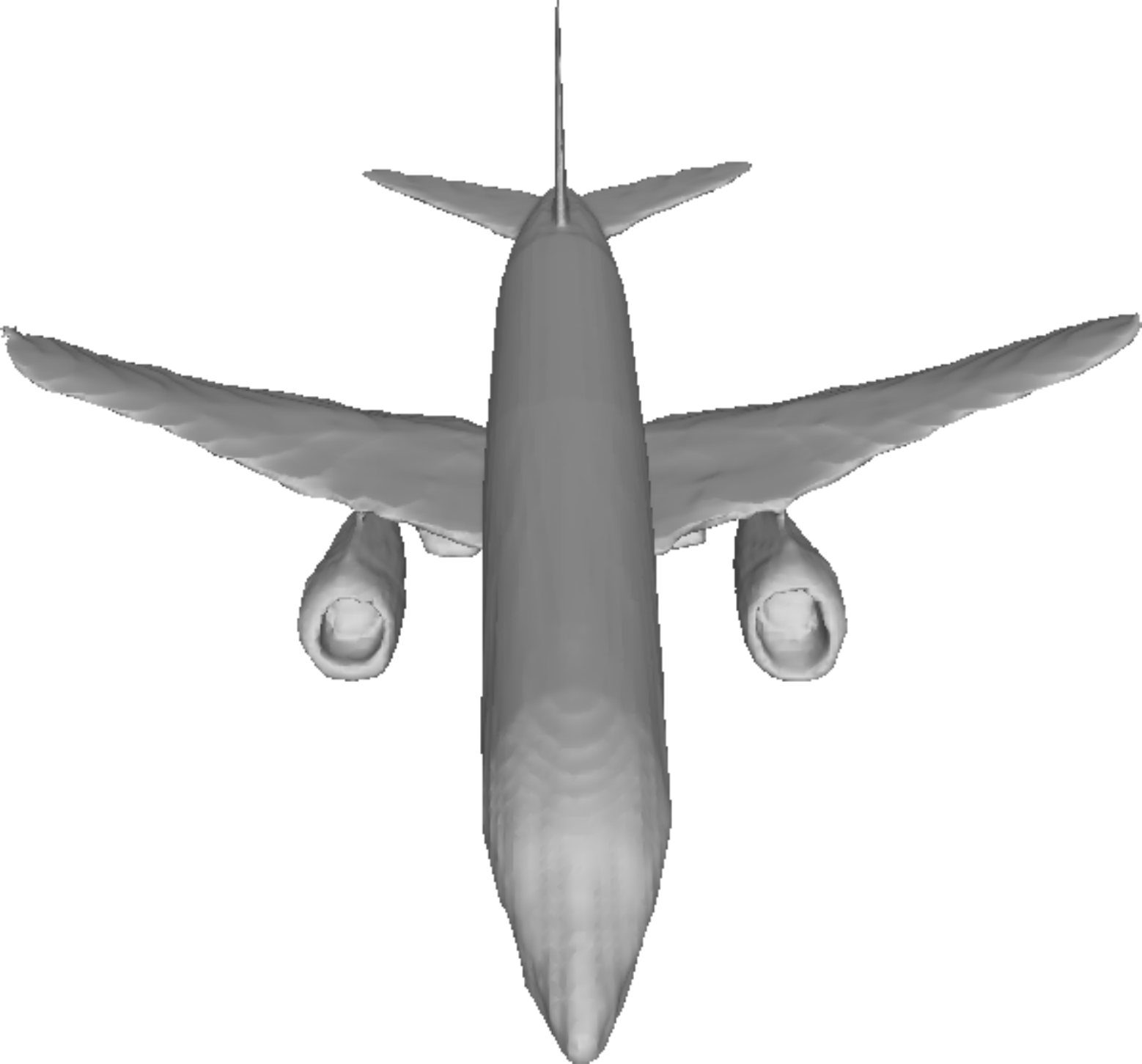}
\caption{Second View (ours)}
\end{subfigure}
\hfill
\begin{subfigure}[t]{0.18\linewidth}
\includegraphics[width=0.9\linewidth]{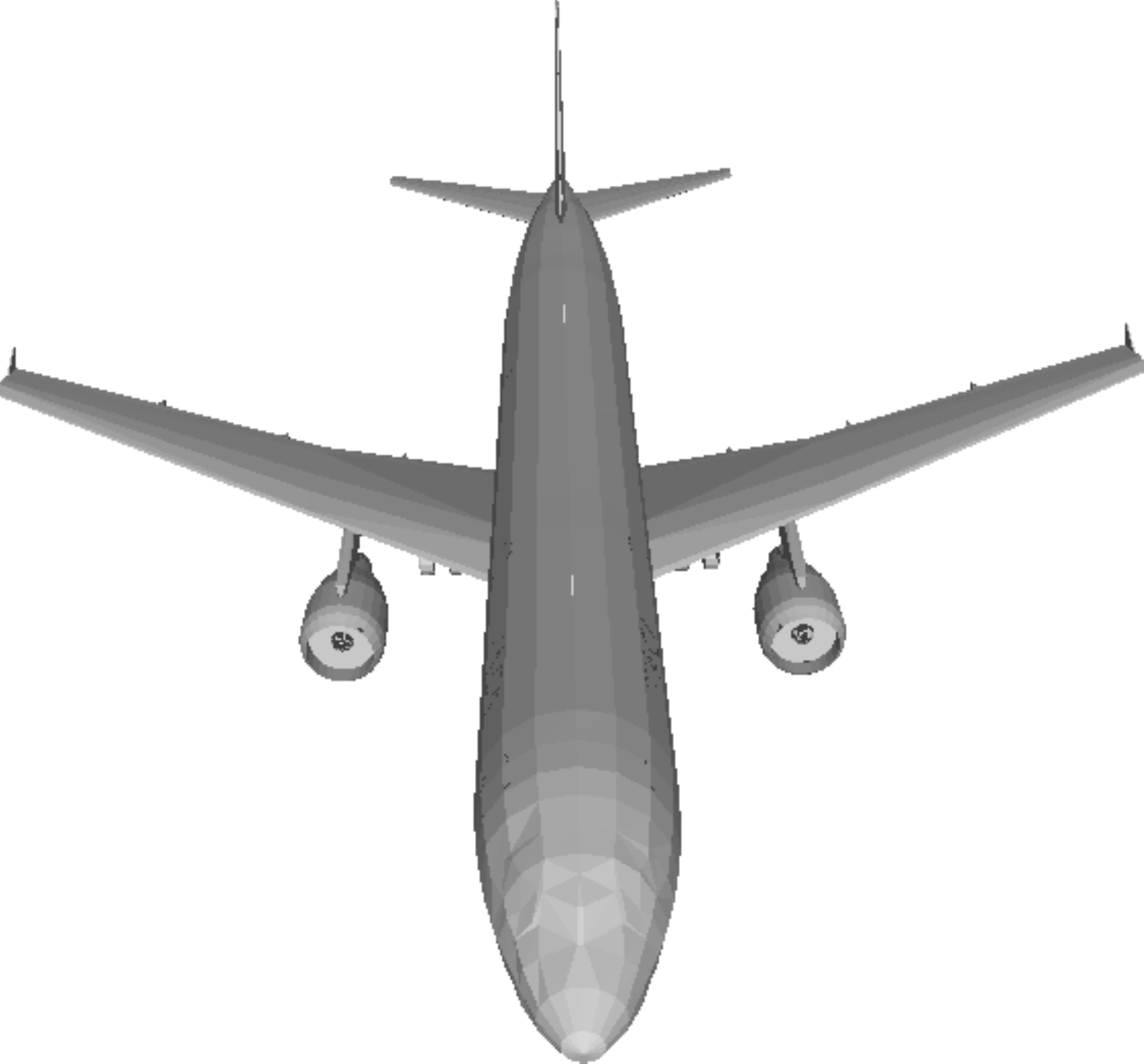}
\caption{Ground truth}
\end{subfigure}
\hfill
\begin{subfigure}[t]{0.167\linewidth}
\includegraphics[width=0.9\linewidth]{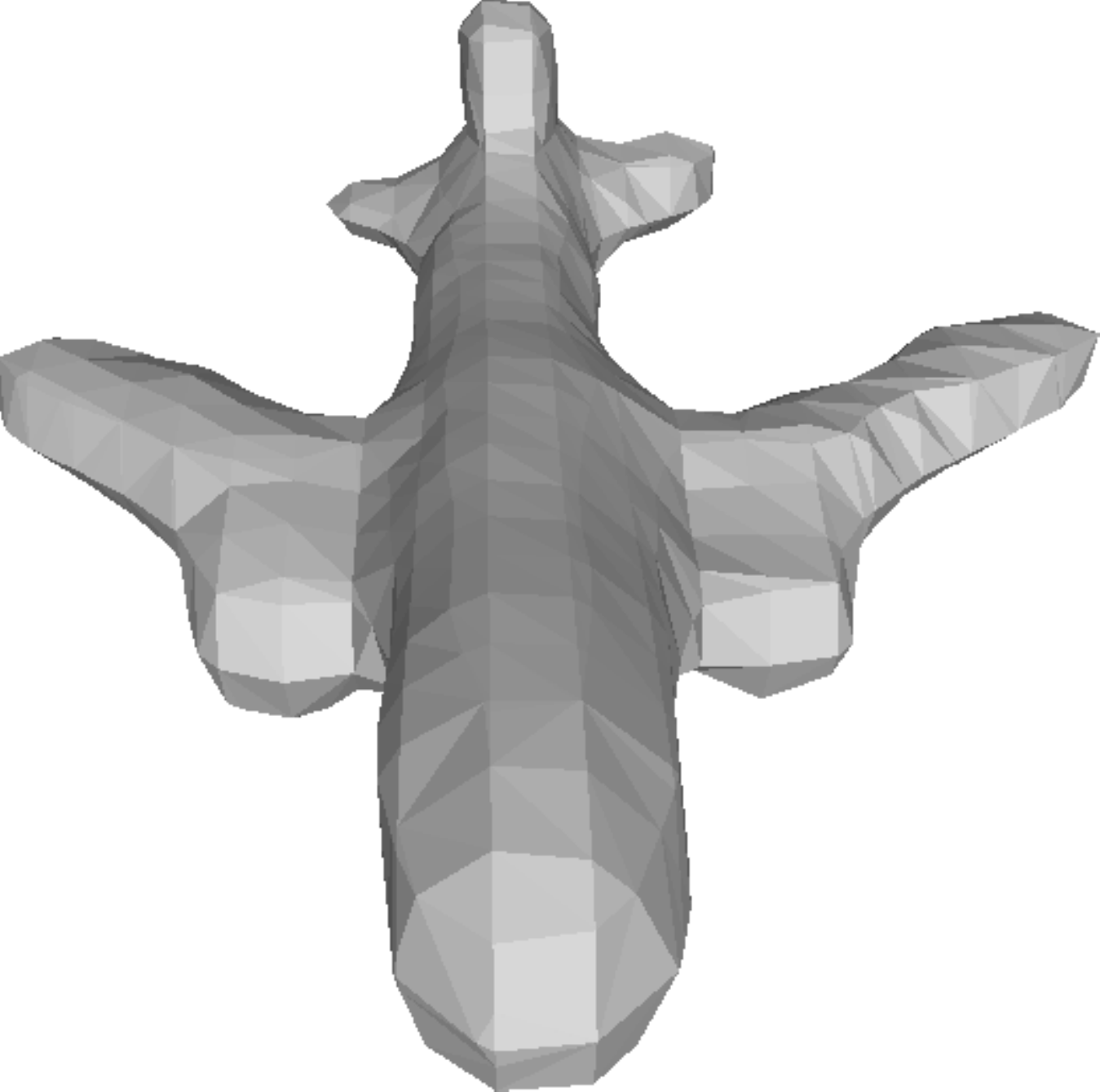}
\caption{3D-EPN}
\end{subfigure}%
	\caption{For a given depth image visualized as a green point cloud, we show a comparison of shape completions from our DeepSDF approach against the true shape and 3D-EPN.}
	\label{fig: completion}
\end{figure*}

\begin{figure} 
\begin{subfigure}[t]{0.49\linewidth}
\includegraphics[width=\linewidth]{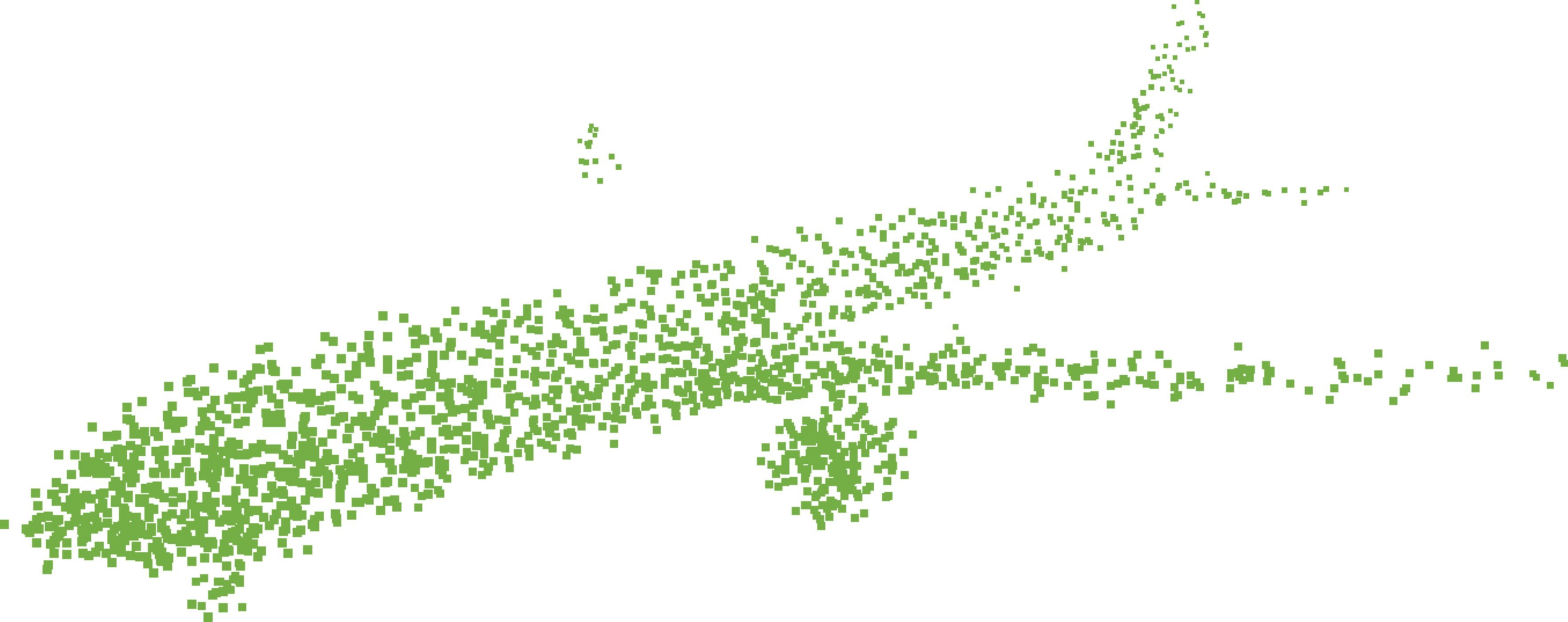}
\caption{Noisy Input Point Cloud}
\end{subfigure}
\hfill
\begin{subfigure}[t]{0.49\linewidth}
\includegraphics[width=\linewidth]{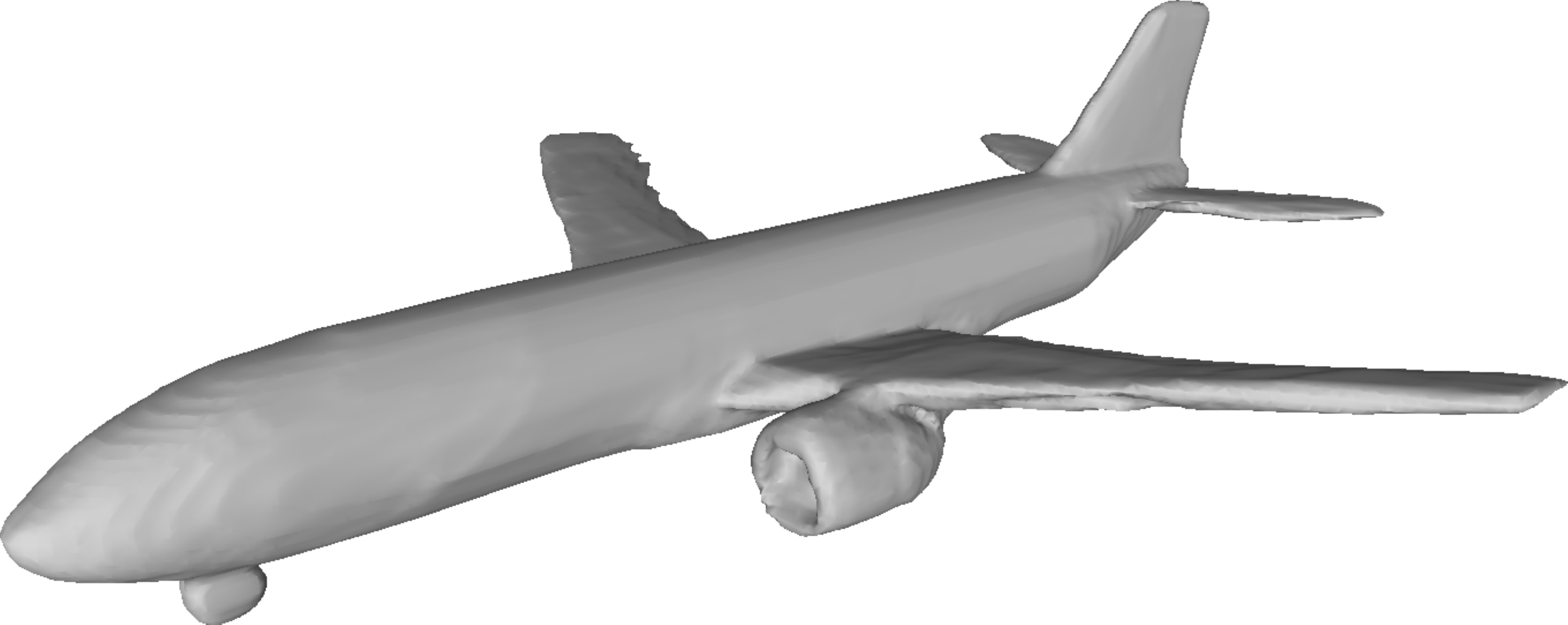}
\caption{Shape Completion}
\end{subfigure}
	\caption{Demonstration of DeepSDF shape completion from a partial noisy point cloud. Input here is generated by perturbing the 3D point cloud positions generated by the ground truth depth map by 3\% of the plane length. We provide a comprehensive analysis of robustness to noise in the supplementary material.}
	\label{fig:noisyCompletion}
\end{figure}

To generate SDF point samples from the depth image observation, we sample two points for each depth observation, each of them located $\eta$ distance away from the measured surface point (along surface normal estimate). With small $\eta$ we approximate the signed distance value of those points to be $\eta$ and $-\eta$, respectively. We solve for Eq.~\ref{eq:testtime} with loss function of Eq.~\ref{eq:clampedL1} using clamp value of $\eta$.
Additionally, we incorporate free-space observations, (i.e. empty-space between surface and camera), by sampling points along the freespace-direction and enforce larger-than-zero constraints.
The freespace loss is $|f_\theta (\bm{z},\bm{x}_j)|$ if $f_\theta (\bm{z},\bm{x}_j)<0$ and $0$ otherwise.

Given the SDF point samples and empty space points, we similarly optimize the latent vector using MAP estimation. Tab.~\ref{tab:5} and Figs.~(\ref{fig: completion}, \ref{fig:noisyCompletion}) respectively shows quantitative and qualitative shape completion results. Compared to one of the most recent completion approaches \cite{dai2017shape} using volumetric shape representation, our continuous SDF approach produces more visually pleasing and accurate shape reconstructions. While a few recent shape completion methods were presented \cite{han2017high, wu2018learning}, we could not find the code to run the comparisons, and their underlying 3D representation is voxel grid which we extensively compare against. 

\subsection{Latent Space Shape Interpolation \label{sec: interp}}
To show that our learned shape embedding is complete and continuous, we render the results of the decoder when a pair of shapes are interpolated in the latent vector space (Fig.~\ref{fig:teaser}). The results suggests that the embedded continuous SDF's are of meaningful shapes and that our representation extracts common interpretable shape features, such as the arms of a chair, that interpolate linearly in the latent space.

\section{Conclusion \& Future Work}

DeepSDF significantly outperforms the applicable benchmarked methods across shape representation and completion tasks and simultaneously addresses the goals of representing complex topologies, closed surfaces, while providing high quality surface normals of the shape. However, while point-wise forward sampling of a shape's SDF is efficient, shape completion (auto-decoding) takes considerably more time during inference due to the need for explicit optimization over the latent vector. We look to increase performance by replacing ADAM optimization with more efficient Gauss-Newton or similar methods that make use of the analytic derivatives of the model.

DeepSDF models enable representation of more complex shapes without discretization errors with significantly less memory than previous state-of-the-art results as shown in Table~\ref{tab:1},  demonstrating an exciting route ahead for 3D shape learning. The clear ability to produce quality latent shape space interpolation opens the door to reconstruction algorithms operating over scenes built up of such efficient encodings. However, DeepSDF currently assumes models are in a canonical pose and as such completion in-the-wild requires explicit optimization over a $SE(3)$ transformation space increasing inference time. Finally, to represent the true \textit{space-of-possible-scenes} including dynamics and textures in a single embedding remains a major challenge, one which we continue to explore.

\appendix

\section*{Supplementary}
\section{Overview}
This supplementary material provides quantitative and qualitative experimental results along with extended technical details that are supplementary to the main paper. We first describe the shape completion experiment with noisy depth maps using DeepSDF (Sec. \ref{sec: noisy}). We then discuss architecture details (Sec. \ref{sec: arch}) along with experiments exploring characteristics and tradeoffs of the DeepSDF design decisions (Sec. \ref{sec: design}). In Sec. \ref{sec. mnist} we compare auto-decoders with variational and standard auto-encoders. Further, additional details on data preparation (Sec.~\ref{sec: data-prep}), training (Sec.~\ref{sec: training-testing}), the auto-decoder learning scheme (Sec. \ref{sec:autodecoder}), and quantitative evaluations (Sec.~\ref{sec: quantitative}) are presented, and finally in Sec.~\ref{sec: additional} we provide additional quantitative and qualitative results. 

\section{Shape Completion from Noisy Depth Maps} \label{sec: noisy}
We test the robustness of our shape completion method by using noisy depth maps as input. Specifically, we demonstrate the ability to complete shapes given partial noisy point clouds obtained from consumer depth cameras. Following \cite{handa2014benchmark}, we simulate the noise distribution of typical structure depth sensors, including Kinect V1 by adding zero-mean Guassian noise to the inverse depth representation of a ground truth input depth image:
\begin{equation}\label{eq: noise}
D_{\text{noise}}=\frac{1}{(1/D) + \mathcal{N}(0,\alpha^2)},
\end{equation}
where $\alpha$ is standard deviation of the normal distribution.

\begin{figure}[t]
\centering
\includegraphics[width=0.53\textwidth]{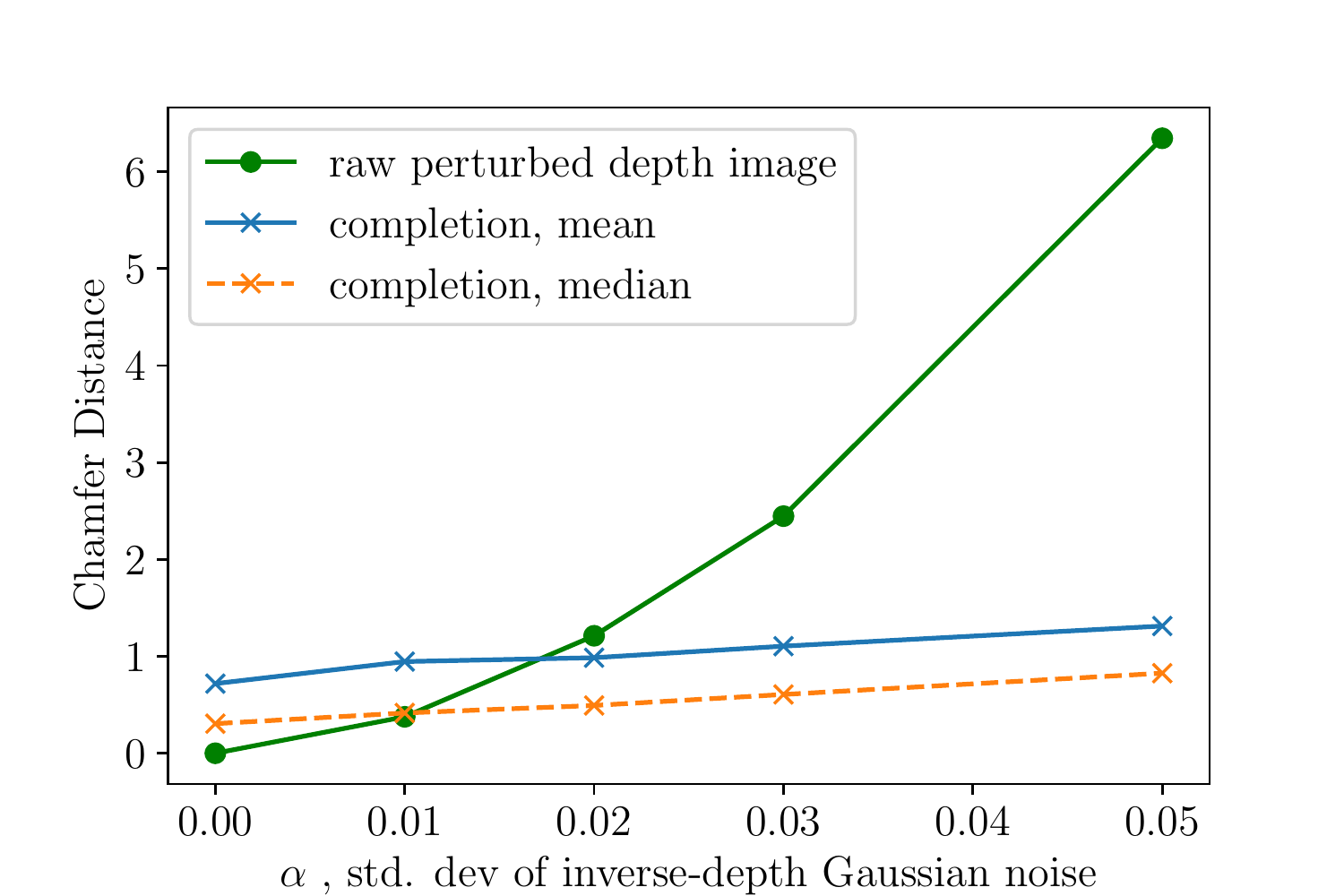}
\caption{Chamfer distance (multiplied by $10^3$) as a function of $\alpha$, the standard deviation of inverse-depth Gaussian noise as shown in Eq.~\ref{eq: noise}, for shape completions on planes from ShapeNet. Green line describes Chamfer distance between the perturbed depth points and original depth points, which shows the superlinear increase with increased noise.  Blue and orange show respectively the mean and median of the shape completion's Chamfer distance (over a dataset of 85 plane completions) relative to the ground truth mesh which deteriorates approximately linearly with increasing standard deviation of noise.  The same DeepSDF model was used for inference, the only difference is in the noise of the single depth image provided from which to perform shape completion.    Example qualitative resuls are shown in Fig.~\ref{fig: noisycompletion}.}
\label{fig: depthmapnoiseplot}
\end{figure}

For the experiment, we synthetically generate noisy depth maps from the ShapeNet \cite{chang2015shapenet} plane models using the same benchmark test set of Dai et al. \cite{dai2017shape} used in the main paper. We perturb the depth values using standard deviation $\alpha$ of $0.01$, $0.02$, $0.03$, and $0.05$. Given that the target shapes are normalized to a unit sphere, one can observe that the inserted noise level is significant  (Fig. \ref{fig: depthmapnoiseplot}).

\begin{figure*} 
\begin{subfigure}[t]{0.195\linewidth}
\includegraphics[width=0.8\linewidth]{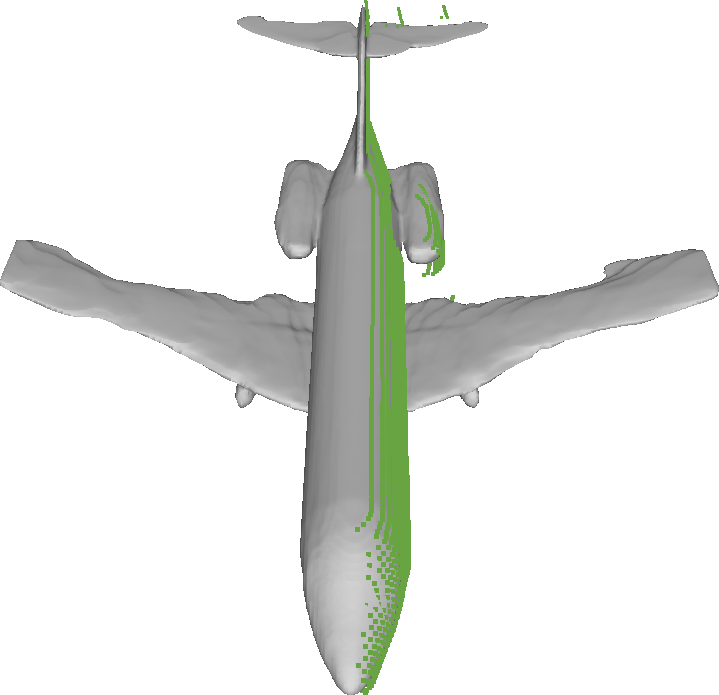}
\caption{No noise}
\end{subfigure}
\hfill
\begin{subfigure}[t]{0.195\linewidth}
\includegraphics[width=0.8\linewidth]{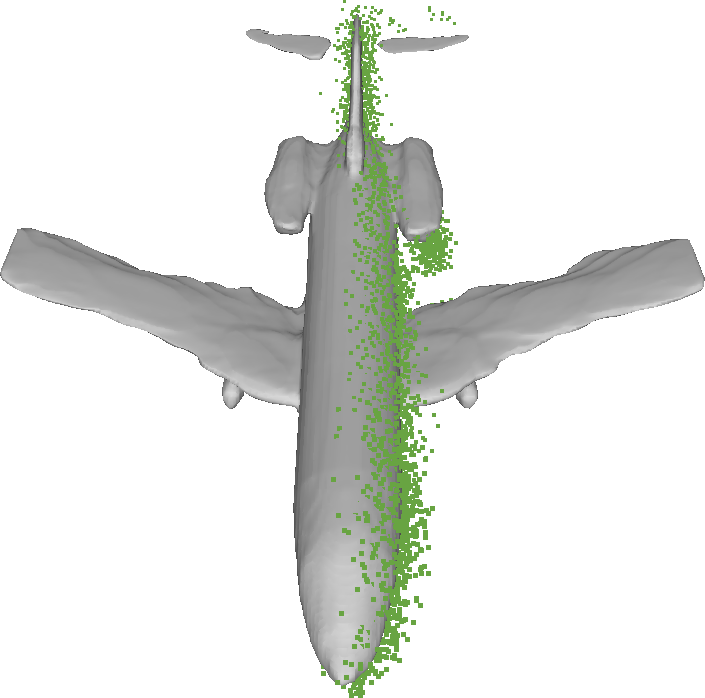}
\caption{$\alpha=0.01$}
\end{subfigure}
\hfill
\begin{subfigure}[t]{0.195\linewidth}
\includegraphics[width=0.8\linewidth]{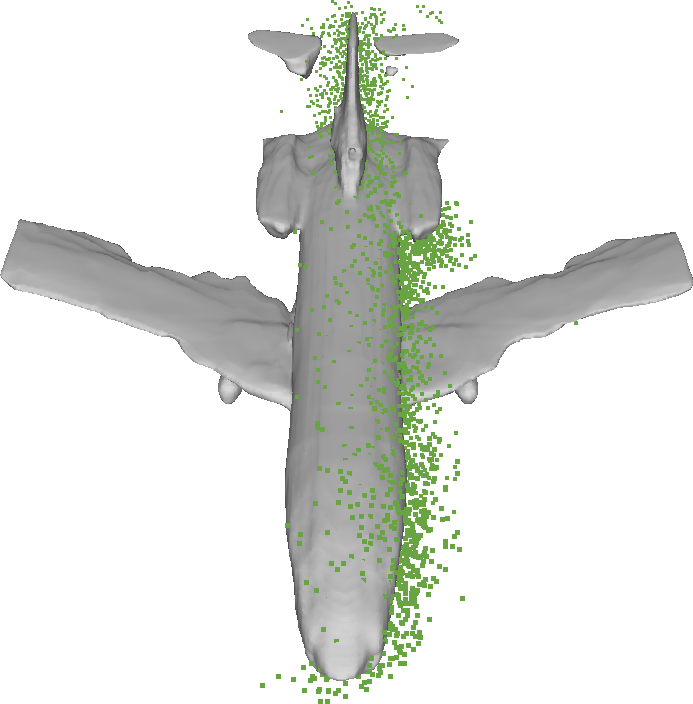}
\caption{$\alpha=0.02$}
\end{subfigure}
\hfill
\begin{subfigure}[t]{0.195\linewidth}
\includegraphics[width=0.8\linewidth]{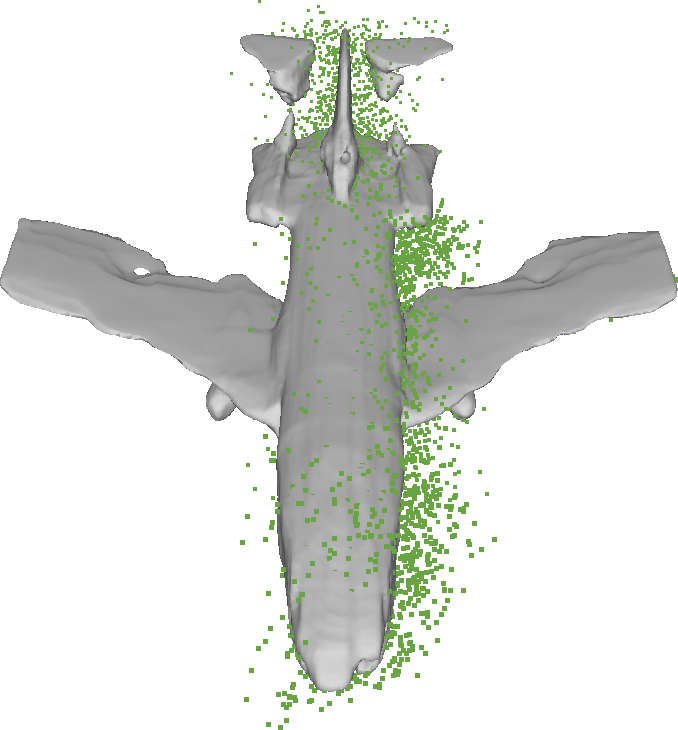}
\caption{$\alpha=0.03$}
\end{subfigure}
\hfill
\begin{subfigure}[t]{0.18\linewidth}
\includegraphics[width=0.8\linewidth]{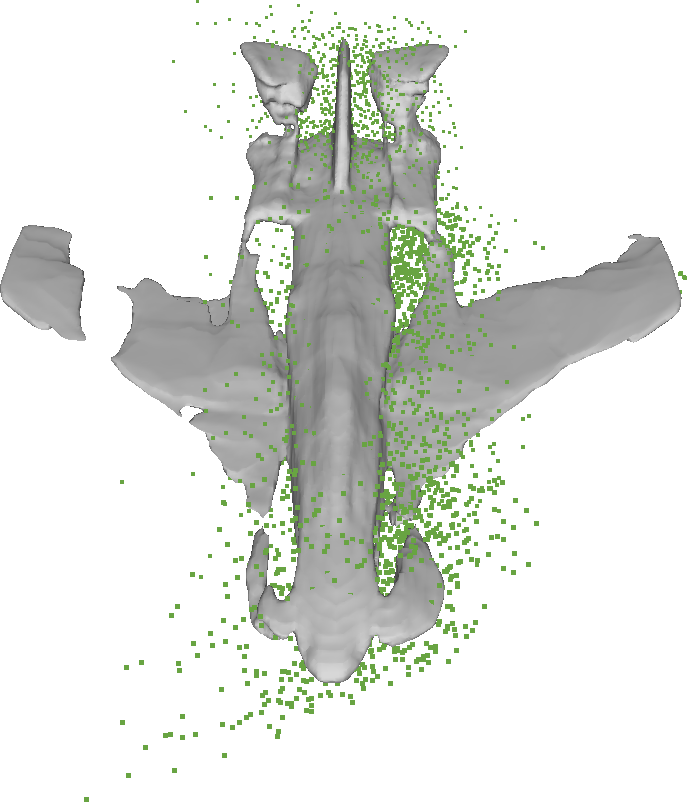}
\caption{$\alpha=0.05$}
\end{subfigure}
\hfill
	\caption{Shape completion results obtained from the partial and noisy input depth maps shown below. Input point clouds are overlaid on each completion to illustrate the scale of noise in the input. } \label{fig: noisycompletion}

\hfill
\begin{subfigure}[t]{0.17\linewidth}
\includegraphics[width=0.8\linewidth]{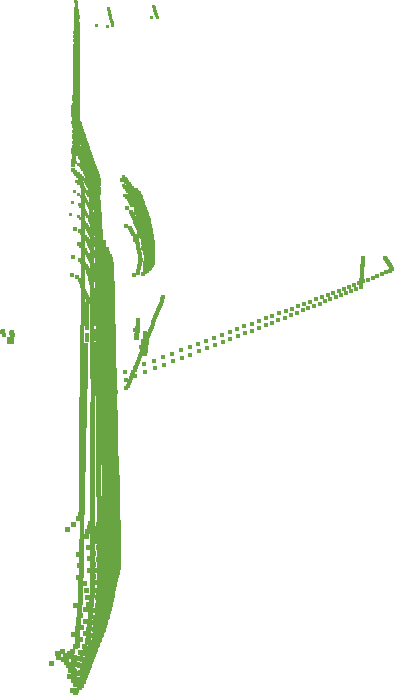}
\caption{No noise}
\end{subfigure}
\begin{subfigure}[t]{0.18\linewidth}
\includegraphics[width=0.8\linewidth]{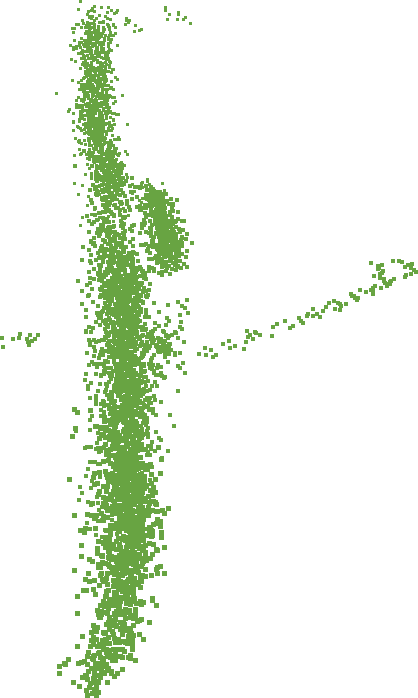}
\caption{$\alpha=0.01$}
\end{subfigure}
\begin{subfigure}[t]{0.19\linewidth}
\includegraphics[width=0.8\linewidth]{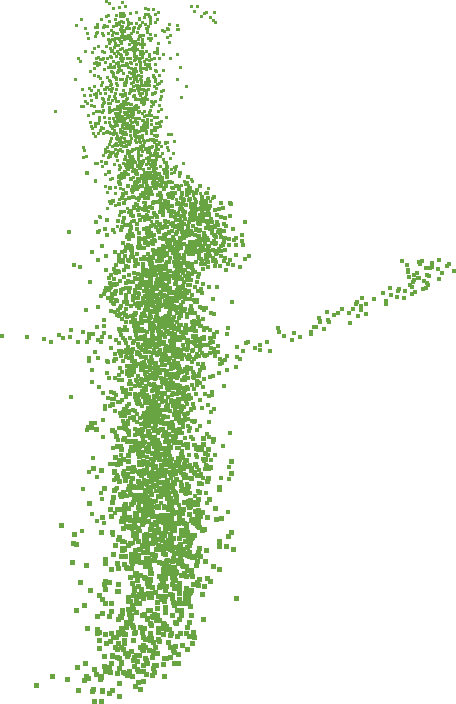}
\caption{$\alpha=0.02$}
\end{subfigure}
\begin{subfigure}[t]{0.18\linewidth}
\includegraphics[width=0.8\linewidth]{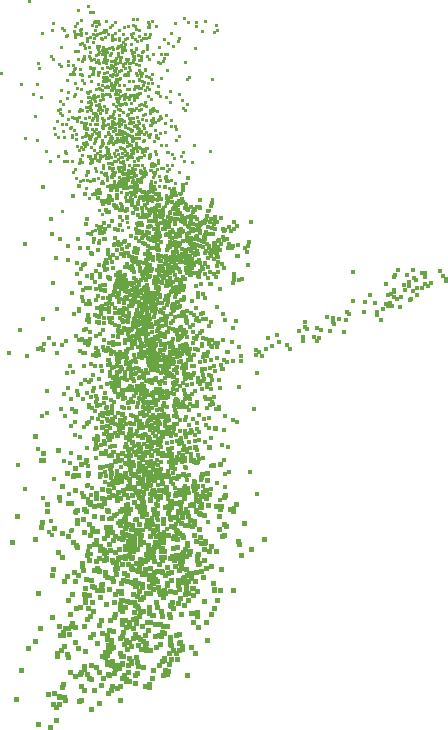}
\caption{$\alpha=0.03$}
\end{subfigure}
\begin{subfigure}[t]{0.22\linewidth}
\includegraphics[width=0.8\linewidth]{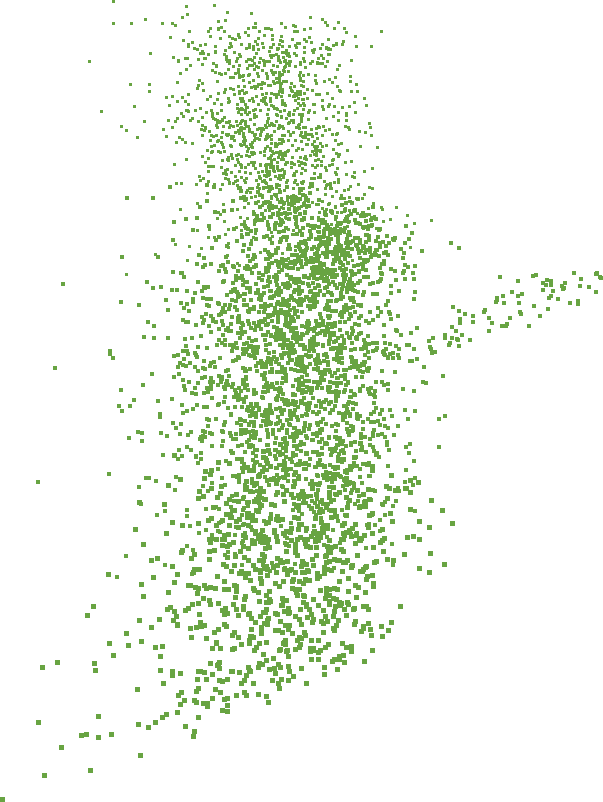}
\caption{$\alpha=0.05$}
\end{subfigure}

	\caption{Visualization of partial and noisy point-clouds used to test shape completion with \textit{DeepSDF}. Here, $\alpha$ is the standard deviation of Gaussian noise in Eq. \ref{eq: noise}. Corresponding completion results are shown above.}
	\label{fig: noise}
\end{figure*}

The shape completion results with respect to added Guassian noise on the input synthetic depth maps are shown in Fig.~\ref{eq: noise}.  The Chamfer distance of the inferred shape versus the ground truth shape deteriorates approximately linearly with increasing standard deviation of the noise. Compared to the Chamfer distance between raw perturbed point cloud and ground truth depth map, which increases super-linearly with increasing noise level (Fig. \ref{eq: noise}), the shape completion quality using DeepSDF degrades much slower, implying that the shape priors encoded in the network play an important role regularizing the shape reconstruction.

\section{Network Architecture} \label{sec: arch}
\begin{figure*} 
\centering
\includegraphics[width=\linewidth]{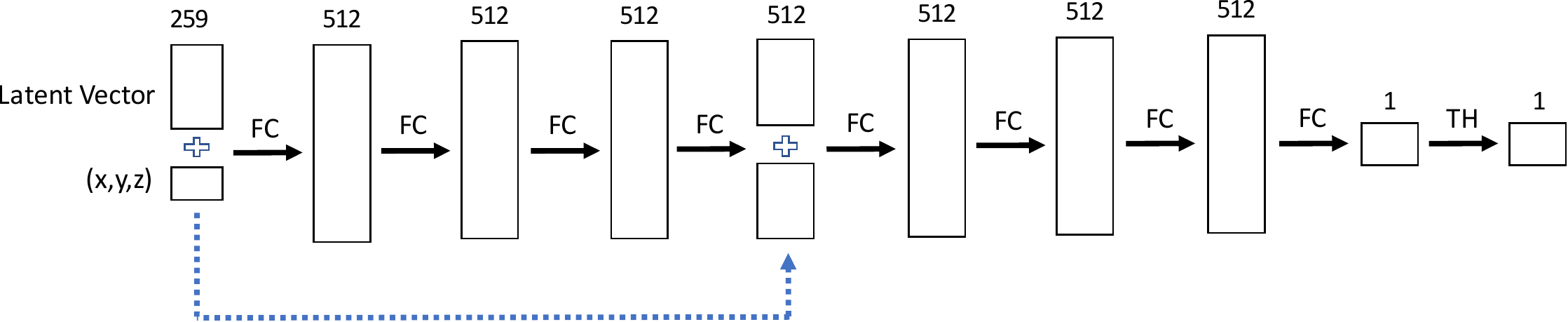}
\caption{DeepSDF architecture used for experiments. Boxes represent vectors while arrows represent operations. The feed-forward network is composed of 8 fully connected layers, denoted as ``FC'' on the diagram. We used 256 and 128 dimensional latent vectors for reconstruction and shape completion experiments, respectively. The latent vector is concatenated, denoted ``+'', with the xyz query, making 259 length vector, and is given as input to the first layer. We find that inserting the latent vector again to the middle layers significantly improves the learning, denoted as dotted arrow in the diagram:  the 259 vector is concatenated with the output of fourth fully connected layer to make a 512 vector. Final SDF value is obtained with hypberbolic tangent non-linear activation denoted as ``TH''.}
\label{fig: architecture}
\end{figure*}

Fig. \ref{fig: architecture} depicts the overall architecture of DeepSDF. For all experiments in the main paper we used a network composed of 8 fully connected layers each of which are applied with weight-normalization, and each intermediate vectors are processed with RELU activation and 0.2 dropout except for the final layer.  A skip connection is included at the fourth layer.

\section{DeepSDF Network Design Decisions}  \label{sec: design}
In this section, we study system parameter decisions that affect the accuracy of SDF regression, thereby providing insight on the tradeoffs and scalability of the proposed algorithm.

\subsection{Effect of Network Depth on Regression Accuracy} \label{sec: capacity}
In this experiment we test how the expressive capability of DeepSDF varies as a function of the number of layers. Theoretically, an infinitely deep feed-forward network should be able to memorize the training data with arbitrary precision, but in practice this is not true due to finite compute power and the vanishing gradient problem, which limits the depth of the network. 

We conduct an experiment where we let DeepSDF memorize SDFs of 500 chairs and inspect the training loss with varying number of layers. As described in Fig. \ref{fig: architecture}, we find that applying the input vector (latent vector + xyz query) both to the first and a middle layer improves training. Inspired by this, we split the experiment into two cases: 1) train a regular network without skip connections, 2) train a network by concatenating the input vector to every 4 layers (e.g. for 12 layer network the input vector will be concatenated to the 4th, and 8th intermediate feature vectors).

\begin{figure}[t]
\centering
\includegraphics[width=\linewidth]{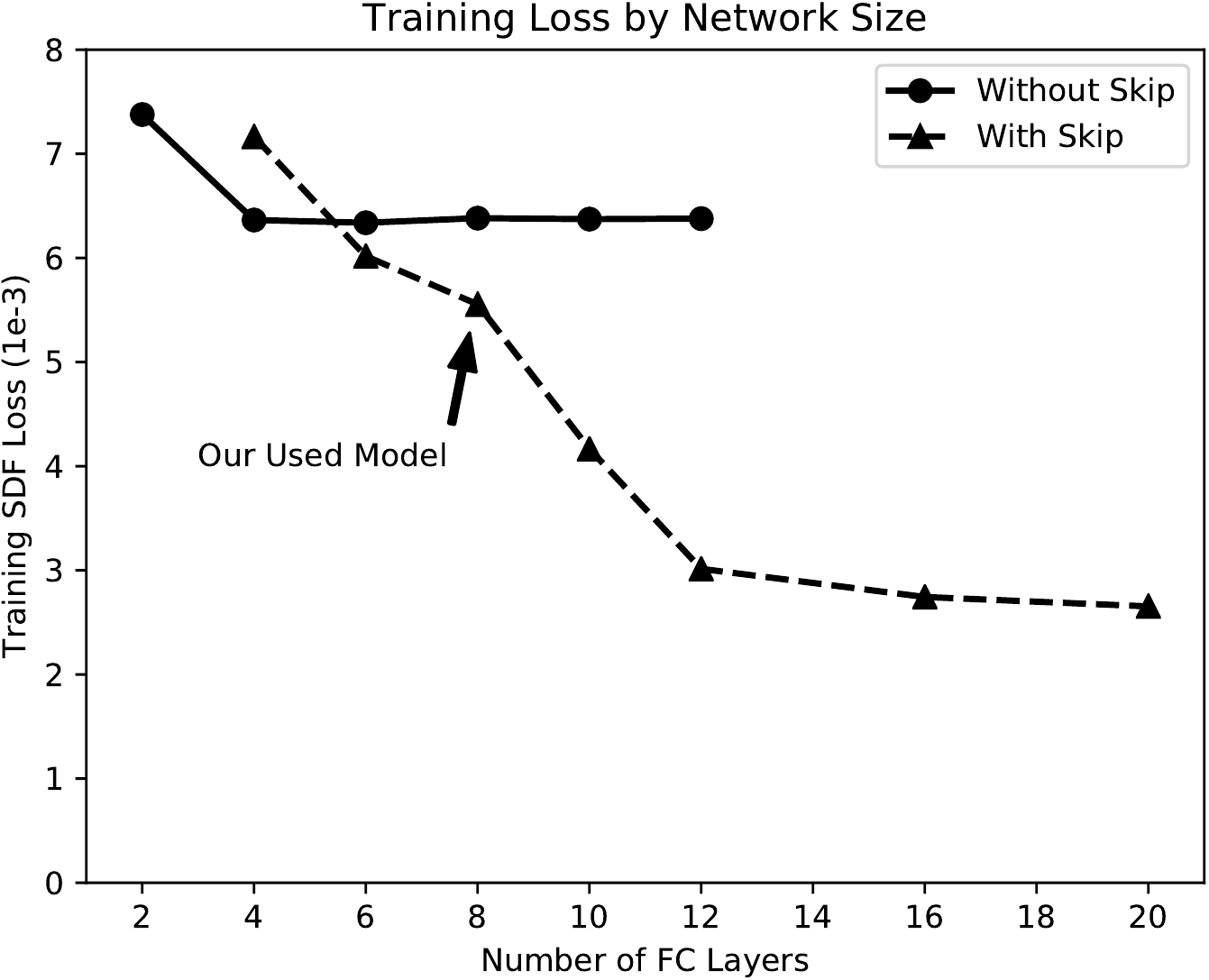}
\caption{Regression accuracy (measured by the SDF loss used in training) as a function of network depth.  Without skip connections, we observe a plateau in training loss past 4 layers.  With skip connections, training loss continues to decrease although with diminishing returns past 12 layers.  The model size chosen for all other experiments, 8 layers, provides a good tradeoff between speed and accuracy.}
\label{fig: capacity}
\end{figure}

Experiment results in Fig. \ref{fig: capacity} shows that the DeepSDF architecture without skip connections gets quickly saturated at 4 layers while the error keeps decreasing when trained with latent vector skip connections. Compared to the architecture we used for the main experiments (8 FC layers), a  network with 16 layers produces significantly smaller training error, suggesting a possibility of using a deeper network for higher precision in some application scenarios. Further, we observe that the test error quickly decrease from four-layer architecture (9.7) to eight layer one (5.7) and subsequently plateaued for deeper architectures.  However, this does not suggest conclusive results on generalization, as we used the same number of small training data for all architectures even though a network with more number of  parameters tends to require higher volume of data to avoid overfitting.

\subsection{Effect of Truncation Distance on Regression Accuracy}
\begin{figure}[t]
\centering
\includegraphics[width=0.53\textwidth]{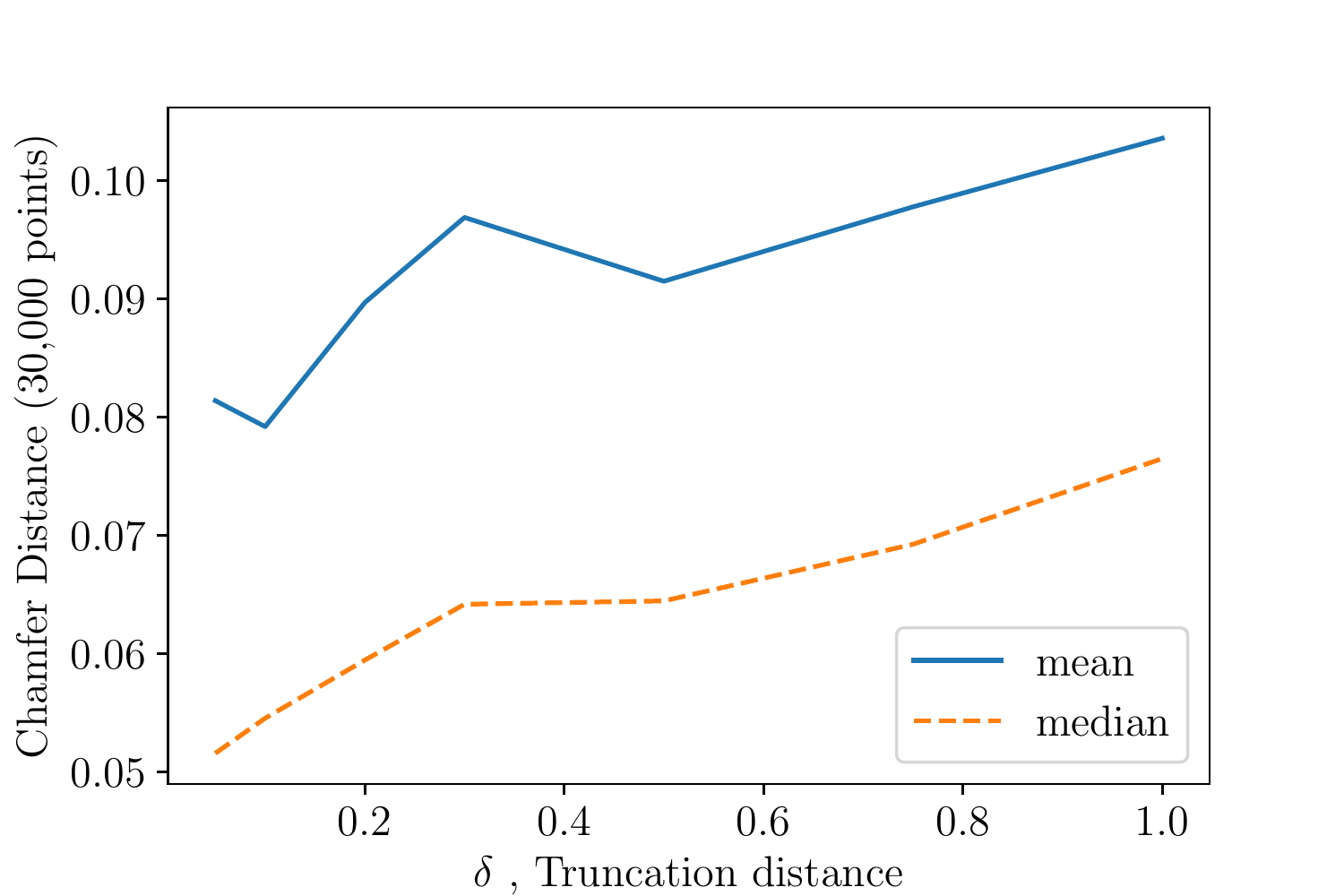}
\caption{Chamfer distance (multiplied by $10^3$) as a function of $\delta$, the truncation distance, for representing a known  small dataset of 100 cars from ShapeNet dataset \cite{chang2015shapenet}.  All models were trained on the same set of SDF samples from these 100 cars.  There is a moderate reduction in the accuracy of the surface, as measured by the increasing Chamfer distance, as the truncation distance is increased between 0.05 and 1.0.  The bend in the curve at $\delta=0.3$ is just expected to be due to the stochasticity inherent in training.  Note that (a) due to the $\tanh()$ activation in the final layer, 1.0 is the maximum value the model can predict, and (b) the plot is dependent on the distribution of the samples used during training.}
\label{fig: truncation}
\end{figure}

We study the effect of the truncation distance ($\delta$ from Eq.~4 of the manuscript) on the regression accuracy of the model.  The truncation distance controls the extent from the surface over which we expect the network to learn a metric SDF.  Fig.~\ref{fig: truncation} plots the Chamfer distance as a function of truncation distance.  We observe a moderate decrease in the accuracy of the surface representation as the truncation distance is increased.  A hypothesis for an explanation is that it becomes more difficult to approximate a larger truncation region (a strictly larger domain of the function) to the same absolute accuracy as a smaller truncation region.  The benefit, however, of larger truncation regions is that there is a larger region over which the metric SDF is maintained -- in our application this reduces raycasting time, and there are other applications as well, such as physics simulation and robot motion planning for which a larger SDF of shapes may be valuable.  We chose a $\delta$ value of 0.01 for all experiments presented in the manuscript, which provides a good tradeoff between raycasting speed and surface accuracy.

\section{Comparison with Variational and Standard Auto-encoders on MNIST} \label{sec. mnist}

\begin{figure*}
    \centering
    \begin{subfigure}[t]{0.32\linewidth}
    \includegraphics[width=\linewidth,clip,trim=118 118 118 120]{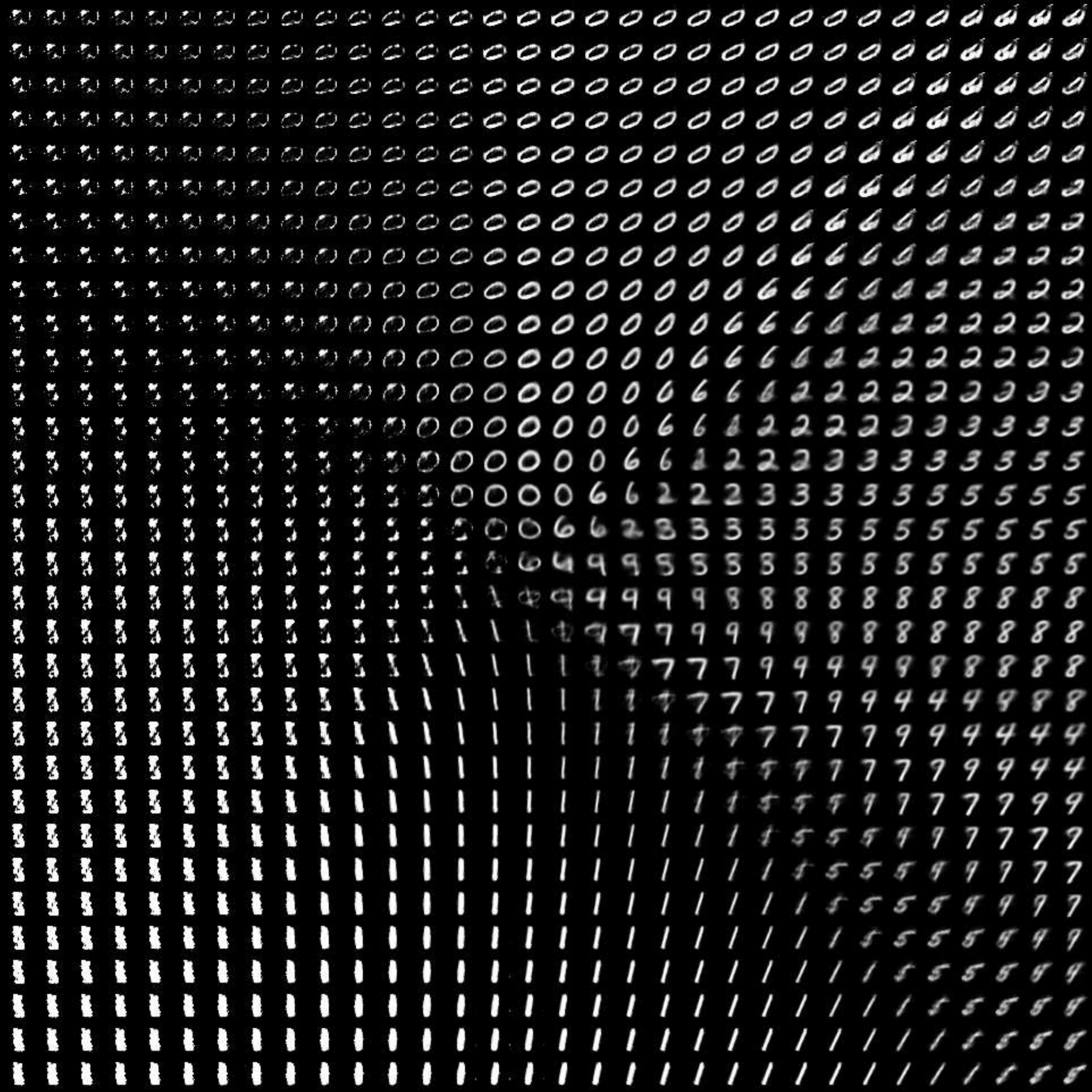}
    \caption{Auto Encoder (AE)}
    \end{subfigure}
    \begin{subfigure}[t]{0.32\linewidth}
    \includegraphics[width=\linewidth,clip,trim=118 118 118 120]{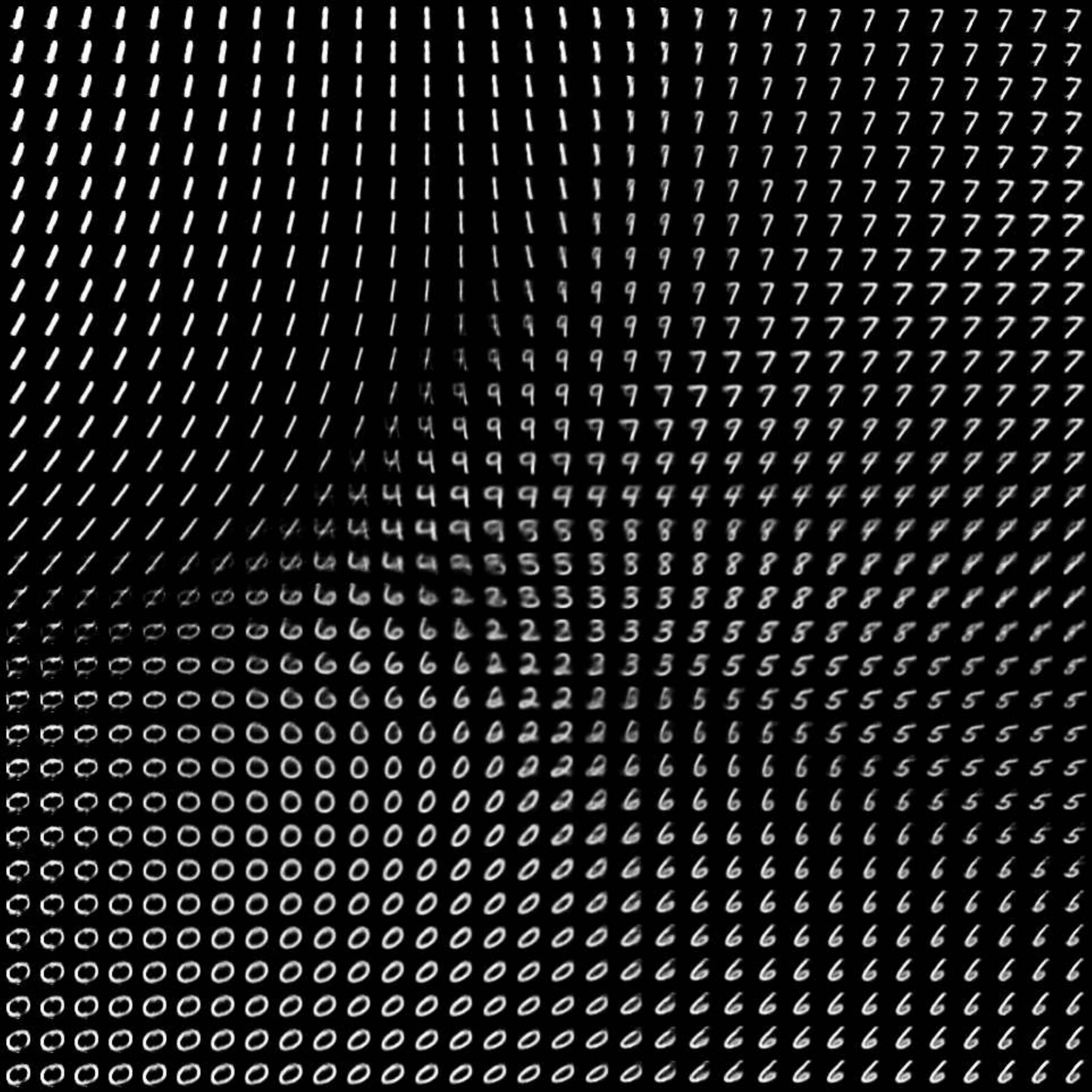}
    \caption{Variational Auto Encoder (VAE)}
    \end{subfigure}
    \begin{subfigure}[t]{0.32\linewidth}
    \includegraphics[width=\linewidth,clip,trim=118 118 118 120]{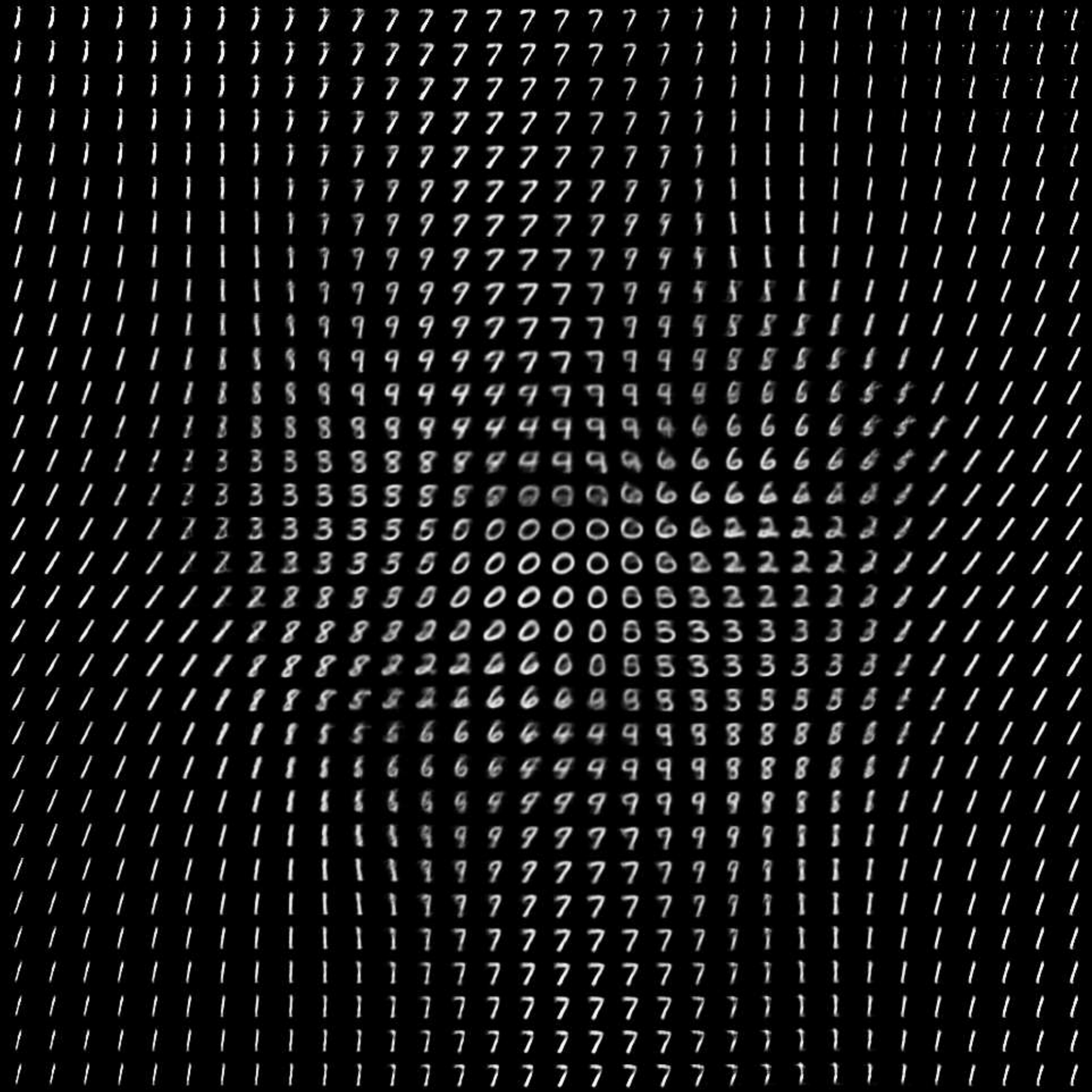}
    \caption{Auto Decoder (AD)}
    \end{subfigure}
    \caption{Comparison of the 2D latent code-space learned by the different methods. Note that large portion of the regular auto-encoder's (AE) latent embedding space contains images that do not look like digits. In contrast, both VAE and AD generate smooth and complete latent space without outstanding artifacts. Best viewed digitally.}
    \label{fig:codeSpaceComparison}
\end{figure*}

\begin{figure}
    \centering
    \hfill
    \begin{subfigure}[t]{0.32\linewidth}
    \includegraphics[width=\linewidth,clip,trim=0 0 0 122]{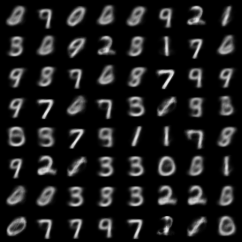}

    \includegraphics[width=\linewidth,clip,trim=0 0 0 122]{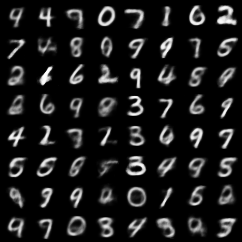}

    \includegraphics[width=\linewidth,clip,trim=0 0 0 122]{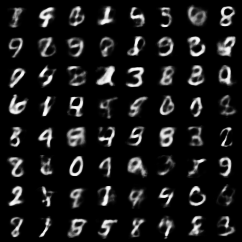}
    \caption{AD}
    \end{subfigure}
    \hfill
    \begin{subfigure}[t]{0.32\linewidth}
    \includegraphics[width=\linewidth,clip,trim=0 0 0 122]{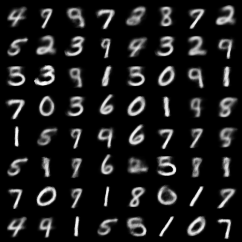}
    
    \includegraphics[width=\linewidth,clip,trim=0 0 0 122]{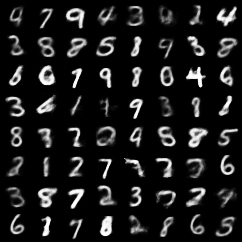}

    \includegraphics[width=\linewidth,clip,trim=0 0 0 122]{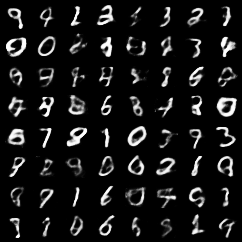}
    \caption{VAE}
    \end{subfigure}
    \hfill
    \begin{subfigure}[t]{0.32\linewidth}
    \includegraphics[width=\linewidth,clip,trim=0 0 0 122]{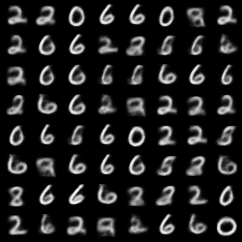}

    \includegraphics[width=\linewidth,clip,trim=0 0 0 122]{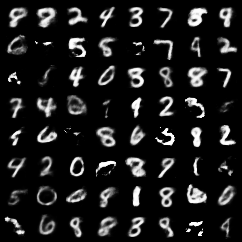}

    \includegraphics[width=\linewidth,clip,trim=0 0 0 122]{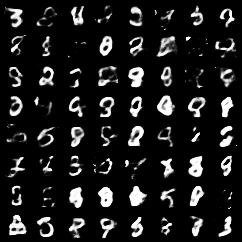}
    \caption{AE}
    \end{subfigure}
        \hfill
    \caption{Visualization of random samples from the latent 2D (top), 5D (middle), and 15D (bottom) code space on MNIST. Note that the sampling from regular auto-encoder (AE) suffers from artifacts. Best viewed digitally.}
    \label{fig:MNISTsamples}
\end{figure}

To compare different approaches of learning a latent code-space for a given datum, we use the MNIST dataset and compare the variational auto encoder (VAE), the standard bottleneck auto encoder (AE), and the proposed auto decoder (AD). As the reconstruction error we use the standard binary cross-entropy and match the model architectures such that the decoders of the different approaches have exactly the same structure and hence theoretical capacity. We show all evaluations for different latent code-space dimensions of 2D, 5D and 15D.

For 2D codes the latent spaces learned by the different methods are visualized in Fig.~\ref{fig:codeSpaceComparison}. All code spaces can reasonably represent the different digits. The AD latent space seems more condensed than the ones from VAE and AE. 
For the optimization-based encoding approach we initialize codes randomly. We show visualizations of such random samples in  Fig.~\ref{fig:MNISTsamples}. Note that samples from the AD- and VAE-learned latent code spaces mostly look like real digits, showing their ability to generate realistic digit images.

\begin{figure*}
    \centering
    \begin{subfigure}[t]{0.32\linewidth}
    \includegraphics[width=\linewidth]{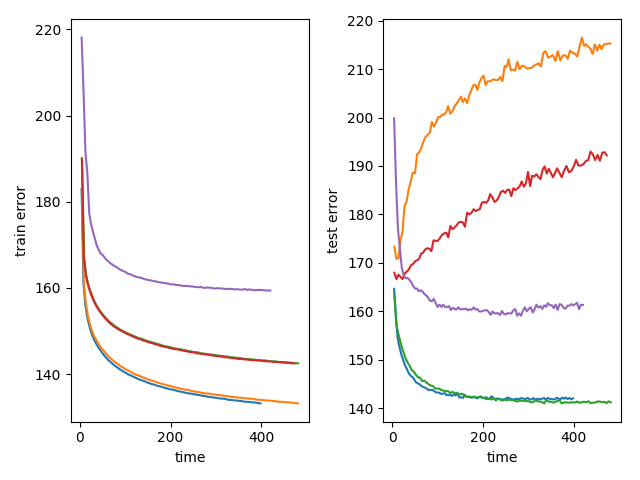}
    \caption{2D}
    \end{subfigure}
    \begin{subfigure}[t]{0.32\linewidth}
    \includegraphics[width=\linewidth]{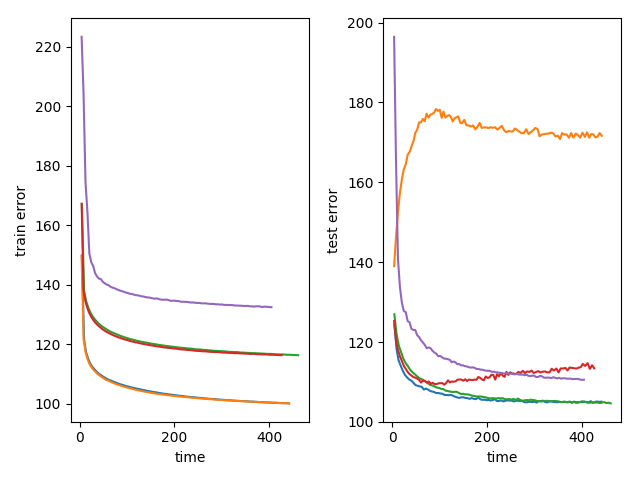}
    \caption{5D}
    \end{subfigure}
    \begin{subfigure}[t]{0.32\linewidth}
    \includegraphics[width=\linewidth]{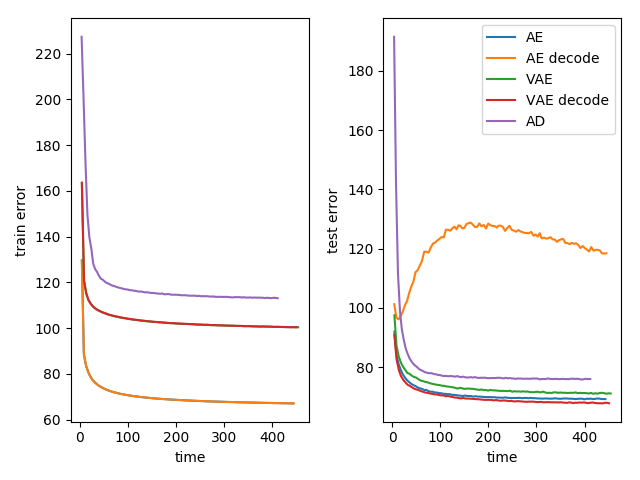}
    \caption{15D}
    \end{subfigure}
    \caption{Train and test error for different dimensions of the latent code for the different approaches.}
    \label{fig:errorComparison}
\end{figure*}

We also compare the train and test reconstruction errors for the different methods in Fig.~\ref{fig:errorComparison}. For VAE and AE we show both the reconstruction error obtained using the learned encoder and obtained via code optimization using the learned decoder only (denoted ``(V)AE decode''). 
The test error for VAE and AE are consistently minimized for all latent code dimensions. ``AE decode '' diverges in all cases hinting at a learned latent space that is poorly suited for optimization-based decoding. Optimizing latent codes using the VAE encoder seems to work better for higher dimensional codes. 
The proposed AD approach works well in all tested code space dimensions. Although ``VAE decode'' has slightly lower test error than AD in 15 dimensions, qualitatively the AD's reconstructions are better as we discuss next.  

\begin{figure*}
    \centering
    \begin{subfigure}[t]{0.16\linewidth}
    \includegraphics[width=\linewidth,clip,trim=0 0 0 29]{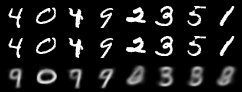}
    
    \includegraphics[width=\linewidth,clip,trim=0 0 0 29]{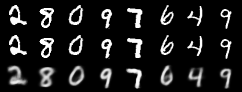}
    
    \includegraphics[width=\linewidth,clip,trim=0 0 0 29]{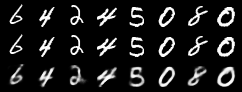}
    \caption{AD}
    \end{subfigure}
    \begin{subfigure}[t]{0.16\linewidth}
    \includegraphics[width=\linewidth]{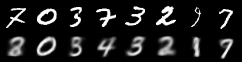}

    \includegraphics[width=\linewidth]{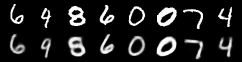}
    
    \includegraphics[width=\linewidth]{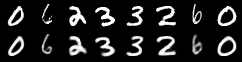}
    \caption{VAE}
    \end{subfigure}
    \begin{subfigure}[t]{0.16\linewidth}
    \includegraphics[width=\linewidth]{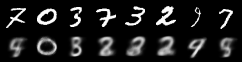}
    
    \includegraphics[width=\linewidth]{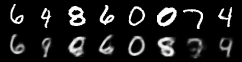}
    
    \includegraphics[width=\linewidth]{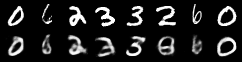}
    \caption{VAE decode}
    \end{subfigure}
    \begin{subfigure}[t]{0.16\linewidth}
    \includegraphics[width=\linewidth]{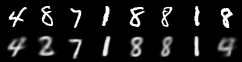}
    
    \includegraphics[width=\linewidth]{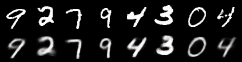}
    
    \includegraphics[width=\linewidth]{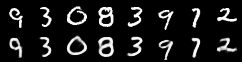}
    \caption{AE}
    \end{subfigure}
    \begin{subfigure}[t]{0.16\linewidth}
    \includegraphics[width=\linewidth]{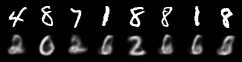}

    \includegraphics[width=\linewidth]{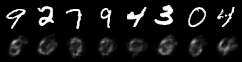}

    \includegraphics[width=\linewidth]{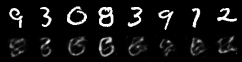}
    \caption{AE decode}
    \end{subfigure}
    \caption{Reconstructions for 2D (top), 5D (middle), and 15D (bottom) code space on MNIST. For each of the different dimensions we plot the given test MNIST image and the reconstruction given the inferred latent code.}
    \label{fig:MNISTreconstructions}
\end{figure*}

In Fig.~\ref{fig:MNISTreconstructions} we show example reconstructions from the test dataset. When using the learned encoders VAE and AE produce qualitatively good reconstructions. When using optimization-based encoding ``AE decode'' performs poorly indicating that the latent space has many bad local minima. While the reconstructions from ``VAE decode`` are, for the most part, qualitatively close to the original, AD's reconstructions more closely resemble the actual digit of the test data. Qualitatively, AD is on par with reconstructions from end-to-end-trained VAE and AE. 

\section{Data Preparation Details}\label{sec: data-prep}
For data preparation, we are given a mesh of a shape to sample spatial points and their SDF values. We begin by normalizing each shape so that the shape model fits into a unit sphere with some margin (in practice fit to sphere radius of 1/1.03). Then, we virtually render the mesh from 100 virtual cameras regularly sampled on the surface of the unit sphere. Then, we gather the surface points by back-projecting the depth pixels from the virtual renderings, and the points' normals are assigned from the triangle to which it belongs. Triangle surface orientations are set such that they are towards the camera. When a triangle is visible from both orientations, however, the given mesh is not watertight, making true SDF values hard to calculate, so we discard a mesh with more than 2\% of its triangles being double-sided. For a valid mesh, we construct a KD-tree for the oriented surface points.

As stated in the main paper, it is important that we sample more aggressively near the surface of the mesh as we want to accurately model the zero-crossings. Specifically, we sample around 250,000 points randomly on the surface of the mesh, weighted by triangle areas. Then, we perturb each surface point along all xyz axes with mean-zero Gaussian noise with variance 0.0025 and 0.00025 to generate two spatial samples per surface point. For around 25,000 points we uniformly sample within the unit sphere. For each collected spatial samples, we find the nearest surface point from the KD-tree, measure the distance, and decide the sign from the dot product between the normal and their vector difference.

\section{Training and Testing Details}\label{sec: training-testing}

For training, we find it is important to initialize the latent vectors quite small, so that similar shapes do not diverge in the latent vector space -- we used $\mathcal{N}(0,0.01^2)$. Another crucial point is balancing the positive and negative samples both for training and testing: for each batch used for gradient descent, we set half of the SDF point samples positive and the other half negative.

Learning rate for the decoder parameters was set to be 1e-5 * $B$, where $B$ is number of shapes in one batch. For each shape in a batch we subsampled 16384 SDF samples. Learning rate for the latent vectors was set to be 1e-3. Also, we set the regularization parameter $\sigma=10^{-2}$. We trained our models on 8 Nvidia GPUs approximately for 8 hours for 1000 epochs. For reconstruction experiments the latent vector size was set to be 256, and for the shape completion task we used models with 128 dimensional latent vectors.

\section{Full Derivation of Auto-decoder-based DeepSDF Formulation} \label{sec:autodecoder}
To derive the auto-decoder-based shape-coded DeepSDF formulation we adopt a probabilistic perspective. 
Given a dataset of $N$ shapes represented with signed distance function ${SDF^i}_{i=1}^{N}$, we prepare a set of $K$ point samples and their signed distance values:
\begin{equation} \label{eq:datapoint}
X_i = \{(\bm{x}_j,s_j):s_j=SDF^i(\bm{x}_j)\}\,.
\end{equation}
The SDF values can be computed from mesh inputs as detailed in the main paper.

For an auto-decoder, as there is no encoder, each latent code $\bm{z}_i$ is paired with training shape data $X_i$ and randomly initialized from a zero-mean Gaussian. We use $\mathcal{N}(0,0.001^2)$.
The latent vectors $\{\bm{z}_i\}_{i=1}^N$ are then jointly optimized during training along with the decoder parameters $\theta$.

We assume that each shape in the given dataset $\bm{X}=\{X_i\}_{i=1}^N$ follows the joint distribution of shapes:
\begin{equation}
p_{\theta}(X_i,\bm{z}_i)=p_{\theta}(X_i|\bm{z}_i)p(\bm{z}_i)\,,
\end{equation}
where $\theta$ parameterizes the data likelihood. For a given $\theta$ a shape code $\bm{z}_i$ can be estimated via Maximum-a-Posterior (MAP) estimation:
\begin{equation}\label{eq:codeMAPgeneral}
\hat{\bm{z}_i}=\argmax_{\bm{z}_i}p_{\theta}(\bm{z}_i|X_i)=\argmax_{\bm{z}_i}\log p_{\theta}(\bm{z}_i|X_i)\,.
\end{equation}
We estimate $\theta$ as the parameters that maximizes the posterior across all shapes:
\begin{align} \label{eq:MAP}
\hat{\theta}&=\argmax_{\theta}\sum_{X_i\in \bm{X}} \max_{\bm{z}_i}\log p_{\theta}(\bm{z}_i|X_i) \\ \nonumber
& =\argmax_{\theta}\sum_{X_i\in \bm{X}} \max_{\bm{z}_i} (\log p_{\theta}(X_i|\bm{z}_i) + \log p(\bm{z}_i))\,,
\end{align}
where the second equality follows from Bayes Law.

For each shape $X_i$ defined via point and SDF samples $(\bm{x}_j,\bm{s}_j)$ as defined in 
Eq.~\ref{eq:datapoint} we make a conditional independence assumption given the code $z_i$ to arrive at the decomposition of the posterior $ p_{\theta}(X_i|z_i)$:
\begin{equation} 
p_{\theta}(X_i | z_i)=\prod_{(\bm{x}_j,\bm{s}_j)\in X_i}p_{\theta}(\bm{s}_j | z_i; \bm{x}_j)\,.
\end{equation} 
Note that the individual SDF likelihoods $p_{\theta}(\bm{s}_j | z_i; \bm{x}_j)$ are parameterized by the sampling location $\bm{x}_j$.

To derive the proposed auto-decoder-based DeepSDF approach we express the SDF likelihood via a deep feed-forward network $f_\theta(\bm{z}_i,\bm{x}_j)$ and, without loss of generality, assume that the likelihood takes the form:
\begin{equation}
   p_{\theta}(\bm{s}_j | z_i; \bm{x}_j) = \exp(-\mathcal{L}(f_\theta (\bm{z}_i,\bm{x}_j),s_j)) \,.
\end{equation}
The SDF prediction $\tilde{s}_j = f_\theta (\bm{z}_i,\bm{x}_j)$ is represented using a fully-connected network and $\mathcal{L}(\tilde{s}_j, s_j)$ is a loss function penalizing the deviation of the network prediction from the actual SDF value $s_j$. One example for the cost function is the standard $L_2$ loss function which amounts to assuming Gaussian noise on the SDF values. In practice we use the clamped $L_1$ cost introduced in the main manuscript.

In the latent shape-code space, we assume the prior distribution over codes $p(\bm{z_i})$ to be a zero-mean multivariate-Gaussian with a spherical covariance $\sigma^2 I$. 
Note that other more complex priors could be assumed.
This leads to the final cost function via Eq.~\ref{eq:MAP} which we jointly minimize with respect to the network parameters $\theta$ and the shape codes $\{z_i\}_{i=1}^N$:
\begin{equation} \label{eq:objective}
\argmin_{\theta, \{\bm{z}_i\}_{i=1}^N} \sum_{i=1}^N \left( \sum_{j=1}^K \mathcal{L}(f_\theta (\bm{z}_i,\bm{x}_j),s_j)+\frac{1}{\sigma^{2}}||\bm{z}_i ||_2^2 \right).
\end{equation}

At inference time, we are given SDF point samples $X$ of one underlying shape to estimate the latent code $\bm{z}$ describing the shape. Using the MAP formulation from Eq.~\ref{eq:codeMAPgeneral} with fixed network parameters $\theta$ we arrive at:
 \begin{equation} \label{eq:testtime2}
\hat{\bm{z}}= \argmin_{\bm{z}} \sum_{(\bm{x}_j,\bm{s}_j)\in X} \mathcal{L}(f_\theta (\bm{z},\bm{x}_j),s_j)+\frac{1}{\sigma^{2}}||\bm{z}||_2^2,
\end{equation}
where $\frac{1}{\sigma^2}$ can be used to balance the reconstruction and regularization term.
For additional comments and insights as well as the practical implementation of the network and its training refer to the main manuscript.

\section{Details on Quantitative Evaluations}\label{sec: quantitative}

\subsection{Preparation for Benchmarked Methods}

\subsubsection{DeepSDF}

For quantitative evaluations we converted the DeepSDF model for a given shape into a mesh by using Marching Cubes \cite{lorensen1987marching} with $512^3$ resolution.  Note that while this was done for quantitative evaluation as a mesh, many of the qualitative renderings are instead produced by raycasting directly against the continuous SDF model, which can avoid some of the artifacts produced by Marching Cubes at finite resolution.  For all experiments in representing known or unknown shapes, DeepSDF was trained on ShapeNet v2, while all shape completion experiments were trained on ShapeNet v1, to match 3D-EPN. Additional DeepSDF training details are provided in Sec.~\ref{sec: training-testing}.

\subsubsection{OGN}

For OGN we trained the provided decoder model (``shape\_from\_id'') for 300,000 steps on the same train set of cars used for DeepSDF.  To compute the point-based metrics, we took the pair of both the groundtruth 256-voxel training data provided by the authors, and the generated 256-voxel output, and converted both of these into point clouds of only the surface voxels, with one point for each of the voxels' centers.  Specifically, surface voxels were defined as voxels which have at least one of 6 direct (non-diagonal) voxel neighbors unoccupied.  A typical number of vertices in the resulting point clouds is approximately 80,000, and the points used for evaluation are randomly sampled from these sets.  Additionally, OGN was trained based on ShapeNet v1, while AtlasNet was trained on ShapeNet v2.  To adjust for the scale difference, we converted OGN point clouds into ShapeNet v2 scale for each model.

\subsubsection{AtlasNet}

Since the provided pretrained AtlasNet models were trained multi-class, we instead trained separate AtlasNet models for each evaluation.  Each model was trained with the available code by the authors with all default parameters, except for the specification of class for each model and matching train/test splits with those used for DeepSDF.  The quality of the models produced from these trainings appear comparable to those in the original paper.

Of note, we realized that AtlasNet's own computation of its training and evaluation metric, Chamfer distance, had the limitation that only the vertices of the generated mesh were used for the evaluation.  This leaves the triangles of the mesh unconstrained in that they can connect across what are supposed to be holes in the shape, and this would not be reflected in the metric.  Our evaluation of meshes produced by AtlasNet instead samples evenly from the mesh surface, i.e. each triangle in the mesh is weighted by its surface area, and points are sampled from the triangle faces.

\subsubsection{3D-EPN}

We used the provided shape completion inference results for 3D-EPN, which is in voxelized distance function format. We subsequently extracted the isosurface using MATLAB as described in the paper to obtain the final mesh.

\subsection{Metrics}\label{sec: metrics}

The first two metrics, Chamfer and Earth Mover's, are easily applicable to points, meshes (by sampling points from the surface) and voxels (by sampling surface voxels and treating their centers as points). When meshes are available, we also can compute metrics suited particularly for meshes: mesh accuracy, mesh completion, and mesh cosine similarity. \\ 

\textbf{Chamfer distance} is a popular metric for evaluating shapes, perhaps due to its simplicity \cite{fan2017point}. Given two point sets $S_1$ and $S_2$, the metric is simply the sum of the nearest-neighbor distances for each point to the nearest point in the other point set.
$$d_{CD}(S_1,S_2) = \sum_{x \in S_1} \underset{y \in S_2}{\min}||x-y||_2^2 + \sum_{y \in S_2} \underset{x \in S_1}{\min}||x-y||_2^2$$
Note that while sometimes the metric is only defined one-way (i.e., just $\sum_{x \in S_1} \underset{y \in S_2}{min}||x-y||_2^2$) and this is not symmetric, the sum of both directions, as defined above, is symmetric: $d_{CD}(S_1,S_2) =d_{CD}(S_2,S_1)$.  Note also that the metric is not technically a valid distance function since it does not satisfy the triangle inequality, but is commonly used as a psuedo distance function \cite{fan2017point}. In all of our experiments we report the Chamfer distance for 30,000 points for both $|S_1|$ and $|S_2|$, which can be efficiently computed by use of a KD-tree, and akin to prior work \cite{groueix2018atlasnet} we normalize by the number of points: we report $\frac{d_{CD}(S_1,S_2)}{30,000}$.\\

\textbf{Earth Mover's distance} \cite{rubner1998metric}, also known as the Wasserstein distance, is another popular metric for measuring the difference between two discrete distributions.  Unlike the Chamfer distance, which does not require any constraints on the correspondences between evaluated points, for the Earth Mover's distance a bijection $\phi: S_1 \rightarrow S_2$, i.e. a one-to-one correspondence, is formed.  Formally, for two point sets $S_1$ and $S_2$ of equal size $|S_1| = |S_2|$, the metric is defined via the optimal bijection \cite{fan2017point}:

$$d_{EMD}(S_1,S_2) = \underset{\phi: S_1 \rightarrow S_2}{\min} \sum_{x \in S_1} || x- \phi(x) ||_2$$

Although the metric is commonly approximated in the deep learning literature \cite{fan2017point} by distributed approximation schemes \cite{bertsekas1985distributed} for speed during training, we compute the metric accurately for evaluation using a more modest number of point samples (500) using \cite{pyemd}.

In practice the intuitive, important difference between the Chamfer and Earth Mover's metrics is that the Earth Mover's metric more favors distributions of points that are similarly evenly distributed as the ground truth distribution.  A low Chamfer distance may be achieved by assigning just one point in $S_2$ to a cluster of points in $S_1$, but to achieve a low Earth Mover's distance, each cluster of points in $S_1$ requires a comparably sized cluster of points in $S_2$.\\

\textbf{Mesh accuracy}, as defined in \cite{seitz2006comparison}, is the minimum distance $d$ such that 90\% of generated points are within $d$ of the ground truth mesh.  We used 1,000 points sampled evenly from the generated mesh surface, and computed the minimum distances to the \textit{full} ground truth mesh. To clarify, the distance is computed to the closest point on any face of the mesh, not just the vertices.  Note that unlike Chamfer and Earth Mover's metrics which require sampling of points from both meshes, with this metric the entire mesh for the ground truth is used -- accordingly this metric has lower variance than for example Chamfer distance computed with only 1,000 points from each mesh.  Note also that mesh accuracy does not measure how \textit{complete} the generated mesh is -- a low (good) mesh accuracy can be achieved by only generating one small portion of the ground truth mesh, ignoring the rest.  Accordingly, it is ideal to pair mesh accuracy with the following metric, mesh completion. \\

\textbf{Mesh completion}, also as defined in \cite{seitz2006comparison}, is the fraction of points sampled from the ground truth mesh that are within some distance $\Delta$ (a parameter of the metric) to the generated mesh.  We used $\Delta=0.01$, which well measured the differences in mesh completion between the different methods.  With this metric the {\em full} generated mesh is used, and points (we used 1,000) are sampled from the ground truth mesh (mesh accuracy is vice versa). Ideal mesh completion is 1.0, minimum is 0.0.\\

\textbf{Mesh cosine similarity} is a metric we introduce to measure the accuracy of mesh normals. We define the metric as the mean cosine similarity between the normals of points sampled from the ground truth mesh, and the normals of the nearest faces of the generated mesh. More precisely, given the generated mesh $M_{gen}$ and a set of points with normals $S_{gt}$ sampled from the ground truth mesh, for each point $x_i$ in $S_{gt}$ we look up the closest face $F_{i}$ in $M_{gen}$, and then compute the average cosine similarity between the normals associated with $x_i$ and $F_{i}$, 
%
$$ \text{Cos. sim}(M_{gen},S_{gt}) = \frac{1}{|S_{gt}|} \sum_{x_i \in S_{gt}} \hat{n}_{F_i} \cdot \hat{n}_{x_i} \ , $$ 
where 
each $\hat{n} \in \mathbb{R}^3$ is a unit-norm normal vector. We use $|S_{gt}|=2,500$ and in order to allow for \cite{groueix2018atlasnet} which does not provide oriented normals, we compute the $\min(\cdot)$ over both the generated mesh normal and its flipped normal: $\min(\hat{n}_{F_i} \cdot \hat{n}_{x_i}, -\hat{n}_{F_i} \cdot \hat{n}_{x_i} )$.  Ideal cosine similarity is 1.0, minimum (given the allowed flip of the normal) is 0.0.

\section{Additional Results}\label{sec: additional}
\subsection{Representing Unseen Objects}
\begin{figure*} 
\begin{subfigure}[t]{0.20\linewidth}
\includegraphics[width=0.8\linewidth]{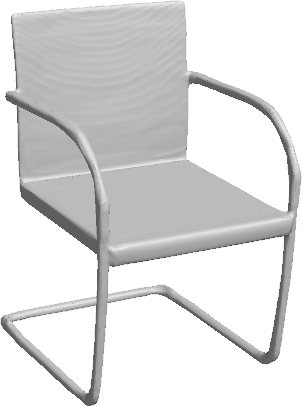}
\end{subfigure}
\hfill
\begin{subfigure}[t]{0.20\linewidth}
\includegraphics[width=0.8\linewidth]{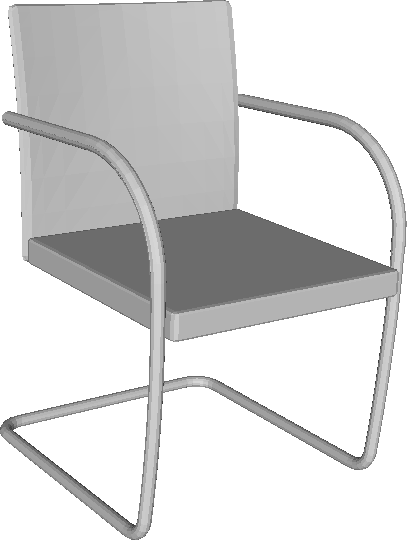}
\end{subfigure}
\hfill
\begin{subfigure}[t]{0.28\linewidth}
\includegraphics[width=0.8\linewidth]{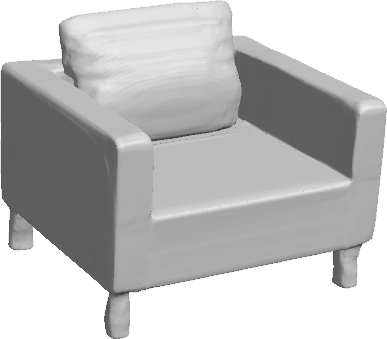}
\end{subfigure}
\hfill
\begin{subfigure}[t]{0.28\linewidth}
\includegraphics[width=0.8\linewidth]{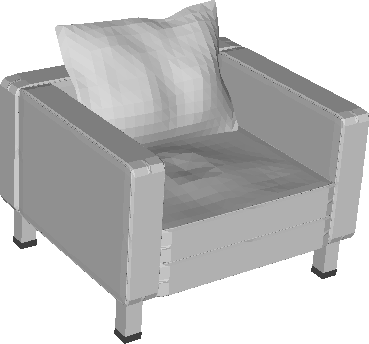}
\end{subfigure}
\hfill

\begin{subfigure}[t]{0.25\linewidth}
\includegraphics[width=0.8\linewidth]{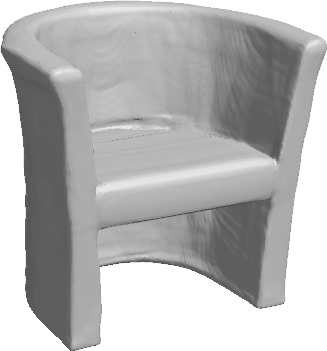}
\end{subfigure}
\hfill
\begin{subfigure}[t]{0.25\linewidth}
\includegraphics[width=0.8\linewidth]{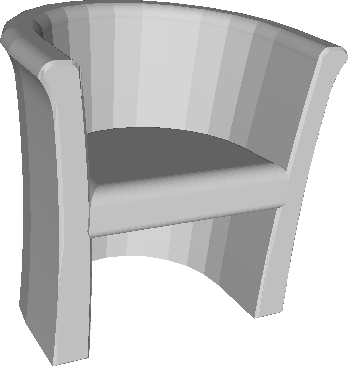}
\end{subfigure}
\hfill
\begin{subfigure}[t]{0.20\linewidth}
\includegraphics[width=0.8\linewidth]{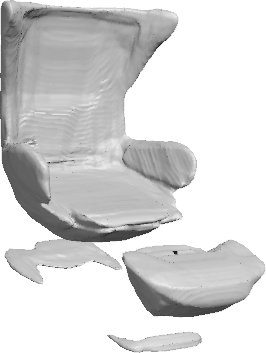}
\end{subfigure}
\hfill
\begin{subfigure}[t]{0.20\linewidth}
\includegraphics[width=0.8\linewidth]{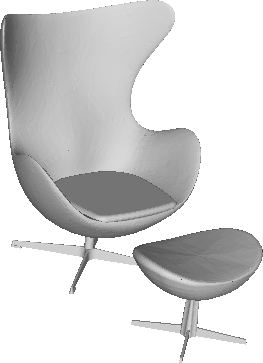}
\end{subfigure}
\hfill
	\caption{Additional test shape reconstruction results. Left to right alternatingly: DeepSDF reconstruction and ground truth.}
	\label{fig: recon}
\end{figure*}

\begin{figure*} 
\begin{subfigure}[t]{0.20\linewidth}
\includegraphics[width=0.8\linewidth]{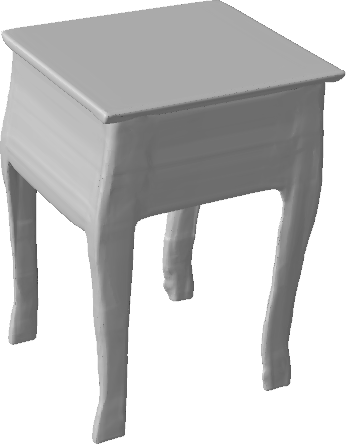}
\end{subfigure}
\hfill
\begin{subfigure}[t]{0.22\linewidth}
\includegraphics[width=0.8\linewidth]{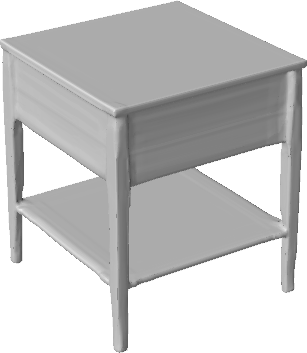}
\end{subfigure}
\hfill
\begin{subfigure}[t]{0.36\linewidth}
\includegraphics[width=0.8\linewidth]{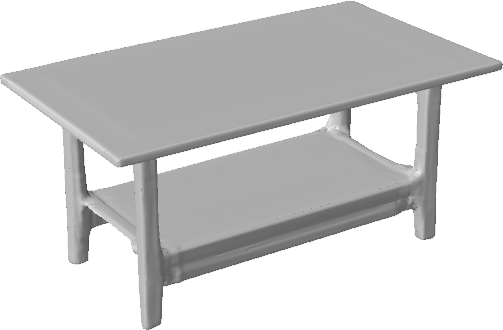}
\end{subfigure}
\hfill
\begin{subfigure}[t]{0.20\linewidth}
\includegraphics[width=0.8\linewidth]{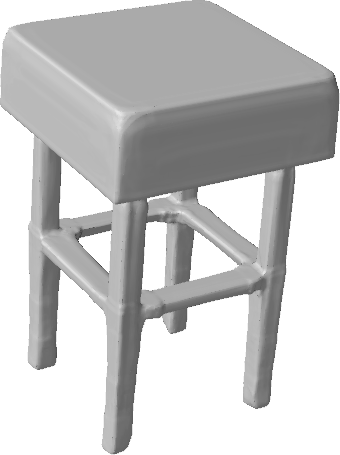}
\end{subfigure}
\hfill

\begin{subfigure}[t]{0.26\linewidth}
\includegraphics[width=0.8\linewidth]{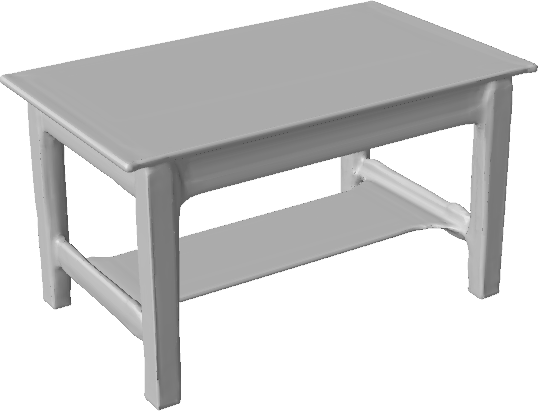}
\end{subfigure}
\hfill
\begin{subfigure}[t]{0.22\linewidth}
\includegraphics[width=0.8\linewidth]{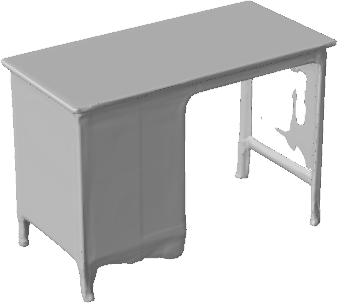}
\end{subfigure}
\hfill
\begin{subfigure}[t]{0.28\linewidth}
\includegraphics[width=0.8\linewidth]{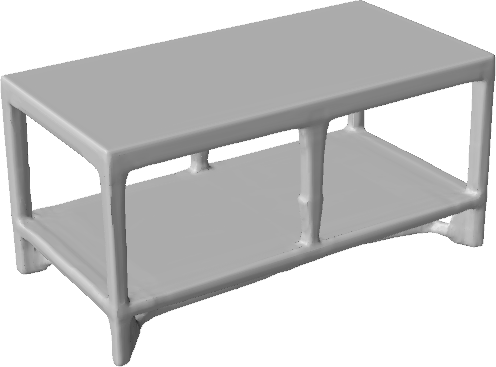}
\end{subfigure}
\hfill
\begin{subfigure}[t]{0.20\linewidth}
\includegraphics[width=0.8\linewidth]{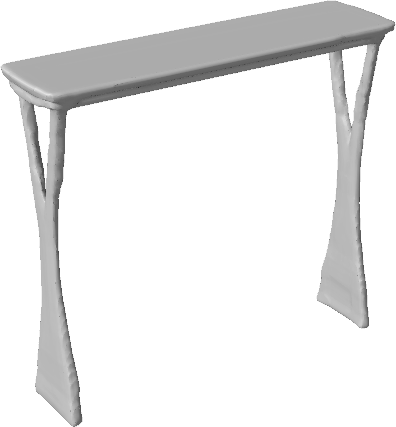}
\end{subfigure}
\hfill
	\caption{Additional test shape reconstruction results for Table ShapeNet class. All of the above images are test shapes represented with our DeepSDF network during inference time, showing the accuracy and expressiveness of the shape embedding.}
	\label{fig: recon2}
\end{figure*}

We provide additional results on representing test objects with trained DeepSDF (Fig. \ref{fig: recon}, \ref{fig: recon2}). We provide additional data with the additional metrics, mesh completion and mesh cosine similarity, for the comparison of methods contained in the manuscript (Tab.~\ref{tab: unknowncompletion}). 
The success of this task for DeepSDF implies that 1) high quality shapes similar to the test shapes exist in the embedding space, and 2) the codes for the shapes can be found through simple gradient descent.

 \begin{table}[t]
\centering
\footnotesize
 \begin{tabular}{|l|c|c|c|c|c|}
 \hhline{|=|=|=|=|=|=|}
  \multicolumn{1}{|l|}{Mesh comp., mean} & \multicolumn{1}{c|}{chair} & \multicolumn{1}{c|}{plane} & \multicolumn{1}{c|}{table} & \multicolumn{1}{c|}{lamp} & \multicolumn{1}{c|}{sofa}  \\
 \hline
 AtlasNet-Sph.  & 0.668 & 0.862 & 0.755 & 0.281 & 0.641  \\
 AtlasNet-25      & 0.723 & 0.887 & 0.785 & 0.528 & 0.681  \\
 DeepSDF        & \bf{0.947} & \bf{0.943} & \bf{0.959} & \bf{0.877} & \bf{0.931}  \\
 \hline
   \hhline{|=|=|=|=|=|=|}
  \multicolumn{1}{|l}{Mesh comp., median} &  \multicolumn{1}{c}{ \ } &  \multicolumn{1}{c}{ \ } &  \multicolumn{1}{c}{ \ } &  \multicolumn{1}{c}{ \ }  &  \multicolumn{1}{c|}{} \\
 \hline
  AtlasNet-Sph.  & 0.686 & 0.930 & 0.795 & 0.257 & 0.666  \\
 AtlasNet-25      & 0.736 & 0.944 & 0.825 & 0.533 & 0.702  \\
 DeepSDF        & \bf{0.970} & \bf{0.970} & \bf{0.982} & \bf{0.930} & \bf{0.941}  \\
 \hline
  \hhline{|=|=|=|=|=|=|}
  \multicolumn{1}{|l}{Cosine sim., mean} &  \multicolumn{1}{c}{ \ } &  \multicolumn{1}{c}{ \ } &  \multicolumn{1}{c}{ \ } &  \multicolumn{1}{c}{ \ }  &  \multicolumn{1}{c|}{} \\
 \hline
  AtlasNet-Sph.  & 0.790 & 0.840 & 0.826 & 0.719 & 0.847  \\
 AtlasNet-25      & 0.797 & 0.858 & 0.835 & 0.725 & 0.826  \\
 DeepSDF        & \bf{0.896} & \bf{0.907} & \bf{0.916} & \bf{0.862} & \bf{0.917}  \\
 \hline

   \end{tabular}
   \caption{Comparison of metrics  for representing unknown shapes (U) for various classes of ShapeNet. Mesh completion as defined in \cite{seitz2006comparison} i.e. the fraction of groundtruth sampled points that are within a $\Delta$ (we used $\Delta=0.01$) of the generated mesh, and mean cosine similarity is of normals for nearest groundtruth-generated point pairs.  Cosine similarity is defined in \ref{sec: metrics}.  Higher is better for all metrics in this table.} 
   \label{tab: unknowncompletion}
\end{table}

\subsection{Shape Completions}
Finally we present additional shape completion results on unperturbed depth images of synthetic ShapeNet dataset (Fig. \ref{fig: completion}), demonstrating the quality of the auto-decoder learning scheme and the new shape representation.

\begin{figure*} 
\begin{subfigure}[t]{0.245\linewidth}
\includegraphics[width=0.8\linewidth]{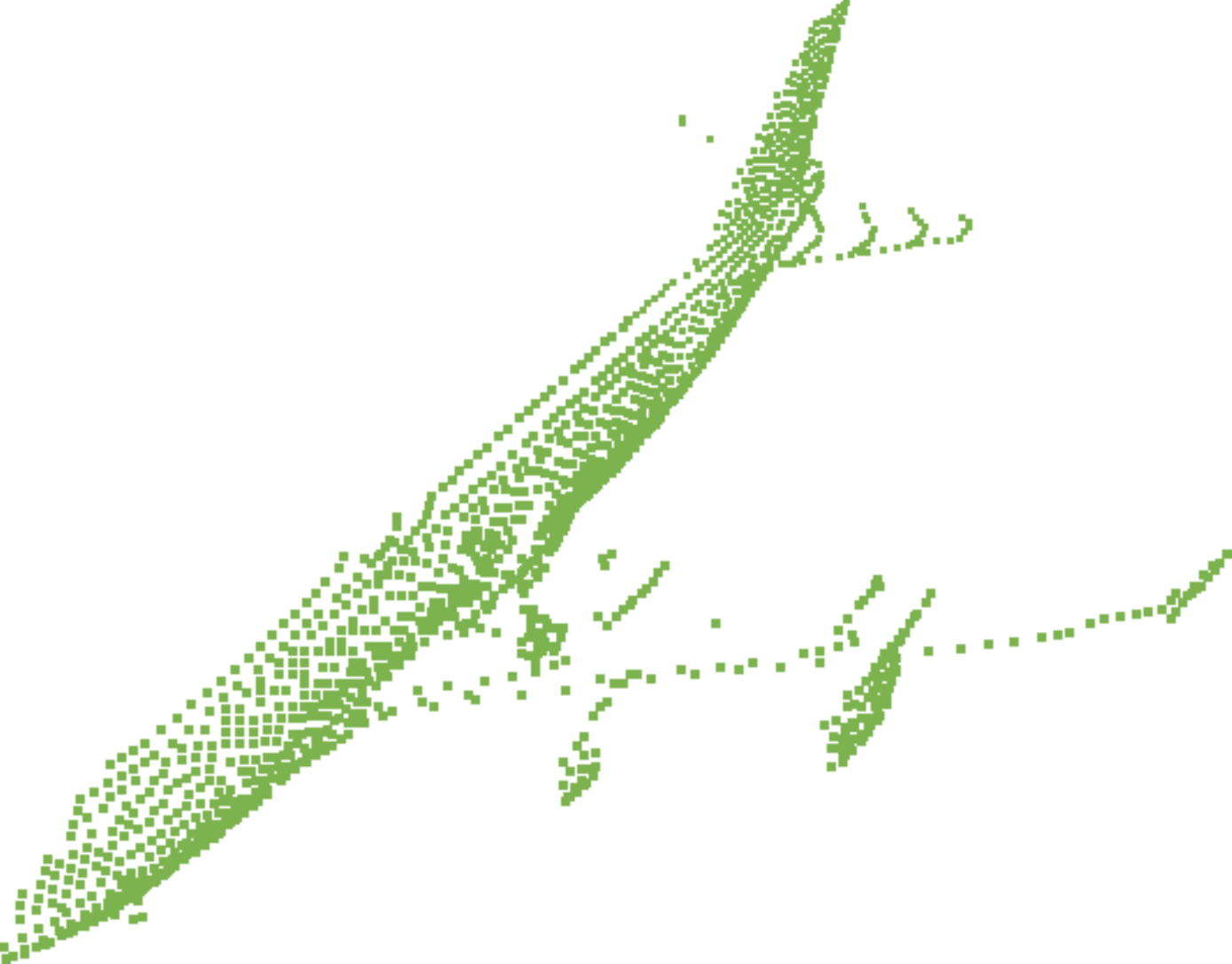}
\end{subfigure}
\hfill
\begin{subfigure}[t]{0.245\linewidth}
\includegraphics[width=0.8\linewidth]{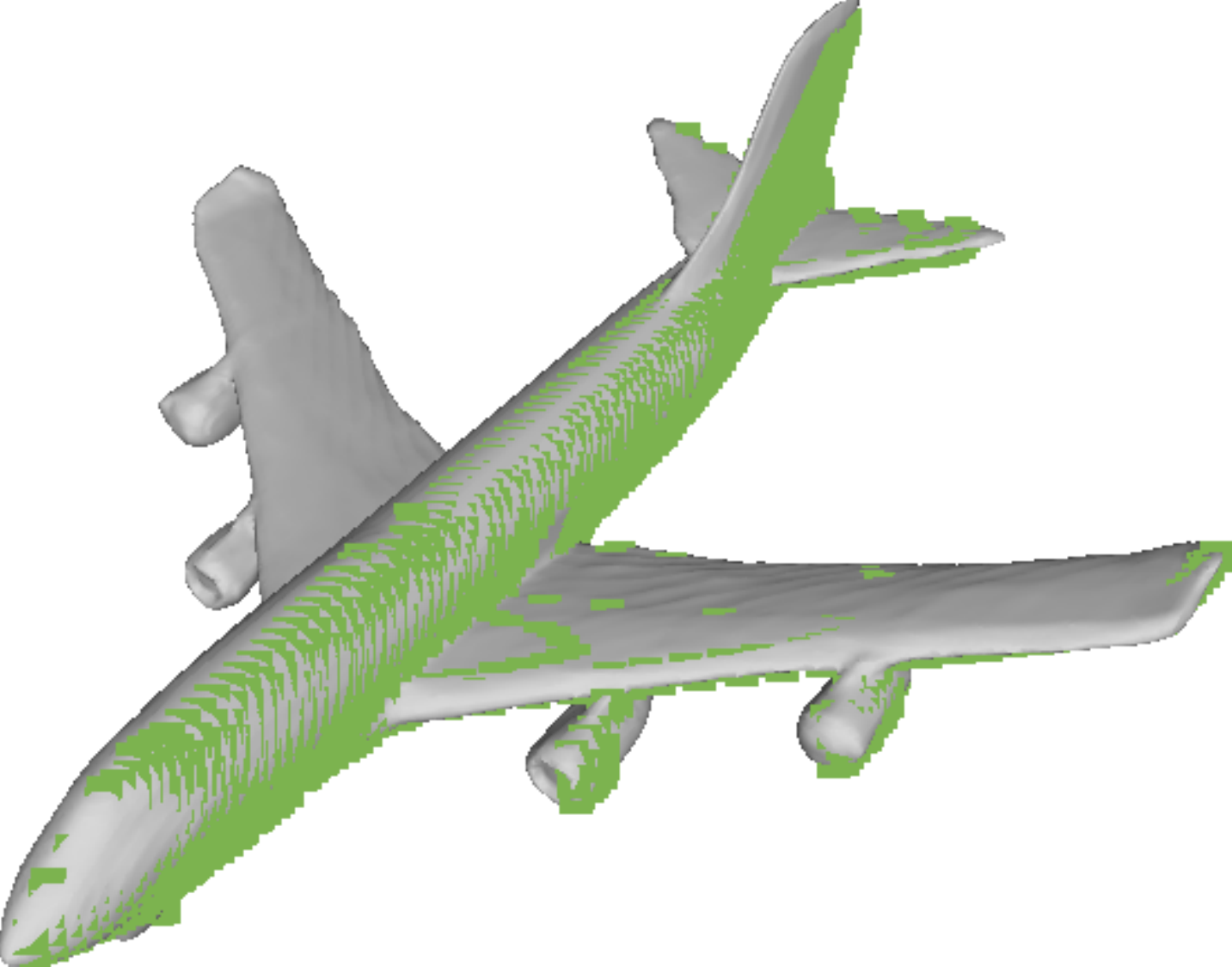}
\end{subfigure}
\hfill
\begin{subfigure}[t]{0.245\linewidth}
\includegraphics[width=0.8\linewidth]{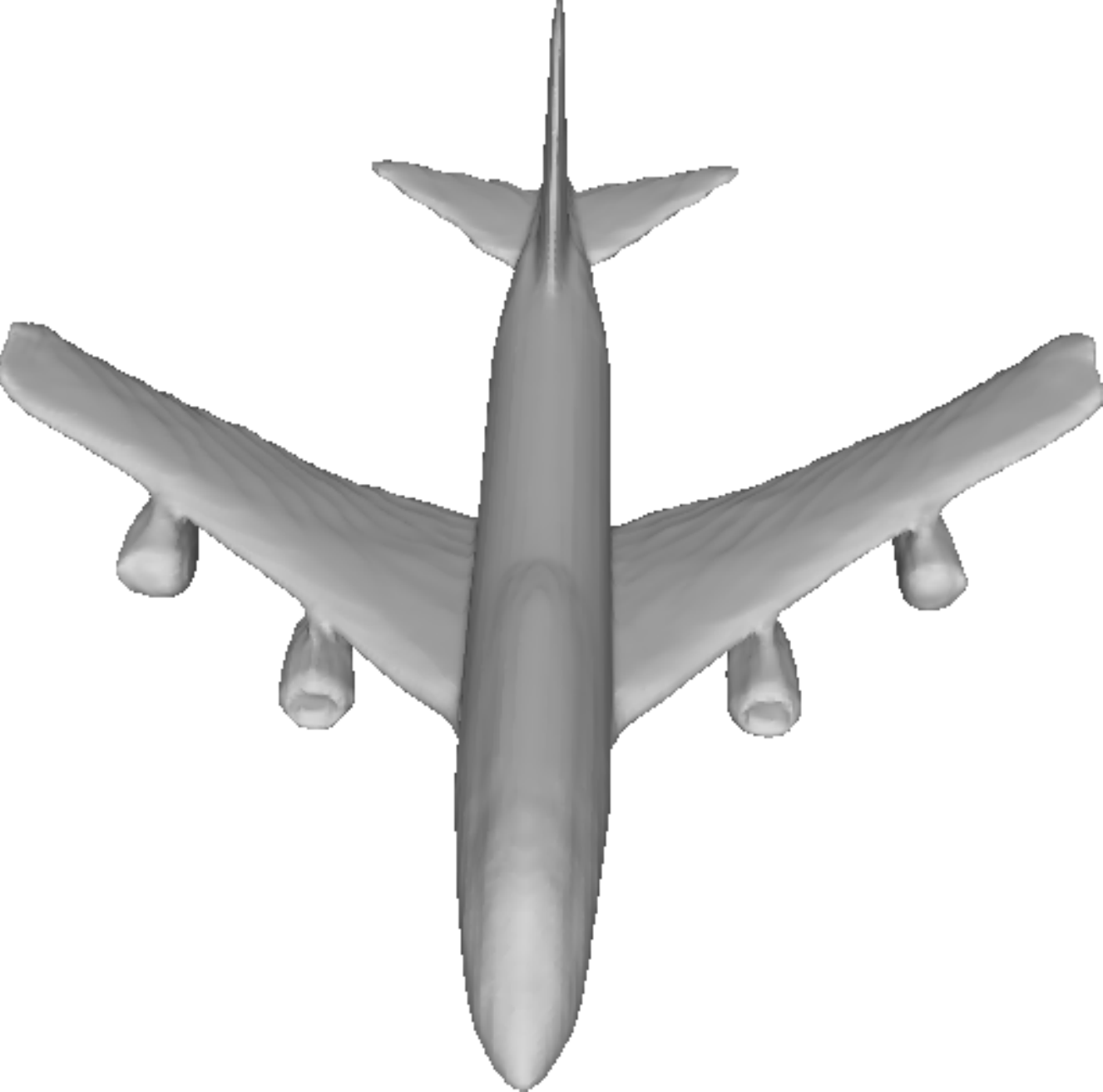}
\end{subfigure}
\hfill
\begin{subfigure}[t]{0.245\linewidth}
\includegraphics[width=0.8\linewidth]{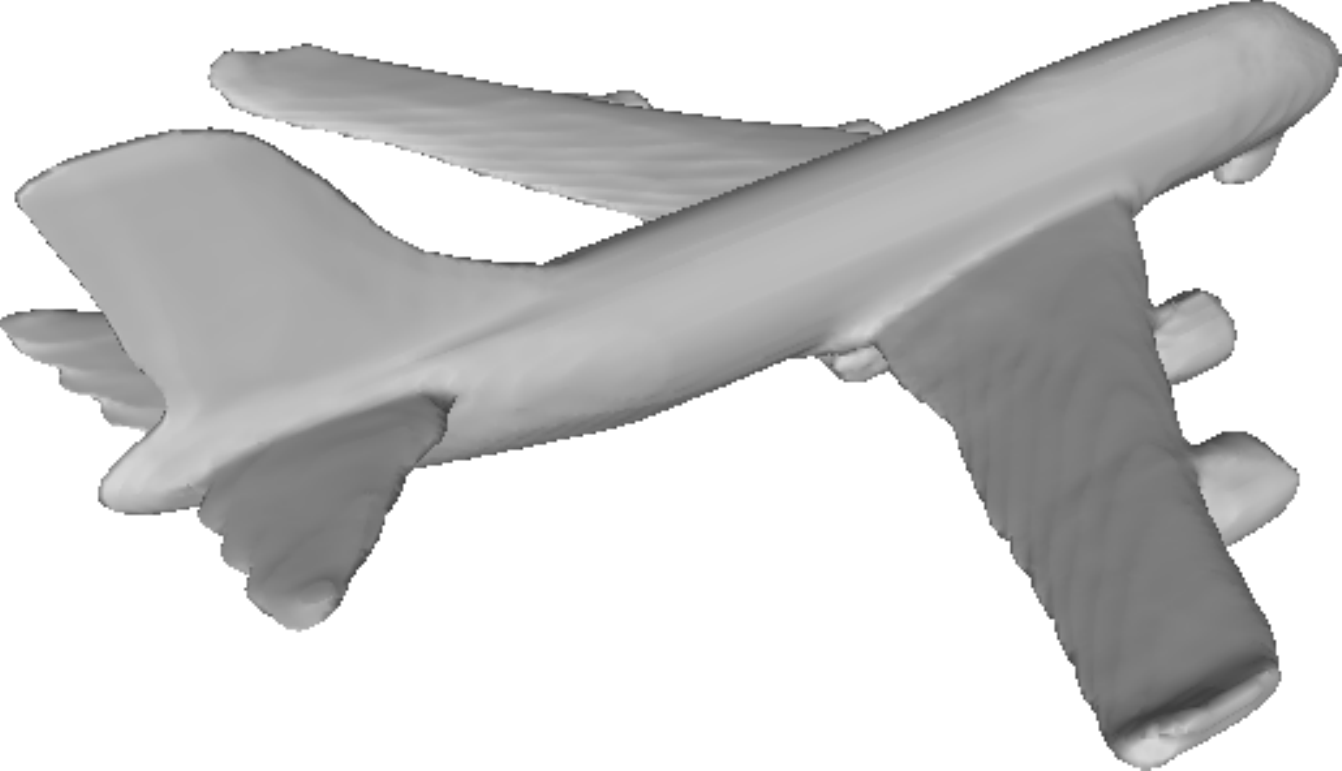}
\end{subfigure}
\hfill

\begin{subfigure}[t]{0.245\linewidth}
\includegraphics[width=0.8\linewidth]{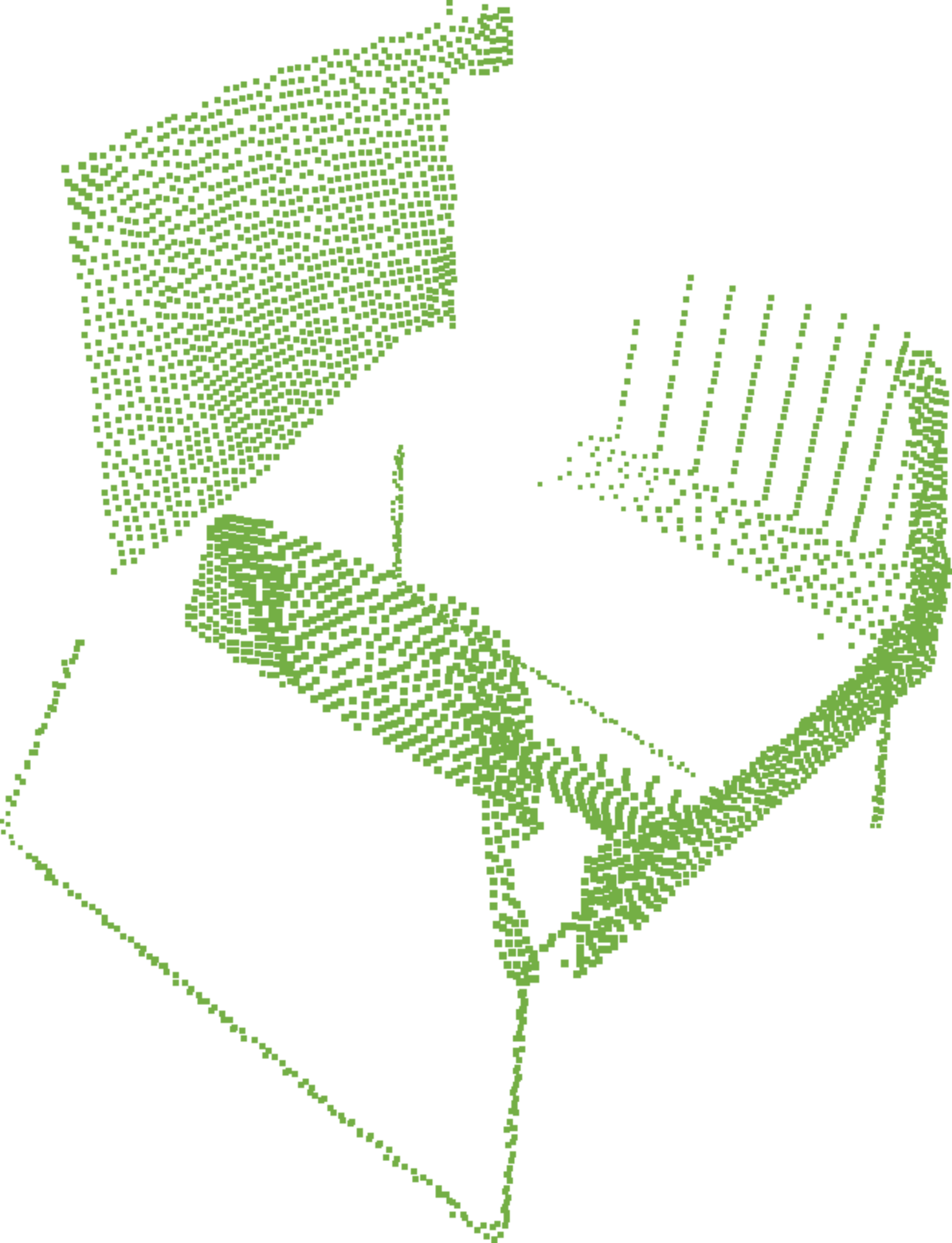}
\end{subfigure}
\hfill
\begin{subfigure}[t]{0.245\linewidth}
\includegraphics[width=0.8\linewidth]{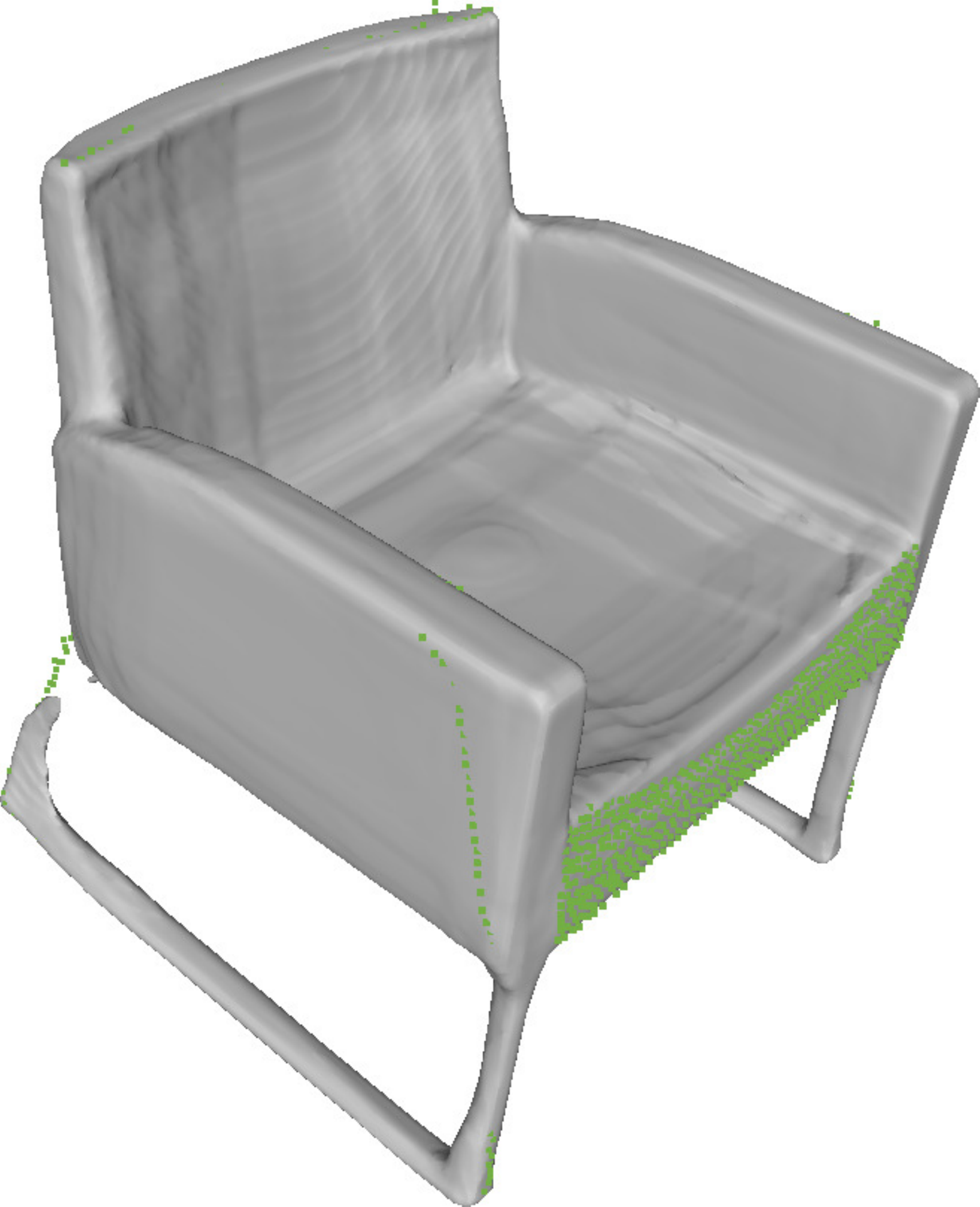}
\end{subfigure}
\hfill
\begin{subfigure}[t]{0.245\linewidth}
\includegraphics[width=0.8\linewidth]{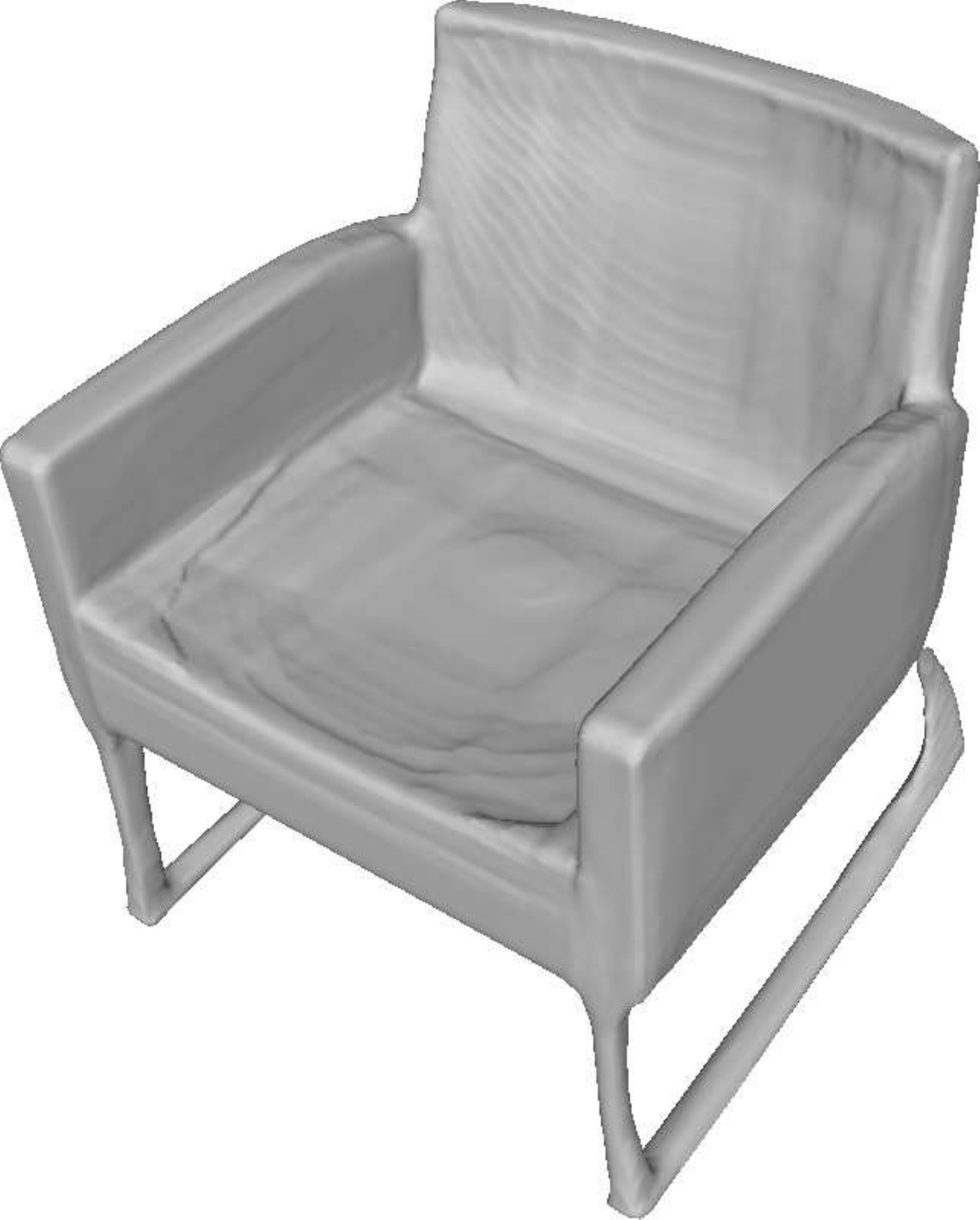}
\end{subfigure}
\hfill
\begin{subfigure}[t]{0.23\linewidth}
\includegraphics[width=0.8\linewidth]{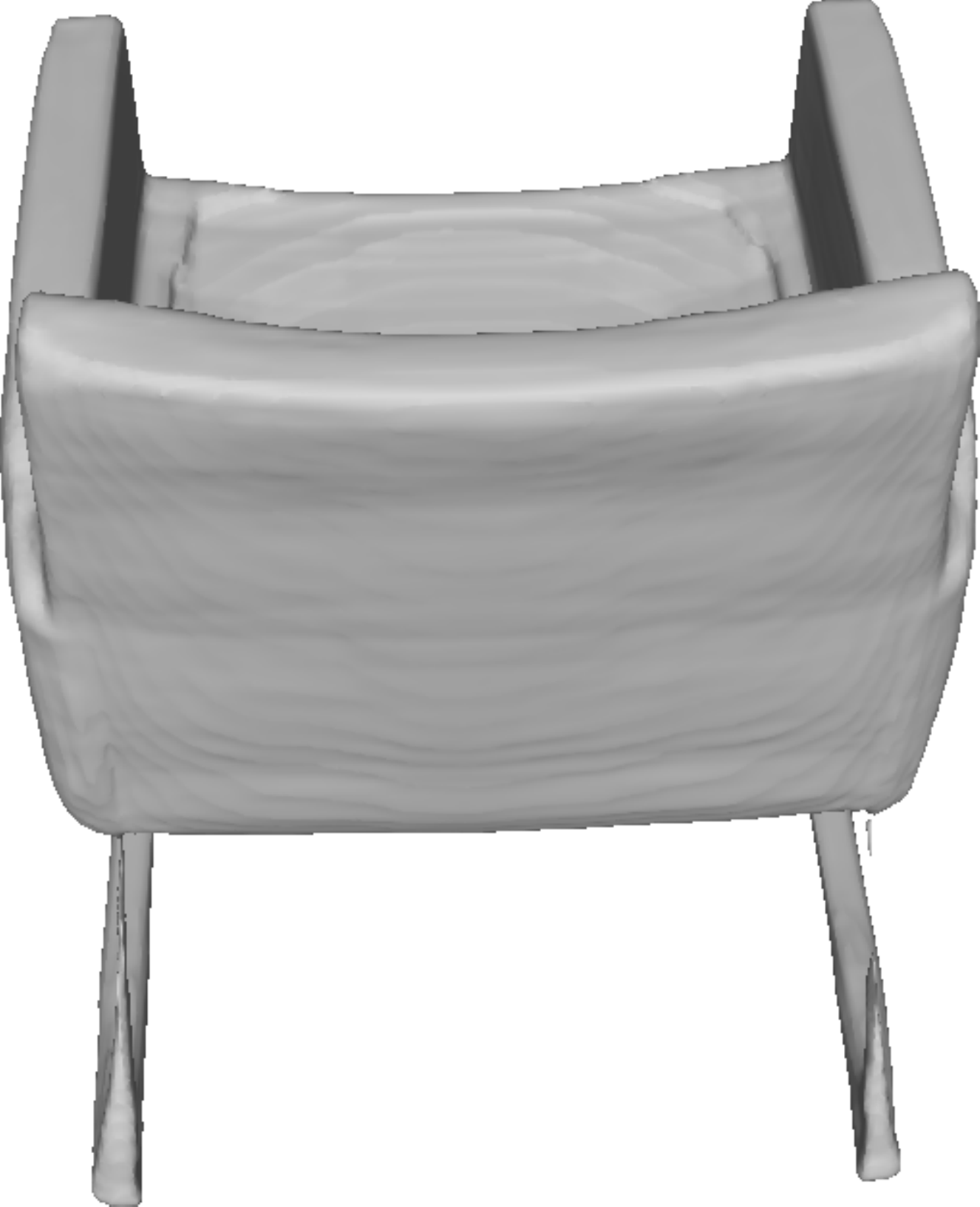}
\end{subfigure}
\hfill

\begin{subfigure}[t]{0.16\linewidth}
\includegraphics[width=0.8\linewidth]{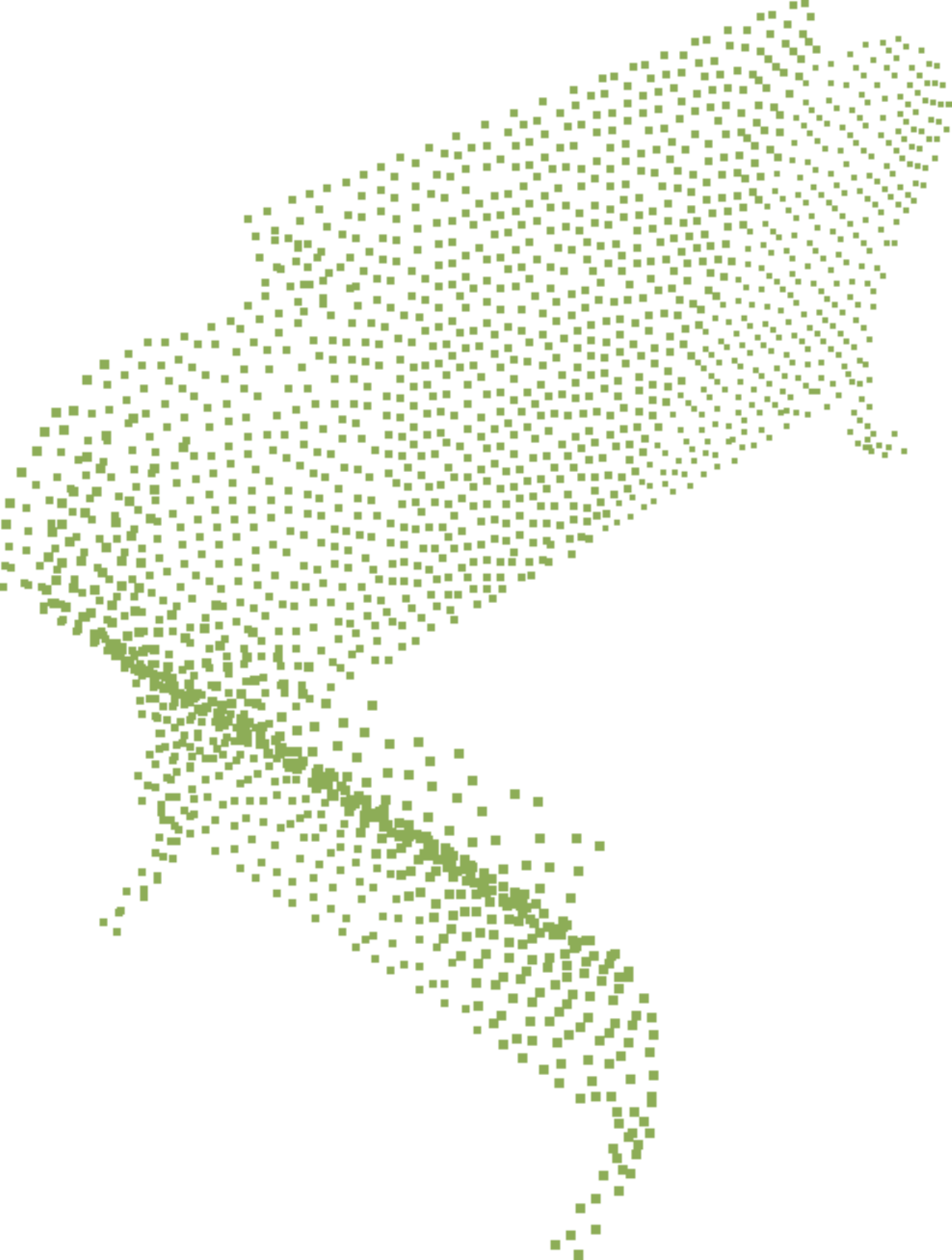}
\end{subfigure}
\hfill
\begin{subfigure}[t]{0.24\linewidth}
\includegraphics[width=0.8\linewidth]{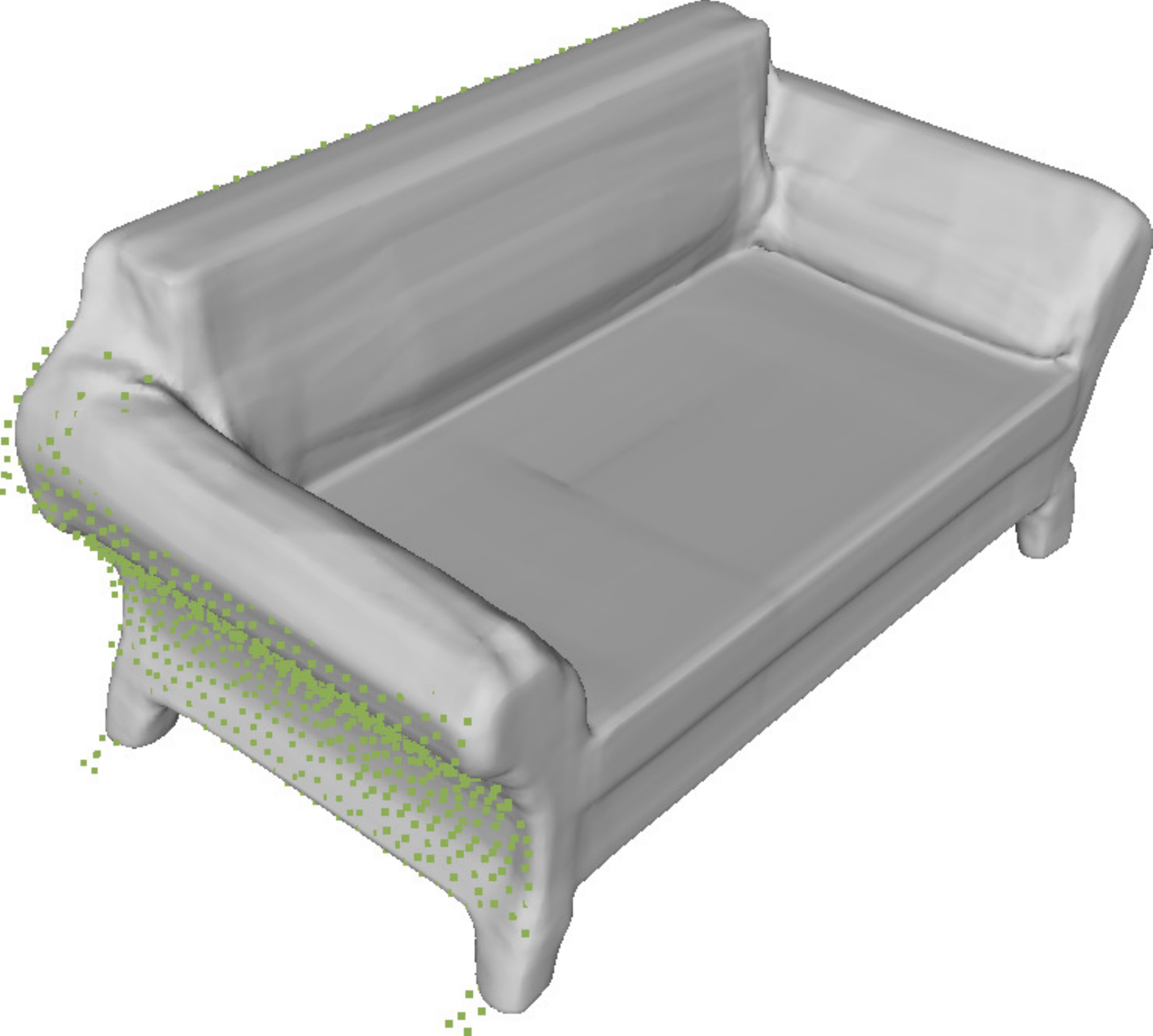}
\end{subfigure}
\hfill
\begin{subfigure}[t]{0.28\linewidth}
\includegraphics[width=0.8\linewidth]{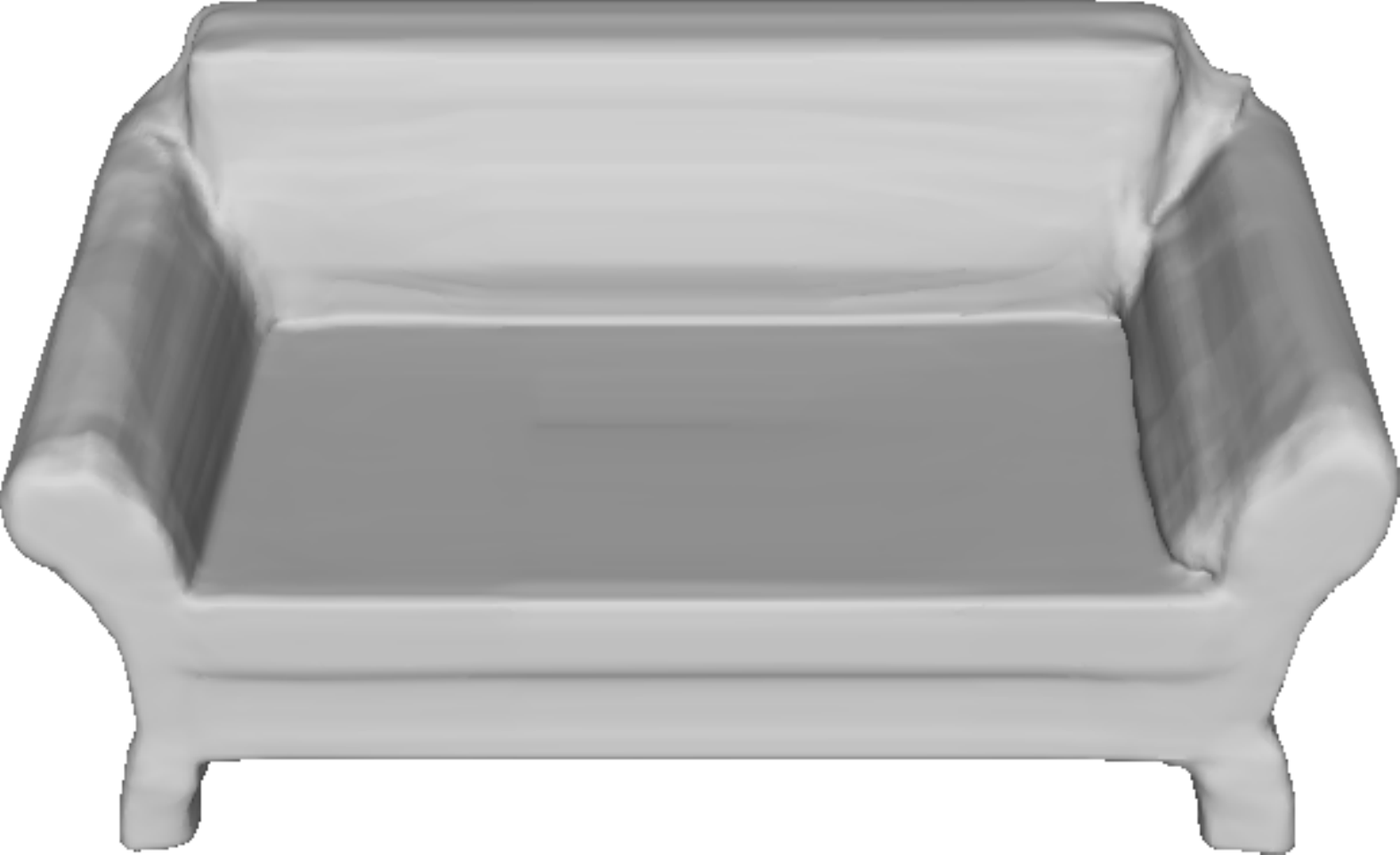}
\end{subfigure}
\hfill
\begin{subfigure}[t]{0.28\linewidth}
\includegraphics[width=0.8\linewidth]{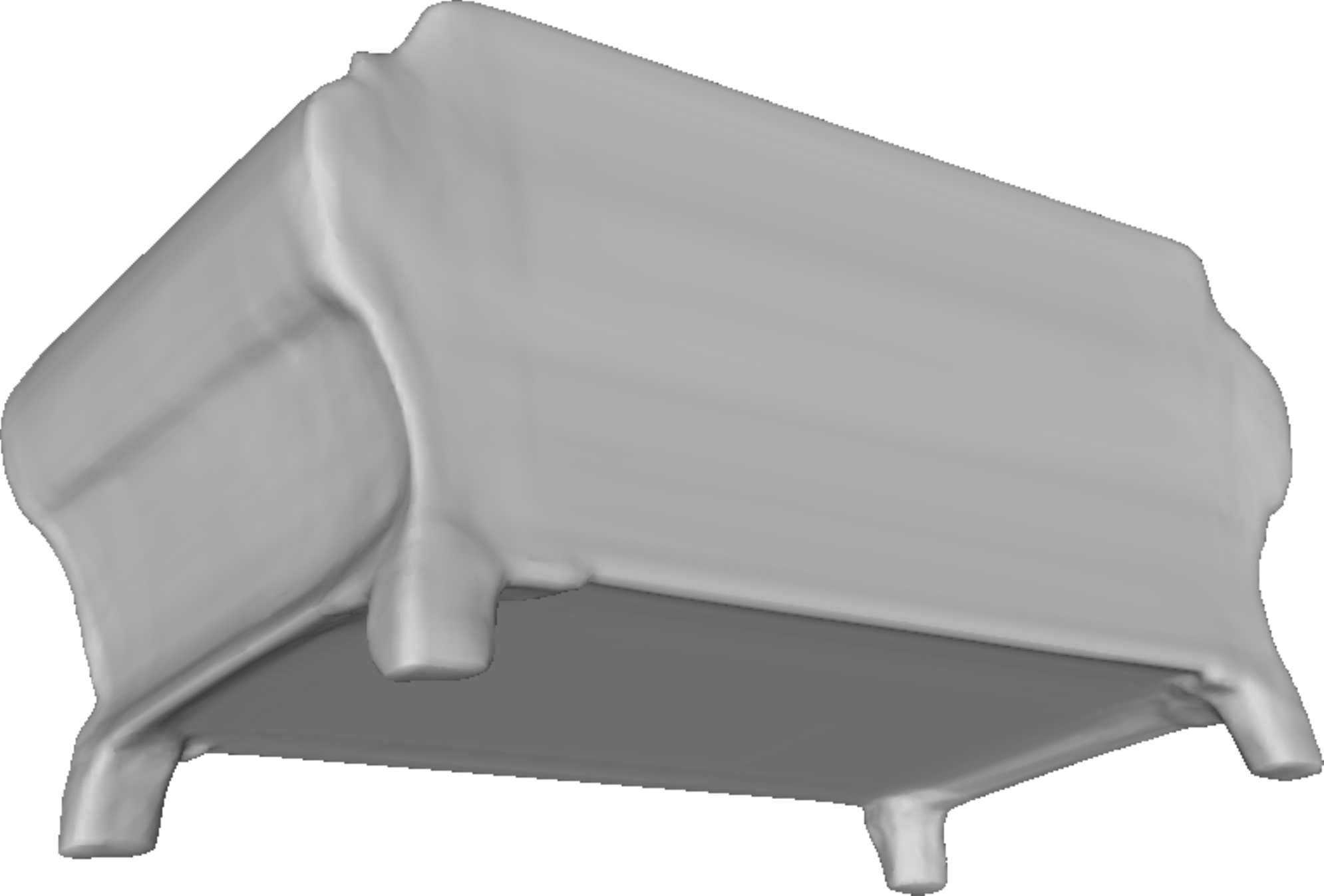}
\end{subfigure}
\hfill

\begin{subfigure}[t]{0.24\linewidth}
\includegraphics[width=0.8\linewidth]{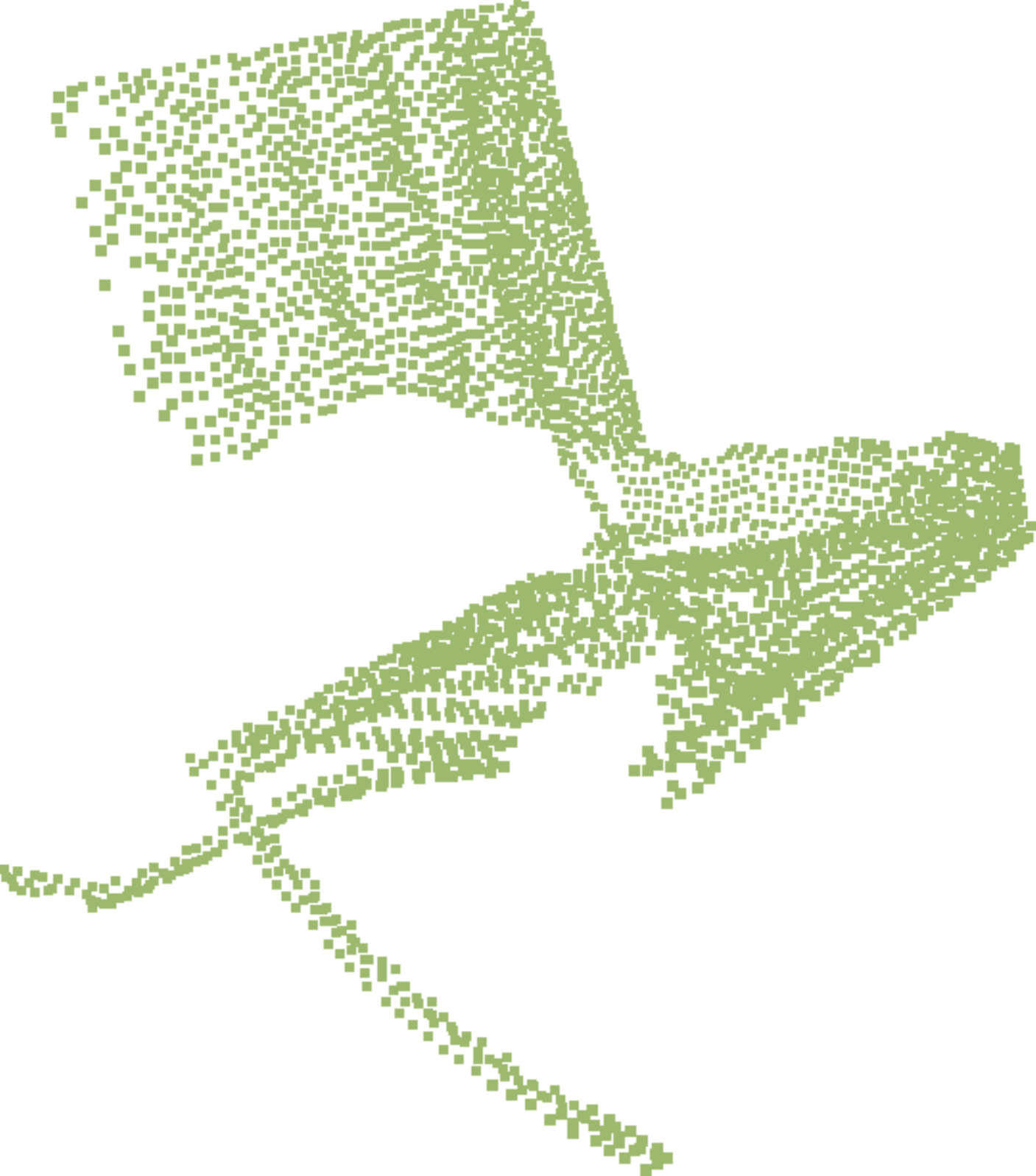}
\end{subfigure}
\hfill
\begin{subfigure}[t]{0.24\linewidth}
\includegraphics[width=0.8\linewidth]{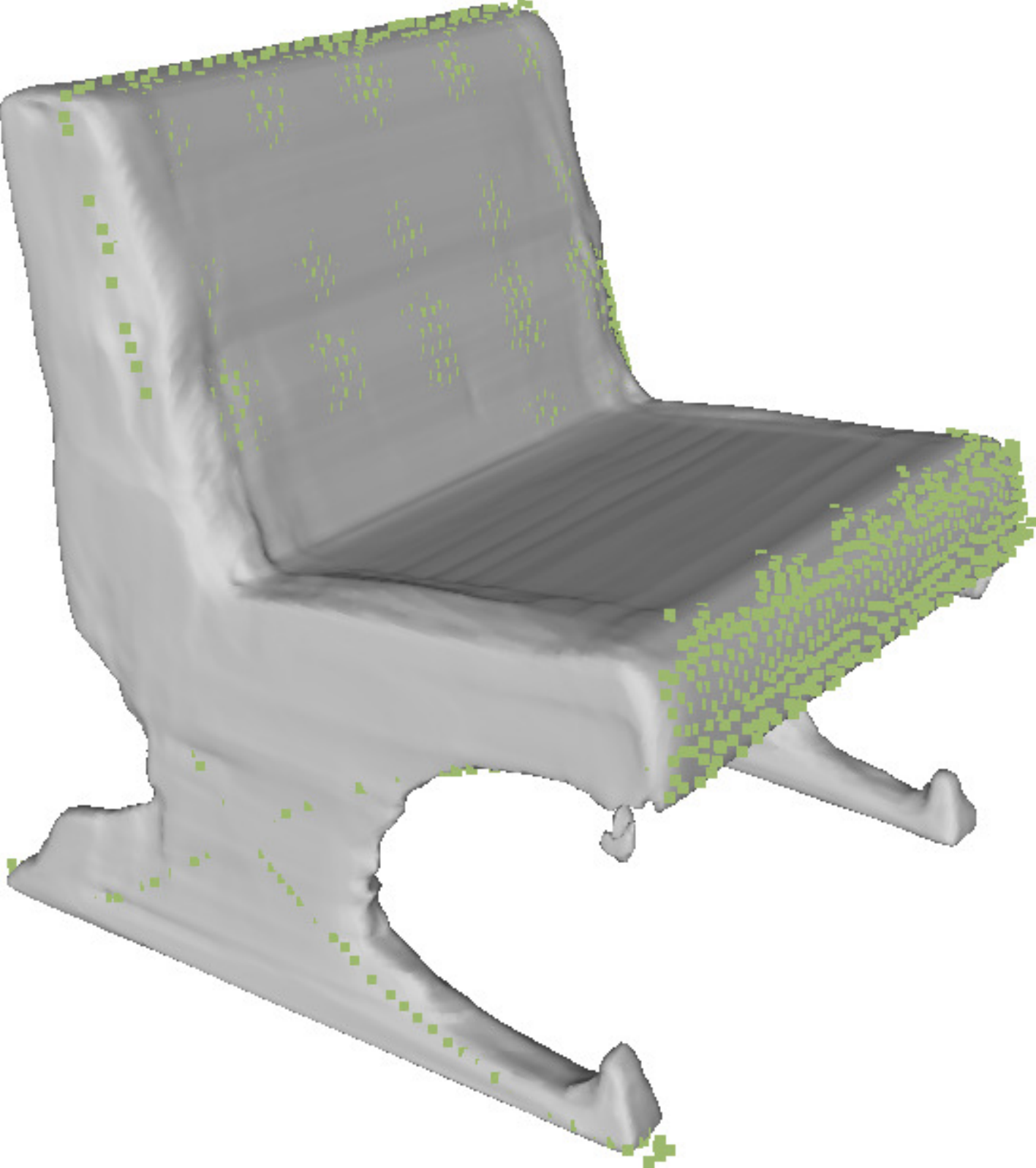}
\end{subfigure}
\hfill
\begin{subfigure}[t]{0.21\linewidth}
\includegraphics[width=0.8\linewidth]{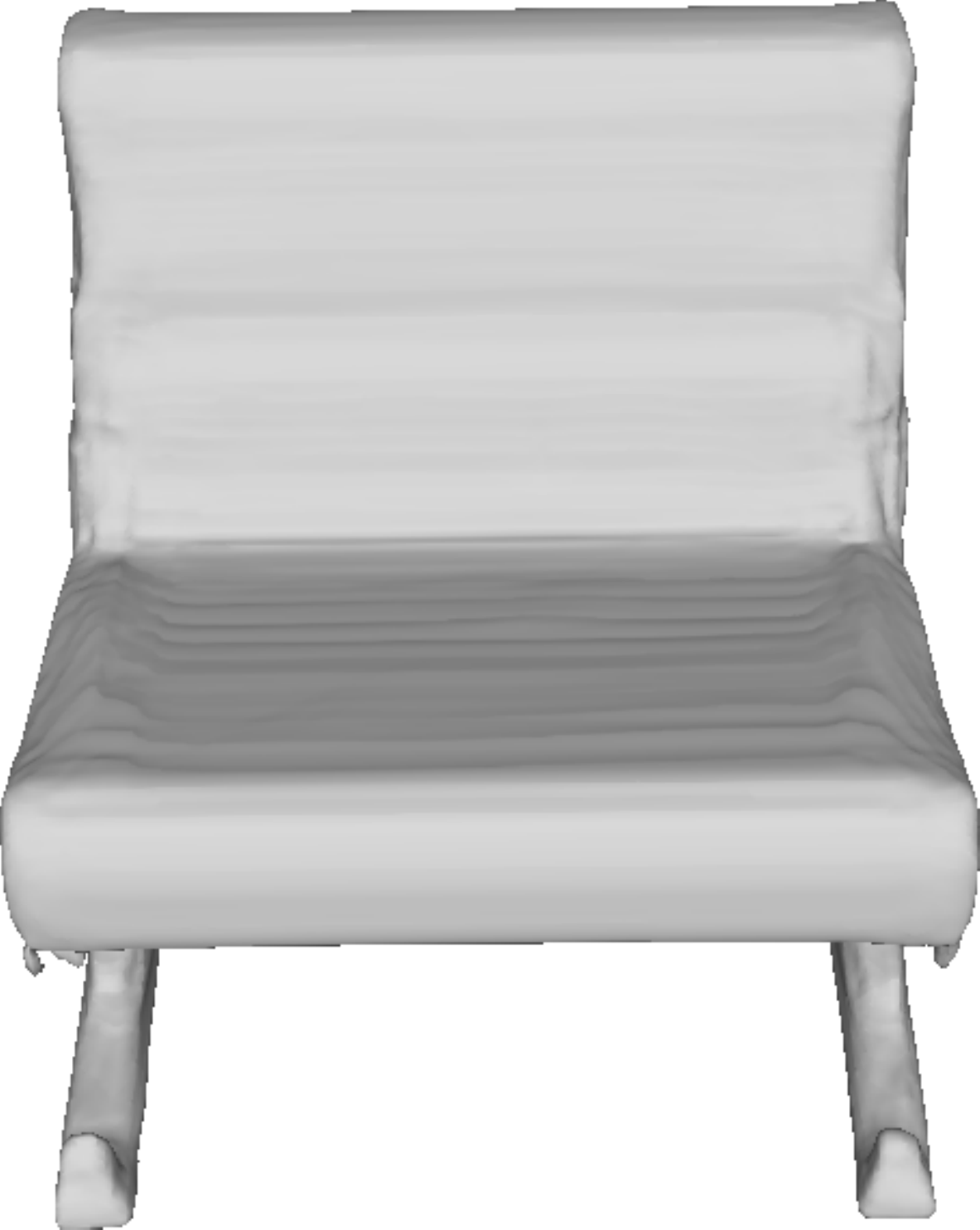}
\end{subfigure}
\hfill
\begin{subfigure}[t]{0.27\linewidth}
\includegraphics[width=0.8\linewidth]{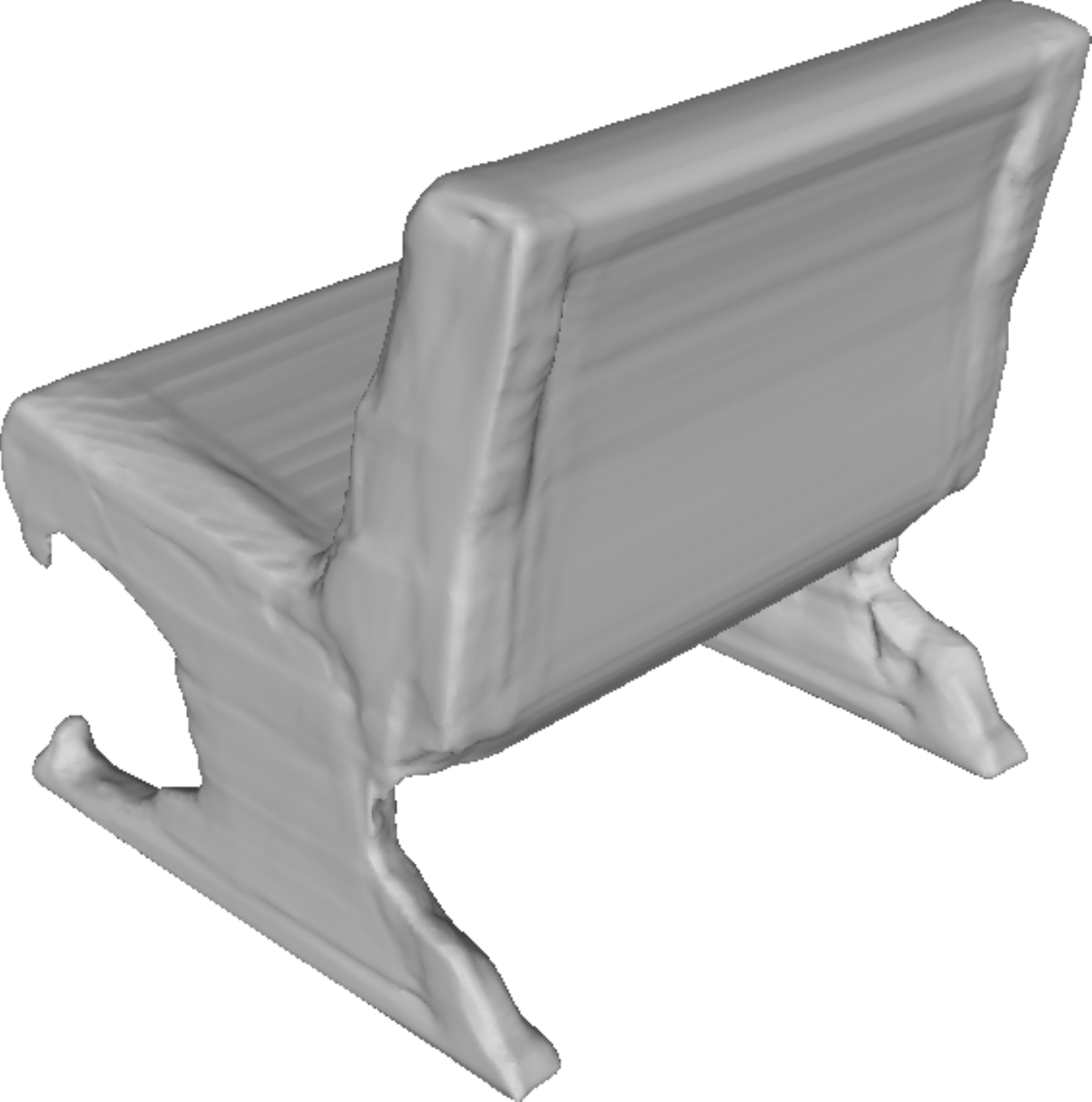}
\end{subfigure}
\hfill
	\caption{Additional shape completion results. Left to Right: input depth point cloud, shape completion using DeepSDF, second view, and third view.}
	\label{fig: completion}
\end{figure*}

{\small
\bibliographystyle{ieee}
\bibliography{egbib}
}

\end{document}